\documentclass[10pt,twocolumn,letterpaper]{article}

\usepackage{cvpr}
\usepackage{times}
\usepackage{epsfig}
\usepackage{graphicx}
\usepackage{amsmath}
\usepackage{amssymb}
\usepackage[utf8]{inputenc} 
\usepackage[usenames]{color}

\usepackage{amsmath}
\usepackage{algorithmic} 
\usepackage{graphicx}
\usepackage{epstopdf}
\usepackage{cite}
\usepackage{algorithm}
\usepackage{amssymb}
\usepackage{epsfig}
\usepackage{array}
\usepackage{subfigure}
\usepackage{bm}
\usepackage{color}
\usepackage{latexsym}
\usepackage{multirow}

\def\swone{0.95\linewidth}

\def\swfive{0.19\linewidth}
\def\swsix{0.164\linewidth}

\usepackage[breaklinks=true,bookmarks=false]{hyperref}

\cvprfinalcopy 

\def\cvprPaperID{12} 

\begin{document}
\title{Unsupervised Image Super-Resolution \\
using Cycle-in-Cycle Generative Adversarial Networks}

\author{
Yuan Yuan$^1$$^2$\thanks{Yuan Yuan and Siyuan Liu are co-first authors. This work was done when they were interns at Sensetime. Contacting email: \href{mailto:yuanyuan@szu.edu.cn}{yuanyuan@szu.edu.cn}}~~
Siyuan Liu$^1$$^3$$^4$~~
Jiawei Zhang$^1$~~
Yongbing Zhang$^3$~~
Chao Dong$^1$~~
Liang Lin$^1$\\
$^1$Sensetime Research\\
$^2$Guangdong Key Laboratory of Intelligent Information Processing, Shenzhen University \\
$^3$Graduate School at Shenzhen, Tsinghua University, Shenzhen \\
$^4$Department of Automation, Tsinghua University, Beijing
$$
}

\maketitle

\begin{abstract}

We consider the single image super-resolution problem in a more general case that the low-/high-resolution pairs and the down-sampling process are unavailable. Different from traditional super-resolution formulation, the low-resolution input is further degraded by noises and blurring. This complicated setting makes supervised learning and accurate kernel estimation impossible. To solve this problem, we resort to unsupervised learning without paired data, inspired by the recent successful image-to-image translation applications. With generative adversarial networks (GAN) as the basic component, we propose a Cycle-in-Cycle network structure to tackle the problem within three steps. First, the noisy and blurry input is mapped to a noise-free low-resolution space. Then the intermediate image is up-sampled with a pre-trained deep model. Finally, we fine-tune the two modules in an end-to-end manner to get the high-resolution output. Experiments on NTIRE2018 datasets demonstrate that the proposed unsupervised method achieves comparable results as the state-of-the-art supervised models.  

\end{abstract}

\section{Introduction}
Recent deep learning based super-resolution (SR) methods have achieved significant improvement either on PSNR values~\cite{singh2014super,huang2015single,yang2010image,kim2010single,
timofte2014a+,ledig2016photo,kim2016accurate,lim2017enhanced} or on visual quality~\cite{ledig2016photo,sajjadi2017enhancenet}. These methods require supervised learning on high-resolution (HR) and low-resolution (LR) image pairs. However, their common assumption that the downscaling factor is known and the input image is noise-free hinders them from practical usages. In real-world scenarios, the SR problem often have the following properties: 1) HR datasets are unavailable, 2) downscaling method is unknown, 3) input LR images are noisy and blurry. This problem is extremely difficult if the input images suffer from different kinds of degradation. For an easier case, in this study, we assume that input images are degraded with the same processing which is complex and unavailable. 

%
%
%
%
%
%

\renewcommand{\tabcolsep}{.0pt}
\begin{figure}
\scriptsize
	\begin{center}
	\includegraphics[width=\swone]{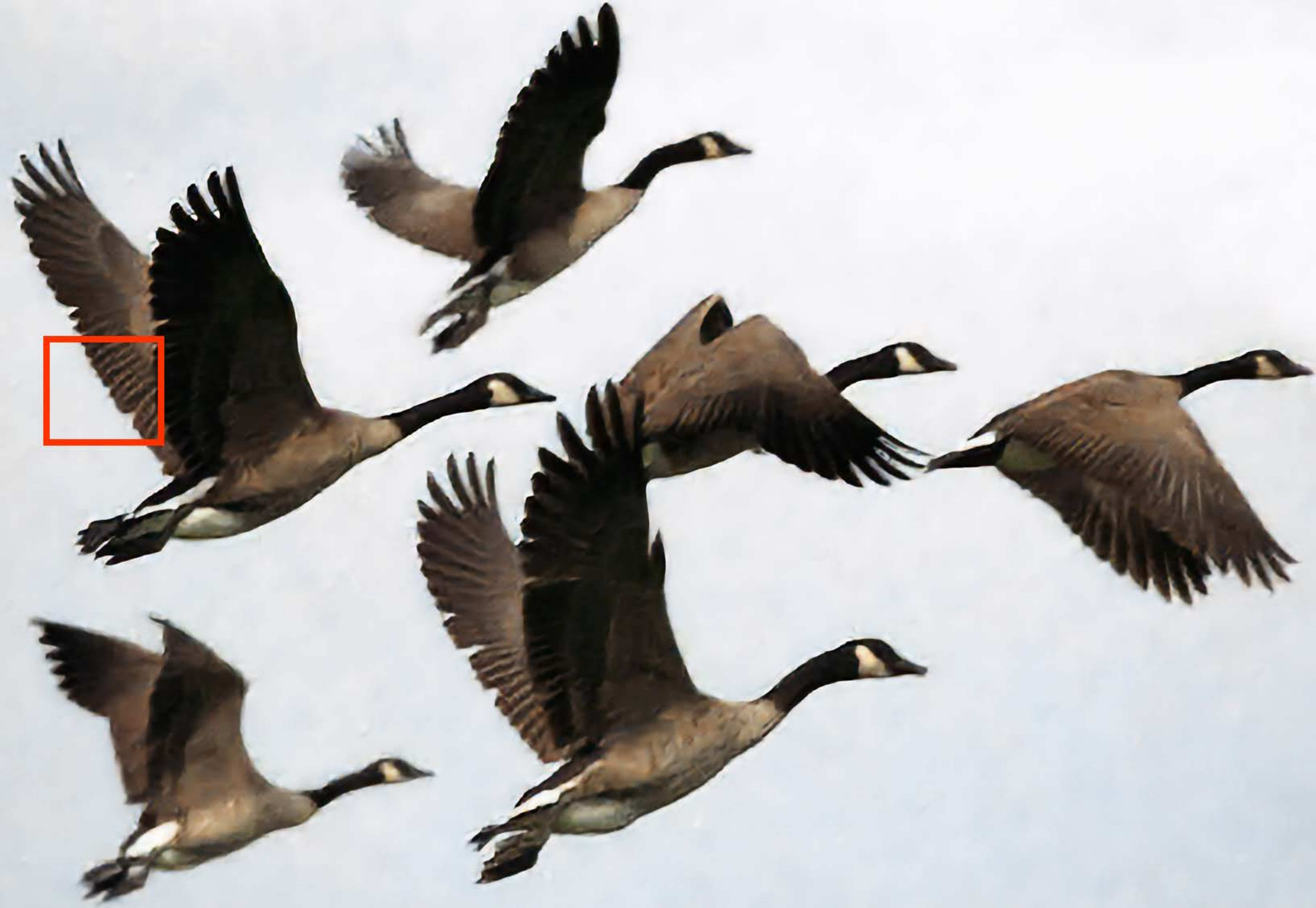}	 \\
		\begin{tabular}{ccccc}
		\includegraphics[width=\swfive]{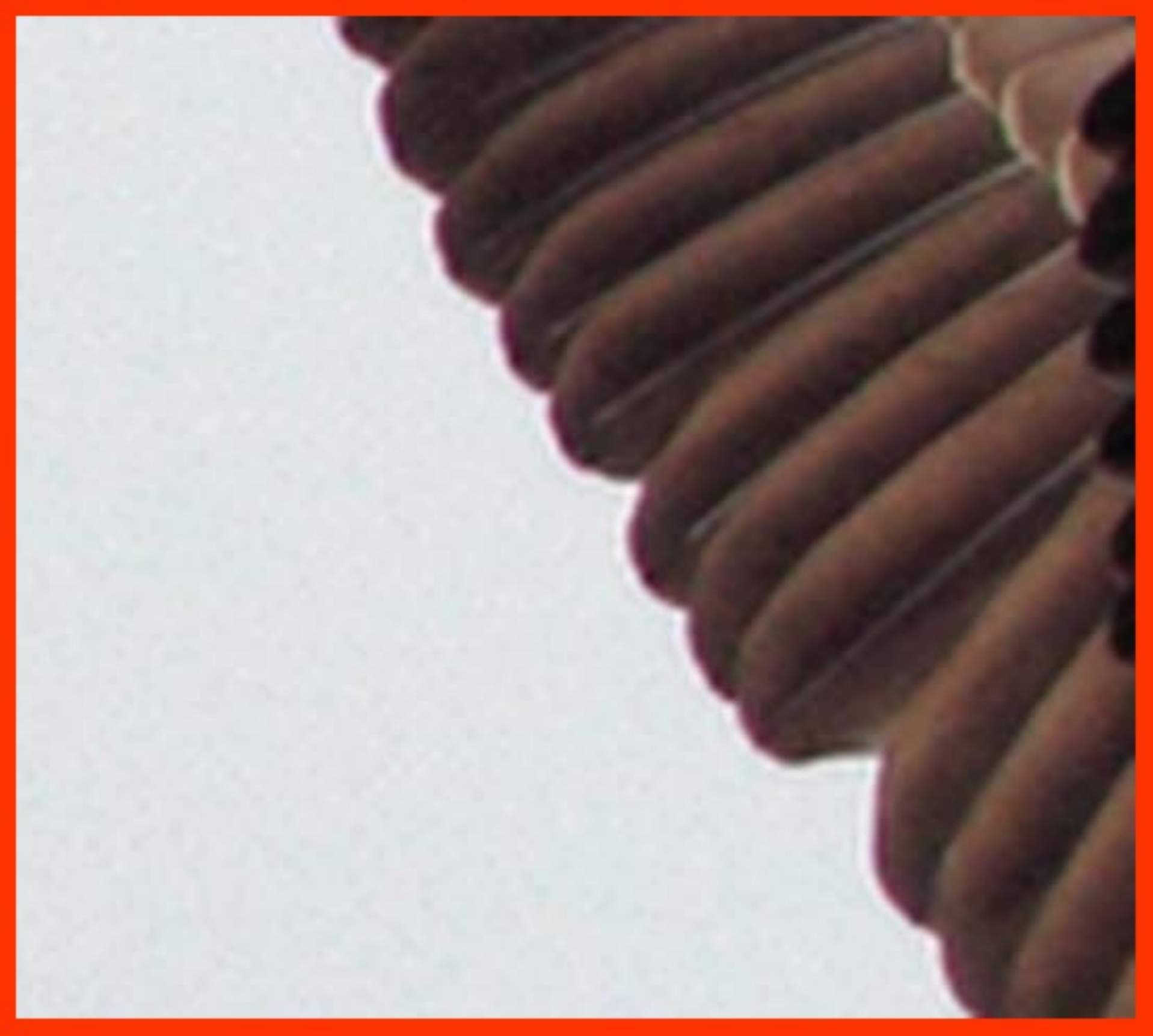} &
		\includegraphics[width=\swfive]{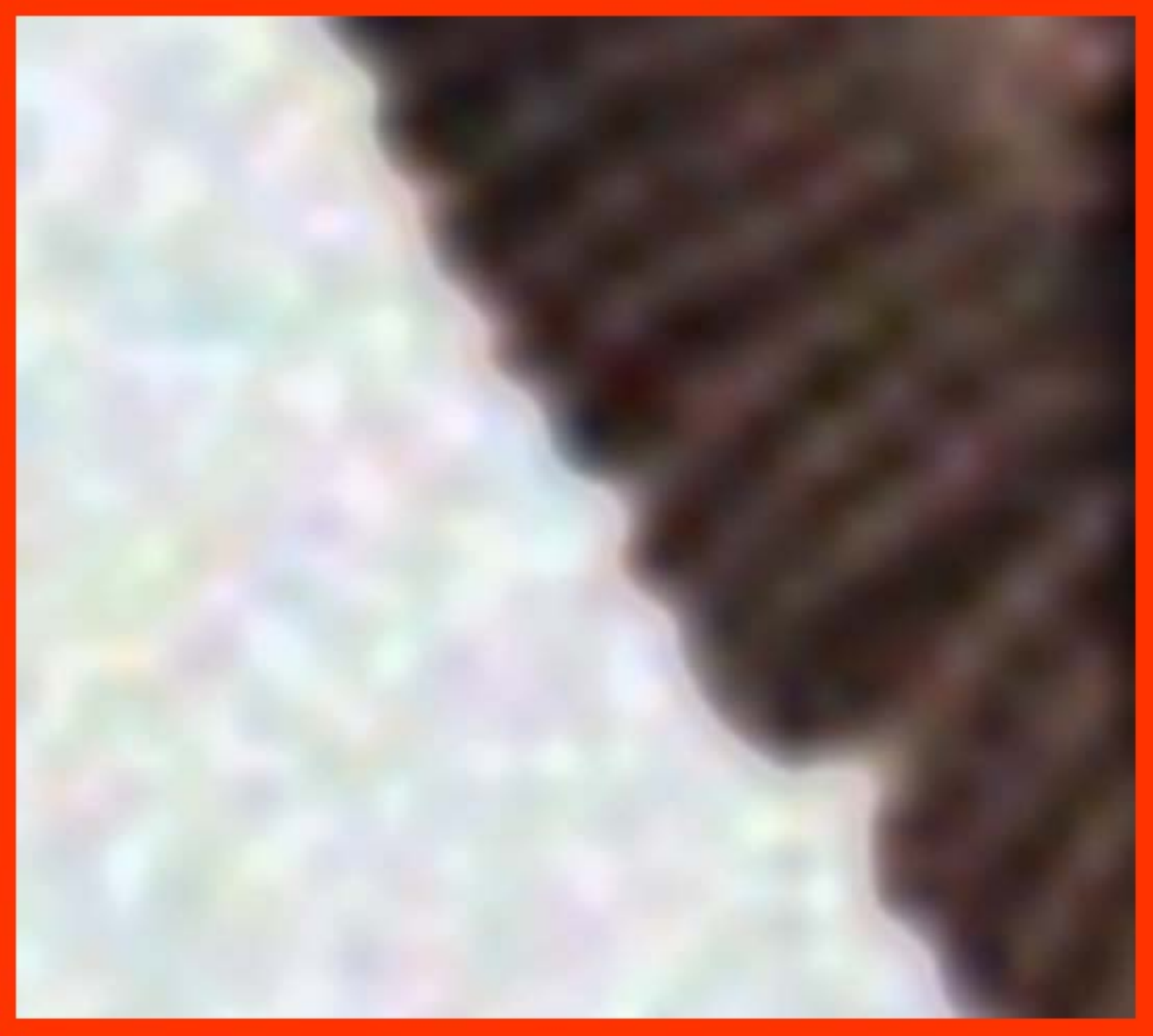} &
		\includegraphics[width=\swfive]{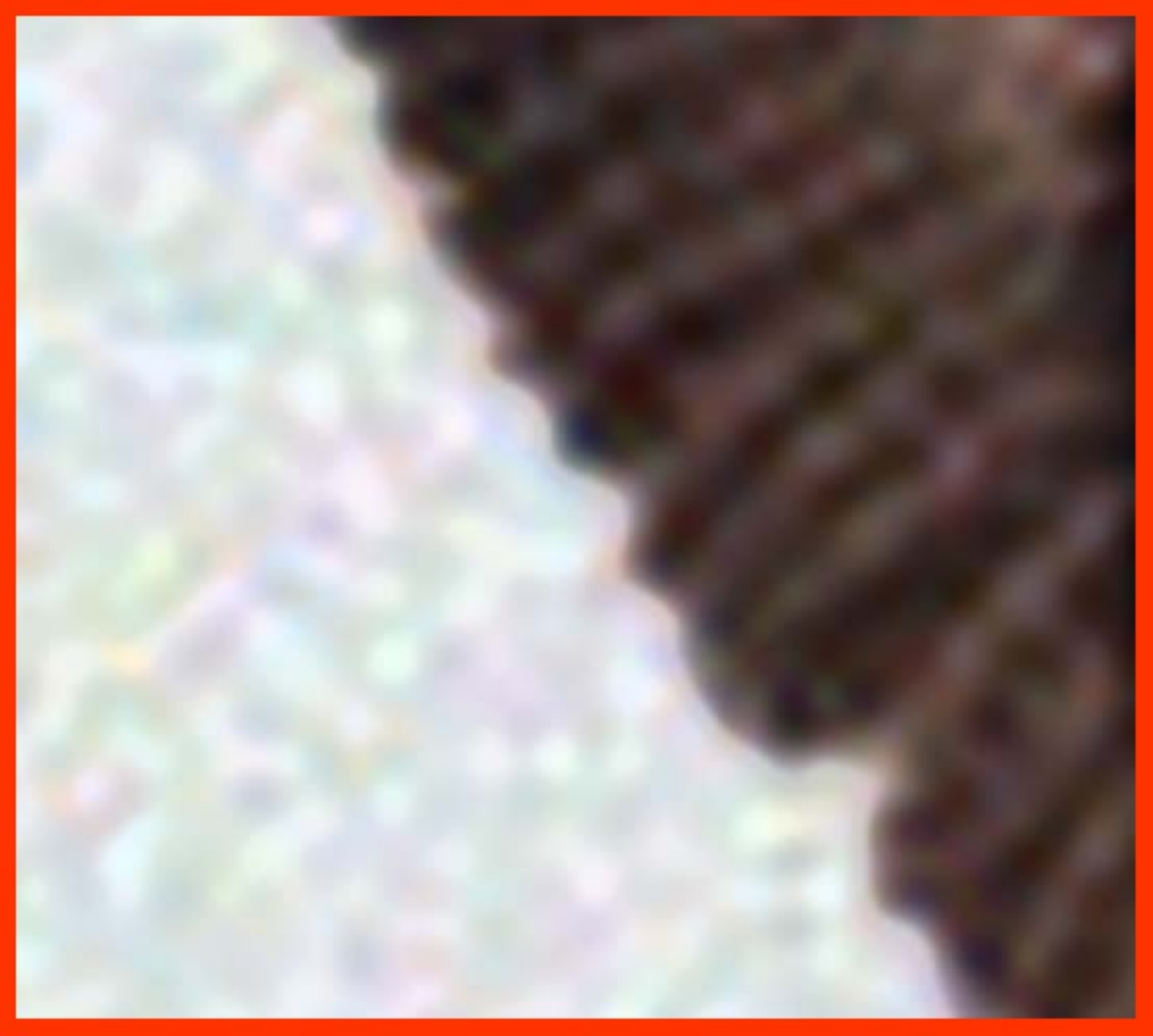} &		
		\includegraphics[width=\swfive]{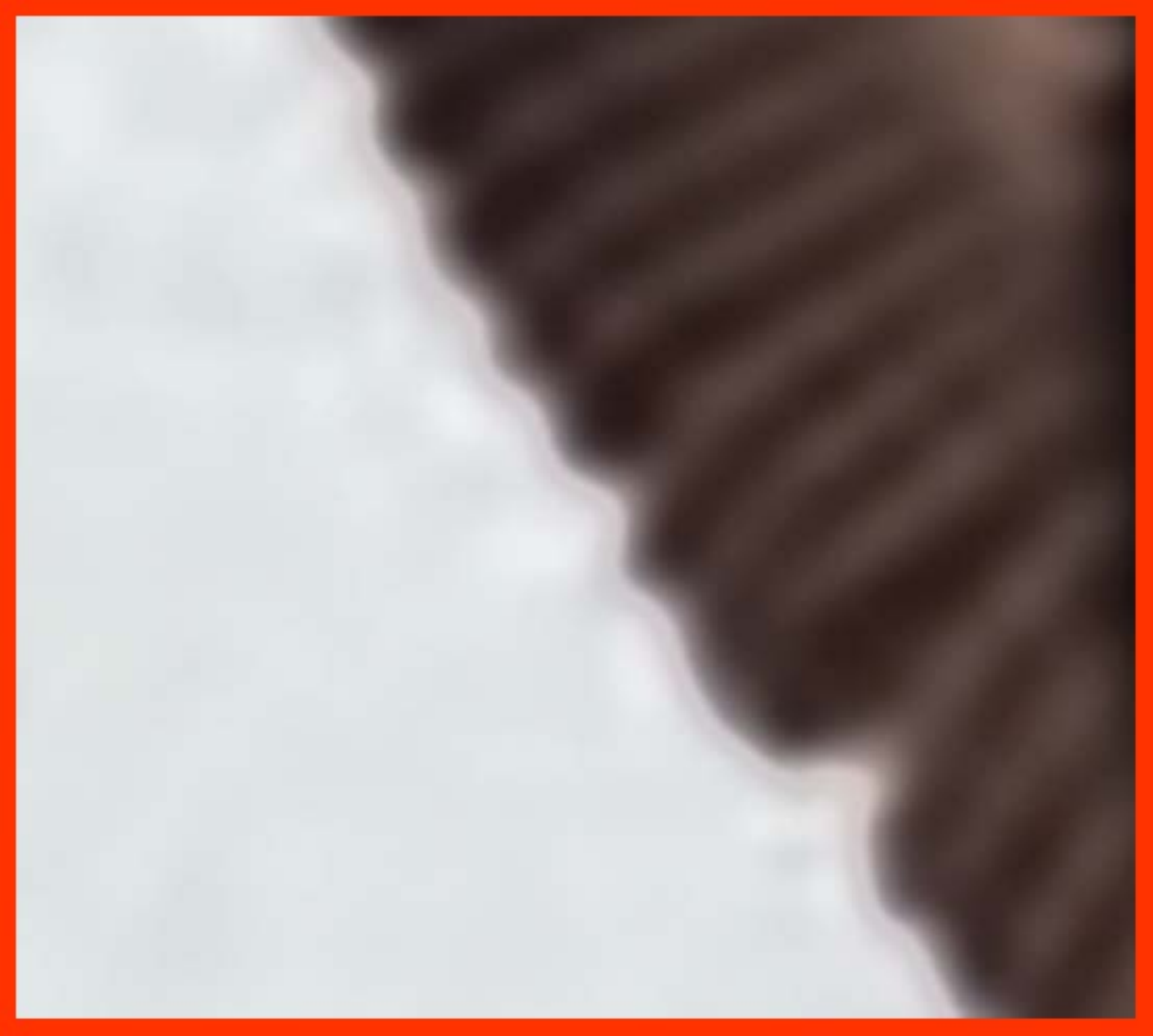} &
		\includegraphics[width=\swfive]{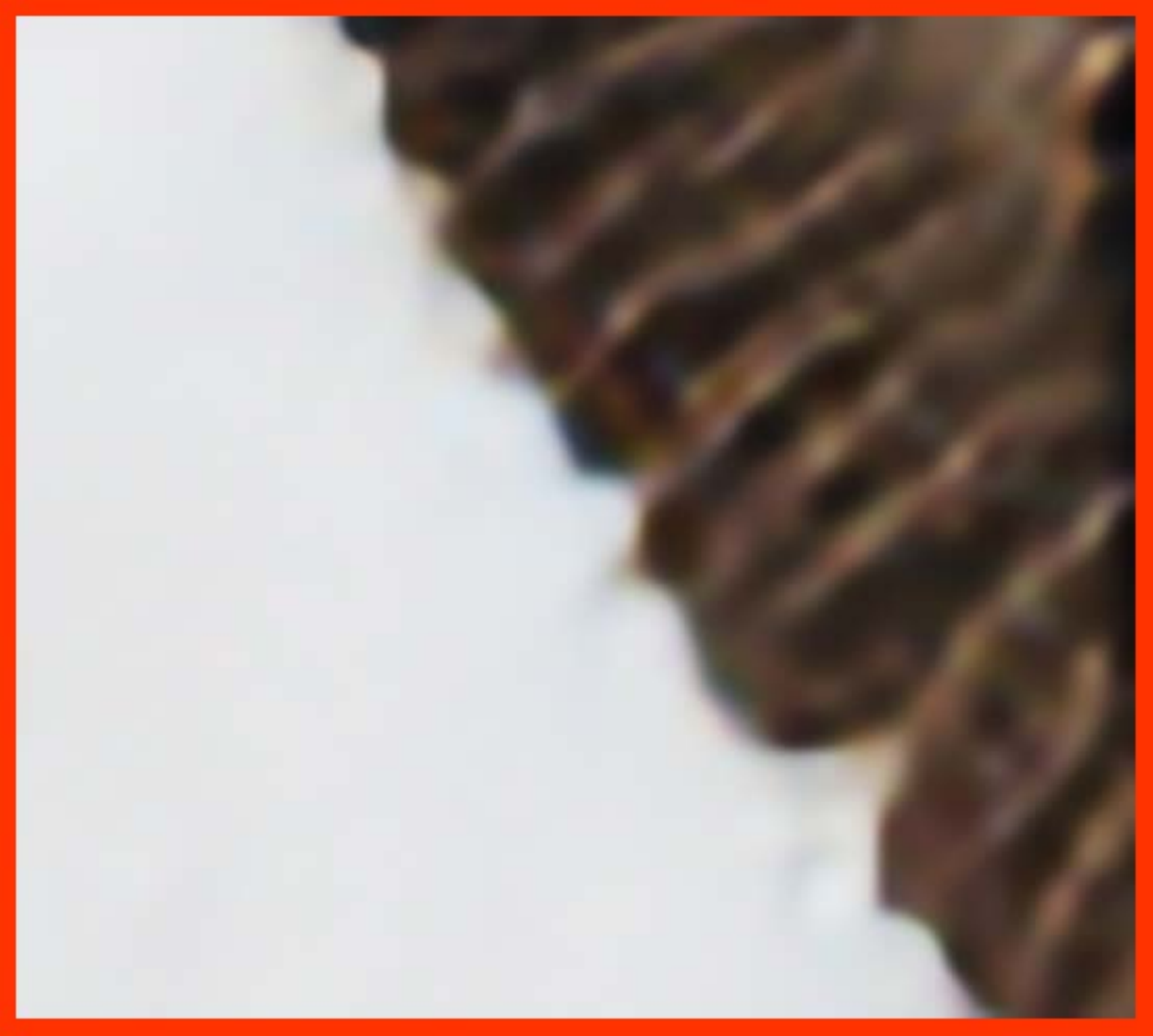} \\
		{Ground Truth} & {Bicubic} & {EDSR~\cite{lim2017enhanced}} & {BM3D+EDSR } & {\textbf{CinCGAN}} \\
		{PSNR/SSIM}  & {29.42/0.82}  & {28.95/0.76} & {30.94/0.91} & {31.01/0.92}  \\		
		
		\end{tabular}
	\end{center}
	\caption{$\times 4$ Super-resolution results of the proposed CinCGAN method for ``0896'' (DIV2K). For comparison, the sub-figures are cropped from results of existing algorithms. When the input is noisy, the results of bicubic interpolation and the EDSR~\cite{lim2017enhanced} model both are in low quality, while CinCGAN learns to reconstruct clean result with fine details. The BM3D+EDSR method means using BM3D for denoising first and then using EDSR for super-resolution.}
	\label{fig:res96}
\end{figure}

Under the above circumstances, models learned from synthetic data tend to generate similar results as traditional methods~\cite{yang2010image, kim2010single} or even simple interpolation.
In Fig.~\ref{fig:res96}, we show the results of bicubic interpolation and the state-of-the-art deep learning model---EDSR~\cite{lim2017enhanced} with a noisy input.
This is mainly due to the data bias between training and testing images. Detailed survey and analysis of deep learning based methods on real data can be found in ~\cite{kohler2017benchmarking}. 

As an alternative choice, blind SR~\cite{michaeli2013nonparametric, wang2005patch, he2009soft}  deal with the real-world data by estimating the down-sampling kernel from internal or external similar patches. However, when the input is noisy, the down-sampling kernel cannot be accurately estimated, and the inverse mapping results are accompanied by amplified noises. There are also works attempting at restoring LR images with addictive Gaussian noises~\cite{zhang2017learning}. But real-world noises may neither be addictive nor follow the standard Gaussian distribution, causing noise estimation infeasible. More generally, LR images may suffer from complex noises, blurry and non-uniform down-sampling kernels, which fail almost all existing blind SR methods. 

Inspired by the development of unsupervised learning in image-to-image translation, such as CycleGAN~\cite{zhu2017unpaired} or WESPE~\cite{ignatov2017wespe}, we intend to investigate unsupervised strategies to overcome this obstacle. In CycleGAN, images are translated between different domains with unpaired training data. They assume that the input image is of the same size as the output image, with only the difference on styles. However, in SR, output images are several times larger than the inputs, making the direct application of CycleGAN impossible. Further, using a bicubic-upsampled image as the input also could not obtain satisfactory results. SR problem is specific as it requires high quality output but not just a different style. 

After exploring several training strategies, we find an effective Cycle-in-Cycle structure, named CinCGAN, which could achieve superior results. The whole pipeline consists of two CycleGANs, while the second GAN covers the first one (See Fig.~\ref{fig:struc}). The first CycleGAN maps the LR image to the clean and bicubic-downsampled LR space. This module ensures that the LR input is fairly denoised/deblurred. We then stack another well-trained deep model with bicubic-downsampling assumption to up-sample the intermediate result to the desired size. Finally, we fine-tune the whole network using adversarial learning in an end-to-end manner. We conduct experiments on the \textbf{NTIRE2018 Super-Resolution Challenge}\footnote{https://competitions.codalab.org/competitions/18024} dataset, and show that the proposed Cycle-in-Cycle structure is much stable at training and achieves competitive performance as supervised deep learning methods. 

The contributions of this work are three-folds: 1) We study a more general super-resolution problem, where the high-resolution ground truth, down-sampling kernel and degradation function are unavailable. 2) We explore several unsupervised training strategies under the above assumption, and show that super-resolution task is different from conventional image-to-image translation. 3) We propose a Cycle-in-Cycle structure that could achieve comparable results as supervised CNN networks.

\label{sec:intro}

\section{Related work}
\subsection{Image Super-Resolution}
Single image super-resolution (SISR) has been widely studied for decades. 
Early approaches either rely on natural image statistics~\cite{zhang2010non}\cite{kim2010single} or pre-defined models~\cite{irani1991improving}\cite{fattal2007image}\cite{sun2008image}. Later, mapping functions between LR images and HR images are investigated, such as sparse coding based SR methods~\cite{yang2010image}\cite{zeyde2010single}.

Recently, deep convolution neural networks (CNN) have shown explosive popularity and powerful capability to improve the quality of SR results. 
Ever since Dong~\cite{dong2016image} first proposed using CNN for SR and achieved the state-of-the-art performance, plenty of CNN architectures have been studied for SISR. Inspired by the VGG~\cite{simonyan2014very} networks used for ImageNet classification, Kim et al.~\cite{kim2016accurate} present a very deep network (VDSR) that learns a residual image. For accelerating the speed of SR, FSRCNN~\cite{dong2016accelerating} and ESPCN~\cite{shi2016real} extract feature maps at the low-resolution space and up-sample the image at the last layer by transposed convolution and sub-pixel convolution, respectively.
All the above mentioned CNN based SR methods aim at minimizing the mean-square error (MSE) between the reconstructed HR image and the ground truth. 
Based on the observation that minimizing MSE will make the SR results overly smooth, SRGAN~\cite{ledig2016photo} combines an adversarial loss~\cite{goodfellow2014generative} and a perceptual loss~\cite{simonyan2014very}\cite{johnson2016perceptual} as the final objective function, and generates visually pleasing images which contain more high frequency details than the MSE-loss based methods. The champion of NTIRE2017 Super-Resolution Challenge~\cite{timofte2017ntire}, EDSR~\cite{lim2017enhanced}, employs deeper and wider networks to achieve the state-of-the-art performance by removing the unnecessary modules in SRResNet~\cite{ledig2016photo}.    

\subsection{Blind Image Super-Resolution}
Although a lot of works focus on SR problems with known degradation/downsamping kernels, little works try to solve blind SR---the degradation operation from HR images to LR images are unavailable.
Estimating the degradation/blur kernel is an essential step for blind SR.
Wang et al. \cite{wang2005patch} propose a probabilistic framework combined with the image co-occurrence prior to estimate the unknown point spread function (PSF) parameters.
According to the property that small image patches will re-appear in natural images, Michaeli and Irani~\cite{michaeli2013nonparametric} present a method that is able to estimate the optimal blur kernel. Another relevant work~\cite{shao2015simple} introduces a convolution consistency constraint and \textit{bi-l$_0$-l$_2$}-norm regularization~\cite{shao2015bi} to guide the blur kernel estimation process, achieving state-of-the-art blind SR performance.

In this work, we investigate how deep learning can be beneficial for addressing blind SR problems.

\begin{figure*}[htb]
\centering
\centerline{\includegraphics[width=0.98\linewidth]{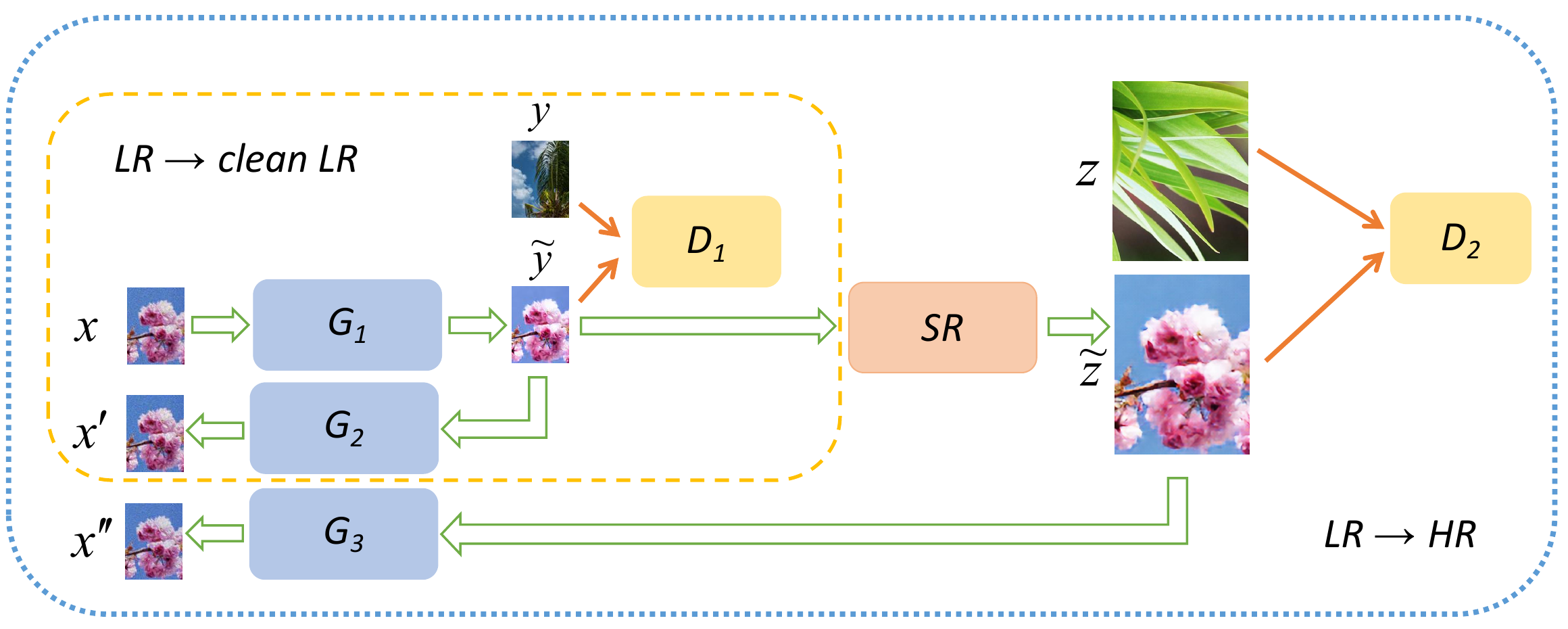}}
\caption{The framework of the proposed CinCGAN, where $G_1$, $G_2$ and $G_3$ are generators and \textit{SR} is a super-resolution network. $D_1$ and $D_2$ are discriminators. 
The $G_1$, $G_2$ and $D_1$ compose the first \textit{LR$\to$clean LR} CycleGAN model, mapping the degrade LR images to clean LR images. 
The $G_1$, \textit{SR}, $G_3$ and $D_2$ compose the second \textit{LR$\to$HR} CycleGAN model, mapping the LR images to HR images.}
\label{fig:struc}
\end{figure*} 

\subsection{Unsupervised Learning}

Existing supervised deep learning methods cannot handle blind SR without LR-HR image pairs.
In real-world scenarios, where paired data is unavailable, it is essential to find a way to realize unsupervised learning.
Recent work on GAN~\cite{goodfellow2014generative} provides a feasible solution, which includes a generator and a discriminator. The generator tries to generate fake images to fool the discriminator, while the discriminator aims at distinguishing the generated results from real data. GAN is widely used to solve the unsupervised learning problems.
DualGAN~\cite{yi2017dualgan} and CycleGAN~\cite{zhu2017unpaired} are two works about image-to-image translation using unsupervised learning, and both of them present an interesting network structure that contains a pair of forward and inverse  generators. The forward generator maps domain X to domain Y, while the inverse generator maps the output back to domain X to maintain cycle consistency. 
Ignatov et al. \cite{ignatov2017wespe} use the similar architecture to design a weakly supervised photo enhancer (WESPE) that translates ordinary photos to DSLR-quality images.

Different from the proposed method, both DualGAN~\cite{yi2017dualgan} and CycleGAN~\cite{zhu2017unpaired} deal with input and output images of the same size, while SR requires the output images several times larger than the inputs.
Utilizing the property of cycle consistency, we present a Cycle-in-Cycle GAN (CinCGAN) to super-resolve the LR images of which the degradation operators are unknown.  
Our method achieves a comparable performance with the state-of-the-art \textit{supervised} CNN based algorithms~\cite{dong2016accelerating,ledig2016photo,lim2017enhanced}.
\label{sec:related}

\section{Proposed Method}

{\flushleft \bf{Problem formulation}}
The conventional formulation of SISR~\cite{yang2010image} is $x = SHz+n$, where $x$ and $z$ denote LR and HR image respectively, $SH$ represents the down-sampling and blurring matrix, and $n$ is the addictive noise. 
Blind SR~\cite{michaeli2013nonparametric, wang2005patch} follow the same assumption, only with unknown $SH$. In this work, we study a more general formulation as $x = f_n(f_d(z))+n$, where $f_d$ is the down-sampling process, $f_n$ is a degradation function that may introduce complex noises, shift and blur. 
Here, we assume that $f_d$, $f_n$ and the paired HR-LR training data are unavailable. Nevertheless, we can obtain a set of LR images that can be used for analysis and unsupervised training. 

{\flushleft \bf{Motivation}}
1) Why applying unsupervised training? As the down-sampling and degradation functions are complex and coupled, it is hard to perform accurate estimation like traditional blind SR methods~\cite{michaeli2013nonparametric, wang2005patch}. The unavailability of HR images in practise also makes supervised training with simulated paired data impractical. This drives us to explore unsupervised learning strategies. 2) What is the difference between SR and image-to-image translation? SR accepts an LR image and outputs a HR image with much larger resolution. Further, SR requires the output to be of high quality, not just a different style. If we directly apply the image-to-image translation methods, we need to up-sample the LR image first by interpolation, which will also enlarge the noisy patterns. Directly applying existing methods like CycleGAN cannot remove such amplified noises, and training becomes very unstable. 
Experiments (in Sec.~\ref{sec:ablation}) also show that when the degradation function varies from image to image, it is difficult to deal with all kinds of images in a single forward pass.

{\flushleft \bf{Solution pipeline}}
Our solution pipeline consists of three steps. First, we learn a mapping from an LR image set $X$ to a ``clean'' LR image set $Y$, where images are noise-free and down-sampled from HR images $Z$ with bicubic kernel. In other words, we deblur and denoise the input images at low resolution. Second, we adopt an existing SR model to super-resolve the intermediate results to the desired resolution. In the end, we combine and fine-tune these two models simultaneously to get the final HR images. 

Under the guidance of the above pipeline, we propose a Cycle-in-Cycle structure named CinCGAN as shown in Fig.~\ref{fig:struc}. To be specific, we adopt two coupled CycleGANs to learn the mapping from $X$ to $Y$ and $Y$ to $Z$, respectively. Unpaired images $x_i \in X$, $y_j \in Y$ and $z_j \in Z$ are used for training\footnote{For simplicity, we omit the subscript $i$ and $j$ in the following.}, where $y_j$ is down-sampled from $z_j$ with bicubic kernel. Details are given in the following.

\subsection{LR Image Restoration} 

The framework of the first CycleGAN that maps an LR image $x$ to a clean LR image $y$ is shown as \textit{LR$\to$clean LR} in Fig.~\ref{fig:struc}. 
Given an input image $x$, the generator $G_1$ learns to generate an image $\tilde{y}$ that looks similar to the clean LR $y$, so as to fool the discriminator $D_1$. Meanwhile, $D_1$ learns to distinguish the generated sample $G_1(x)$ from the real sample $y$. To stabilize the training procedure, we use the least square loss~\cite{mao2016multi} instead of the negative log-likelihood used in \cite{goodfellow2014generative}. The generator-adversarial loss is:
\begin{eqnarray}
\mathcal{L}^{LR}_{GAN} = \frac{1}{N}\sum_i^N ||D_1(G_1(x_i)) - 1||_2, 
\label{eq:ganL1}
\end{eqnarray}
where $N$ is the number of training samples.
To maintain consistency between input $x$ and output $y$, we add a network $G_2$ and let $x' = G_2(G_1(x))$ be identical to the input $x$. Hence, we also use a cycle consistency loss as:
\begin{eqnarray}
\mathcal{L}^{LR}_{cyc} =\frac{1}{N} \sum_i^N || G_2(G_1(x_i)) - x_i||_2.
\end{eqnarray}

In the previous work~\cite{zhu2017unpaired}, the authors introduce an identity loss to preserve color composition between input and output images when they work on painting generation. They claim that the identity loss can help preserve the color of input images. In image SR, we also need to avoid color variation among different iterations, thus we add an identity loss
\begin{eqnarray}
\mathcal{L}^{LR}_{idt} = \frac{1}{N} \sum_i^N ||G_1(y_i) - y_i||_1.
\end{eqnarray}

In addition, we add a total variation (TV) loss to impose spatial smoothness
\begin{eqnarray}
\mathcal{L}^{LR}_{TV} = \frac{1}{N} \sum_i^N ( || \nabla_h G_1(x_i) ||_2 + ||\nabla_w G_1(x_i) ||_2 ),
\end{eqnarray}
where $\nabla_h$ and $\nabla_w$ are functions to compute the horizontal and vertical gradient of $G_1(x_i)$.

In summary, the final objective loss for the \textit{LR$\to$clean LR} model is a weighted sum of the four losses:
\begin{eqnarray}
\mathcal{L}^{LR}_{total} = \mathcal{L}^{LR}_{GAN} + w_1 \mathcal{L}^{LR}_{cyc}  + w_2 \mathcal{L}^{LR}_{idt} + w_3 \mathcal{L}^{LR}_{TV}
\label{eq:objLR}
\end{eqnarray}
where $w_1,w_2,w_3$ are the weights of different losses. 

\begin{figure*}
	\begin{center}
		\begin{tabular}{c}
			\includegraphics[width=0.85\linewidth]{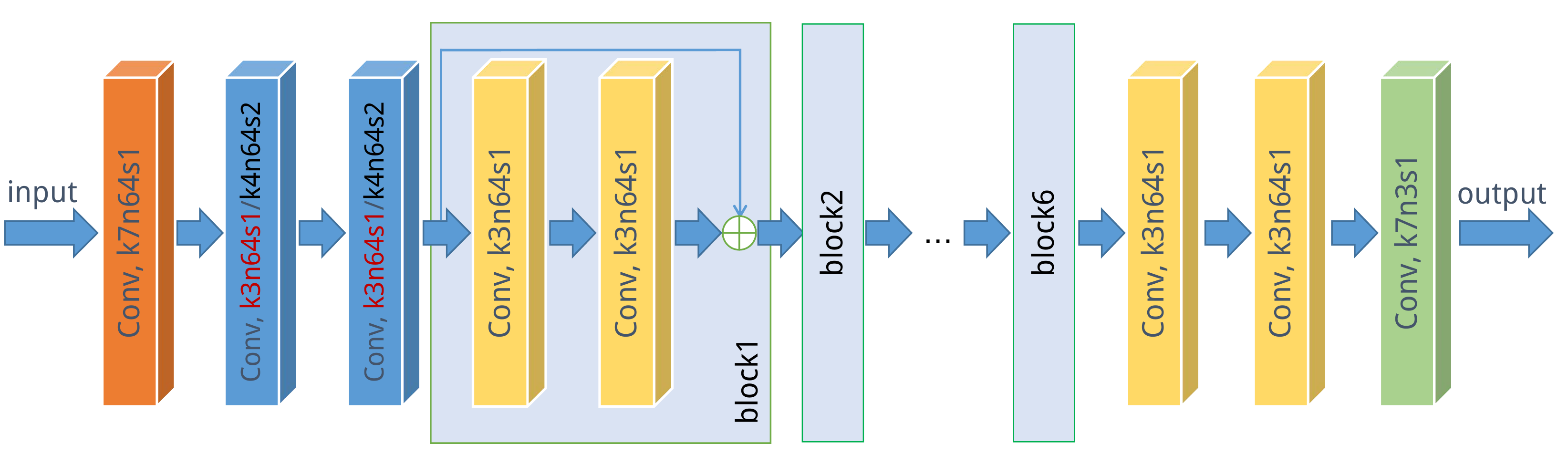} \\
			(a) Generator \\
			\includegraphics[width=0.8\linewidth]{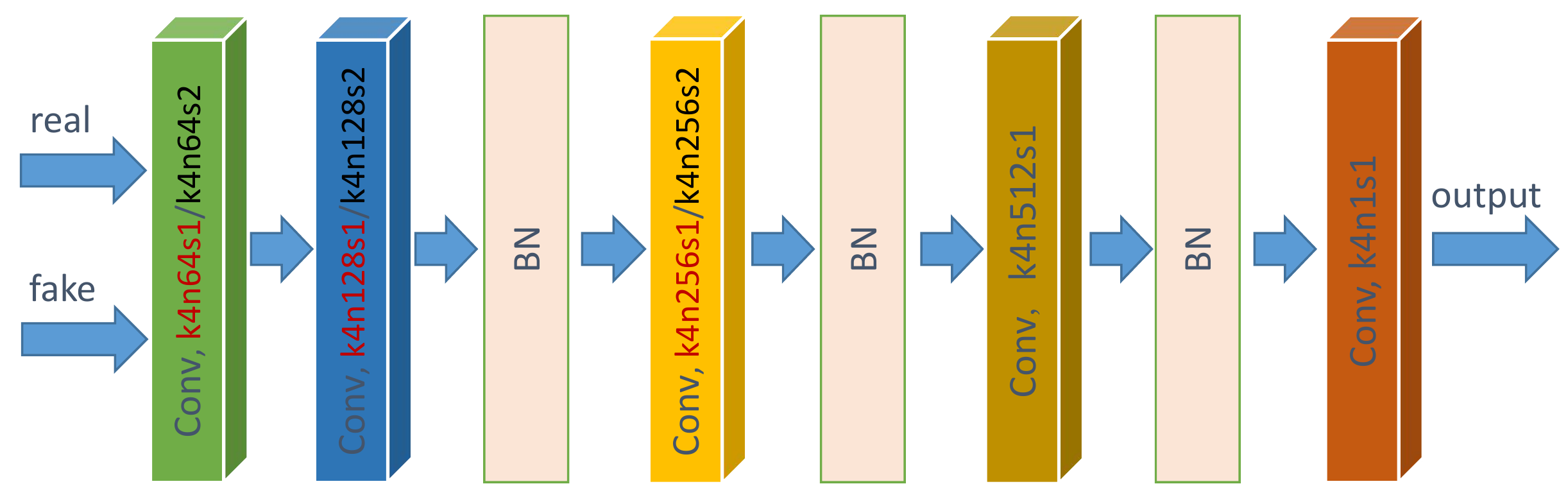} \\
			(b) Discriminator
		\end{tabular}
	\end{center}
	\caption{The generators $G_1$, $G_2$ and $G_3$ share the same framework as (a) and the discriminators $D_1$ and $D_2$ share the same framework as (b). 
For the $2$-nd and $3$-rd convolution layers in generator (a), \textbf{k3n64s1} is for $G_1$ and $G_2$, while \textbf{k4n64s2} is for $G_3$. For the first three convolution layers in discriminator (b), \textbf{k4n64s1}, \textbf{k4n128s1}, and \textbf{k4n256s1} are for $D_1$ and \textbf{k4n64s2}, \textbf{k4n128s2}, and \textbf{k4n256s2} are for $D_2$. Please see text for details.}
\label{fig:arch}
\end{figure*}

\subsection{Jointly Restoration and Super-Resolution}
 
We then investigate how to super-resolve the intermediate image $\tilde{y}$ to the desired size. Recently, the enhanced deep residual network -- EDSR~\cite{lim2017enhanced} has won the first prize in the NTIRE 2017 challenge on single image super-resolution~\cite{agustsson2017ntire}. 
For simplicity, we directly adopt EDSR as the $SR$ network stacked after $G_1$. Similarly, we use a discriminator $D_2$ for adversarial training both $G_1$ and $SR$ networks. We also utilize another generator $G_3$ to ensure cycle consistency between $x$ and the reconstructed $x''$. The GAN loss, cycle loss and TV loss for the  \textit{LR$\to$HR} network are formulated as follows:
\begin{eqnarray}
\mathcal{L}^{HR}_{GAN} =\frac{1}{N} \sum_i^N ||D_2(SR(G_1(x_i))) - 1||_2,
\label{eq:ganHR}
\end{eqnarray}
\begin{eqnarray}
\mathcal{L}^{HR}_{cyc} =\frac{1}{N} \sum_i^N || G_3(SR(G_1(x_i))) - x_i||_2,
\label{eq:cycHR}
\end{eqnarray}

\begin{small}
\begin{align}
\mathcal{L}^{HR}_{TV} =\frac{1}{N} \sum_i^N (|| \nabla_h SR(G_1(x_i)) ||_2 + || \nabla_w SR(G_1(x_i)) ||_2).
\label{eq:tvHR}
\end{align}
\end{small} 
\noindent

For the identity loss, instead of maintaining the tint consistency between input and output, we consider ensuring the $SR$ network can generate adequate quality of super-resolved images. We define a new identity loss as:
\begin{eqnarray}
\mathcal{L}^{HR}_{idt} = \sum_i ||SR(z') - z||_2.
\label{eq:idtHR}
\end{eqnarray}
where $z'$ is down-sampled from $z$ with bicubic kernel. This $\mathcal{L}^{HR}_{idt}$ makes the SR network does not betray its original ambition, such that the produced $\tilde{z}$ can be reasonable SR results.

To sum up, the total loss for fine-tuning the LR to HR networks is
\begin{eqnarray}
\mathcal{L}^{HR}_{total} = \mathcal{L}^{HR}_{GAN} + \lambda_1 \mathcal{L}^{HR}_{cyc}  + \lambda_2 \mathcal{L}^{HR}_{idt} + \lambda_3 \mathcal{L}^{HR}_{TV}
\label{eq:objHR}
\end{eqnarray}
where $\lambda_1,\lambda_2,\lambda_3$, for $i = {1,2,3}$, are weights of each loss.

\subsection{Network Architecture}

The architecture of generators $G_1, G_2, G_3$ and discriminators $D_1, D_2$  are shown in Fig.~\ref{fig:arch}. 
We adapt similar architecture as the work of Zhu et al.~\cite{zhu2017unpaired}, which has shown impressive results for unpaired image-to-image translation.
Here, ``conv'' means convolution layer, where a Leaky ReLU layer with negative slope 0.2 is added right after except for the last convolution layer (we omit it for simplicity). ``BN'' means a batch normalization layer.
The number after symbols $k,n$ and $s$ represents kernel size, number of filters and stride size, respectively. For example, \textbf{k3n64s1} refers to the convolution layer that contains 64 filters, of which the spatial size is 3 and stride is 1. 

For the generators $G_1$ and $G_2$, we use 3 convolution layers at the head and tail, and 6 residual blocks in the middle. 
The generator $G_3$ shares the same architecture as $G_1$ and $G_2$, except for the $2$-nd and $3$-rd convolution layers, where the stride is set to 2 to perform down-sampling. 
As to the discriminator, we use a $70 \times 70$ PatchGAN for $D_2$. Since we up-sample LR images with a scale of $\times$4, the size of input images is usually less than 70 (we use $32 \times 32$ LR images and $128 \times 128$ HR images for training). Hence, we modify the stride of the first three convolution layers as 1 for discriminator $D_1$, such that the respective field of $D_1$ is reduced to $16 \times 16$.
\label{sec:method}

\section{Experiments}
In this section, we first introduce the dataset and details we used for training. We then evaluate the performance of the proposed CinCGAN model by comparing with several state-of-the-art SISR methods. Finally, we perform ablation study to validate the advantages of CinCGAN.

\subsection{Training data}
We take the track 2 dataset from the NTIRE2018 Super-Resolution Challenge for training. The challenge aims to restore a HR image given a degraded LR image. They provide a high-quality image dataset, DIV2K ~\cite{agustsson2017ntire}, which contains 800 training images and 100 validation images. The DIV2K dataset contains almost all kinds of natural scenarios: buildings (indoor and outdoor), forest, lakes, animals, people, etc.
The track 2 dataset is degraded from DIV2K dataset, with down-sampling, blurring, pixel shifting and noises. Although the parameters of the degradation operators are fixed for all images, the blur kernels are randomly generated and their resulting pixel shifts vary from image to image. Hence, the degradation kernels of images in the track 2 dataset are unknown and diverse. 

Since our purpose is to unsupervised train a network without paired LR-HR data, we take the first 400 images (numbered from 1 to 400) from the training LR set as input images $X$, and the other 400 images (numbered from 401 to 800) from the HR set as demanding HR images $Z$. The intermediate clean LR images $Y$ are directly bicubic down-sampled from $Z$. Similar to~\cite{dong2016accelerating}\cite{simonyan2014very}, we augment data with 90 degree rotation and flipping. Our experiments are performed with a scaling factor of $\times$4. We randomly crop $X$ and $Y$ with size $32 \times 32$ and crop $Z$ with size $128 \times 128$.
We conduct testing on the provided 100 validation images. 
Note that, although DIV2K contains paired training dataset, we do not use paired data for supervised training. 

\renewcommand{\tabcolsep}{.1pt}
\begin{figure*}
	\begin{center}
		\begin{tabular}{cccccc}
		\includegraphics[width=\swsix]{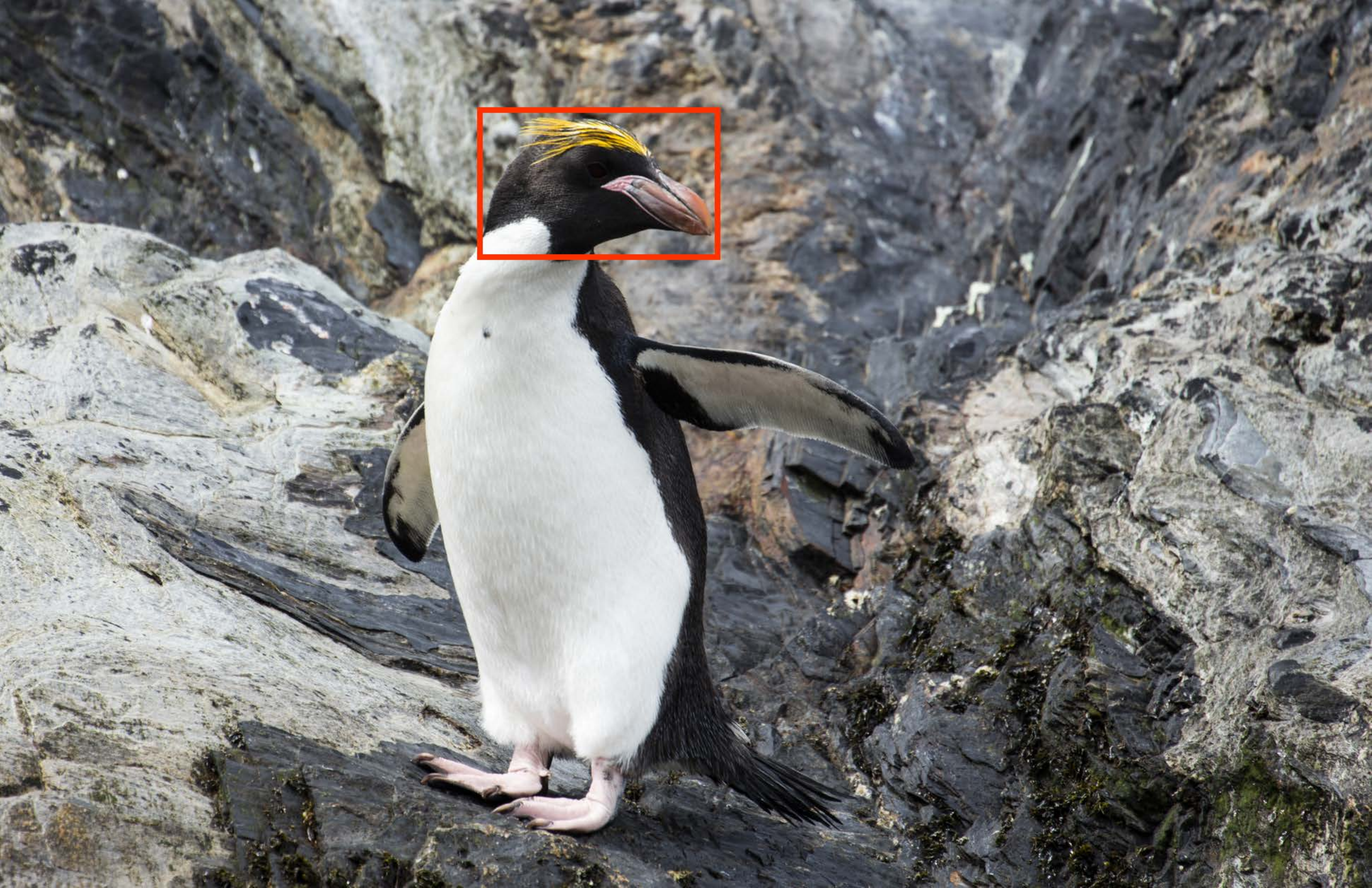} &
		\includegraphics[width=\swsix]{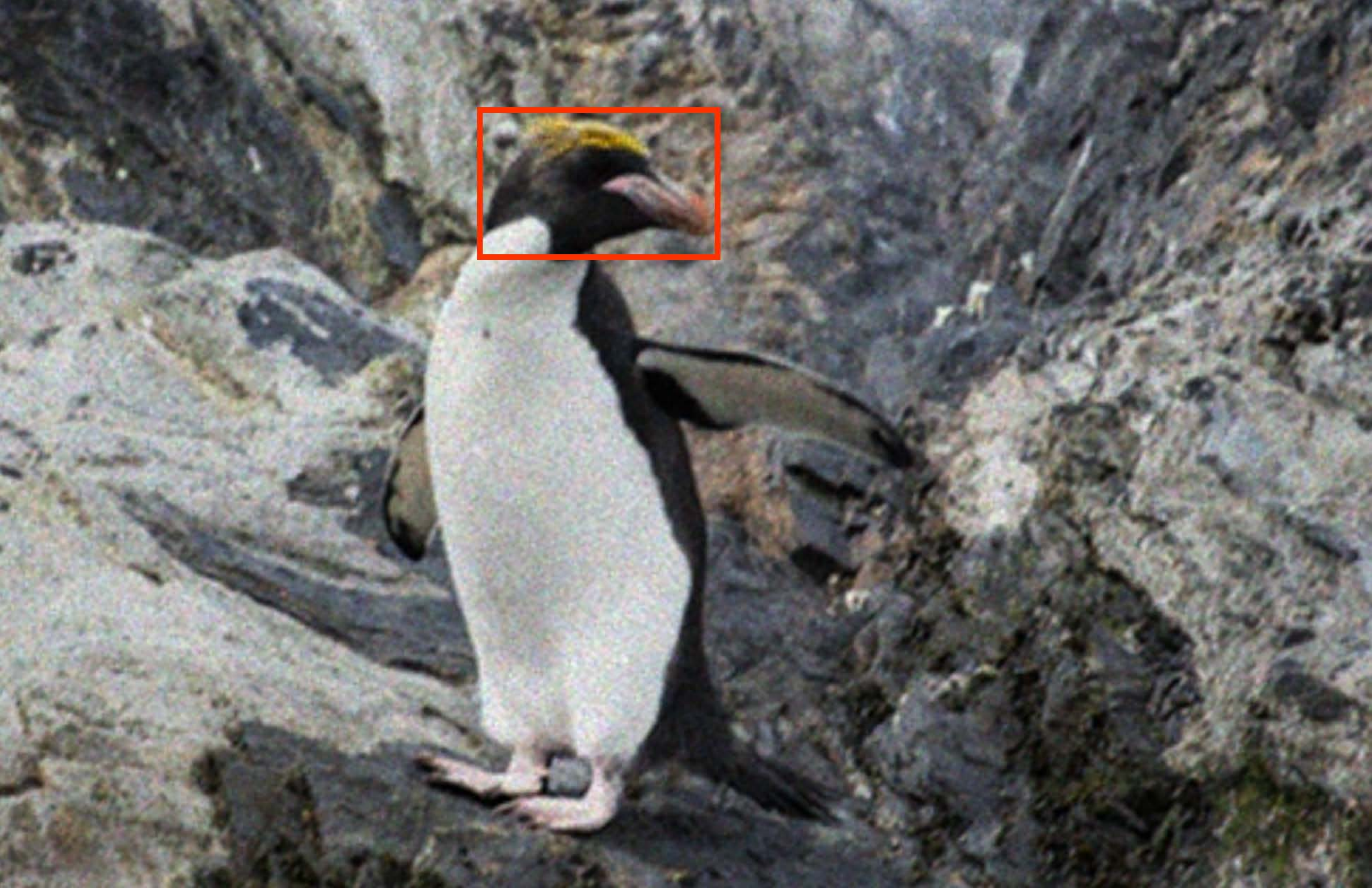} &
		\includegraphics[width=\swsix]{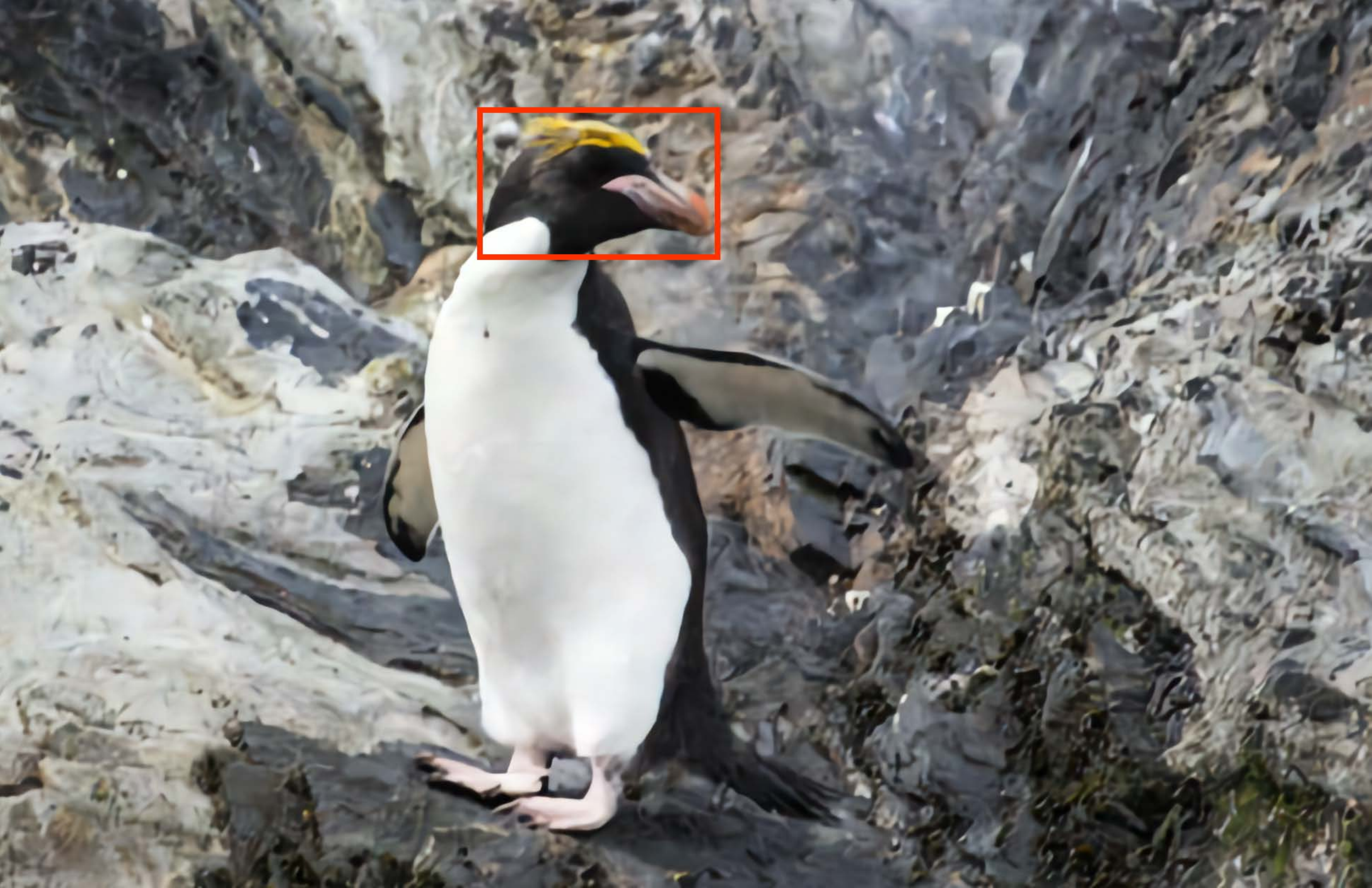} &
		\includegraphics[width=\swsix]{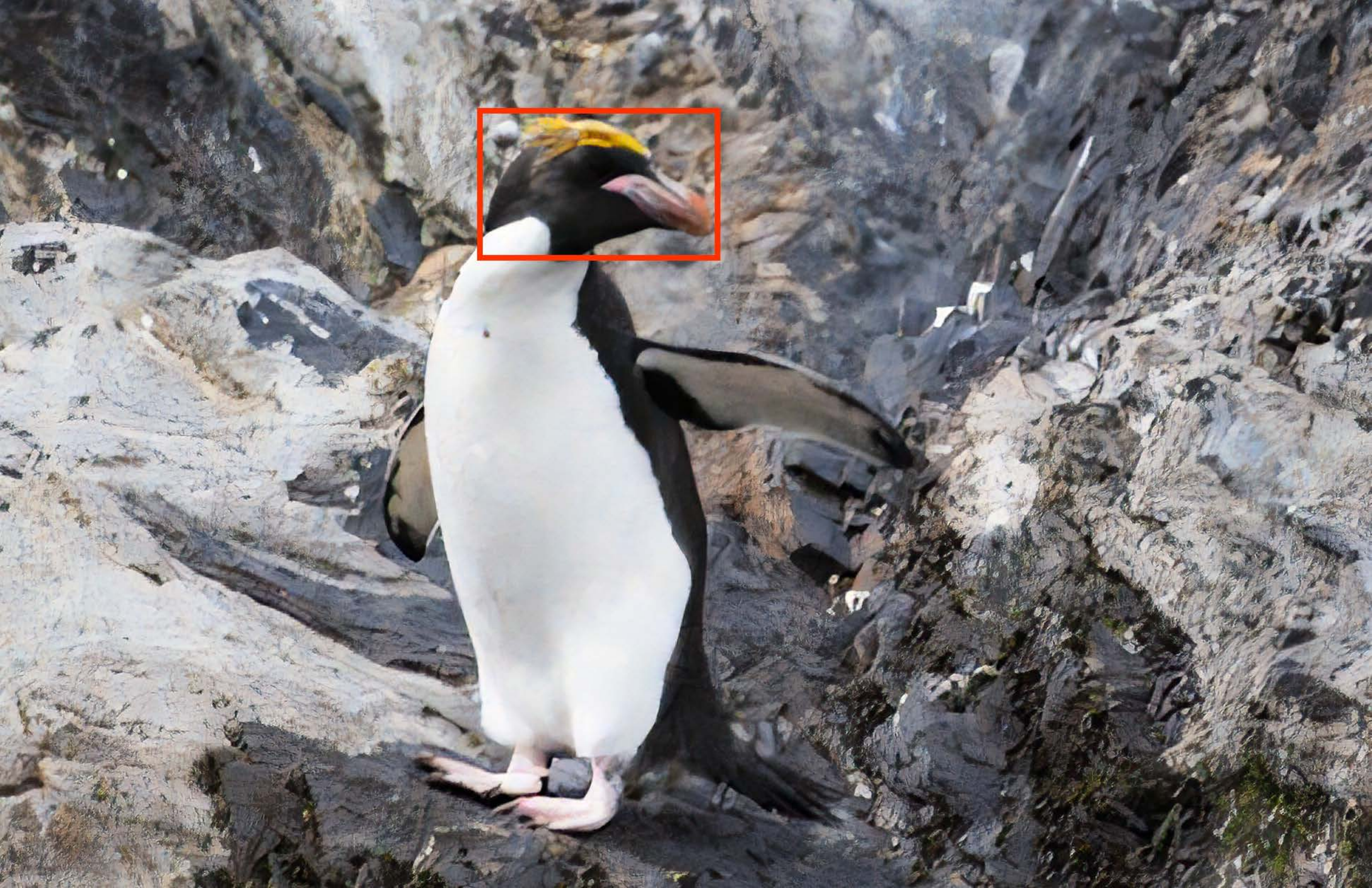} &
		\includegraphics[width=\swsix]{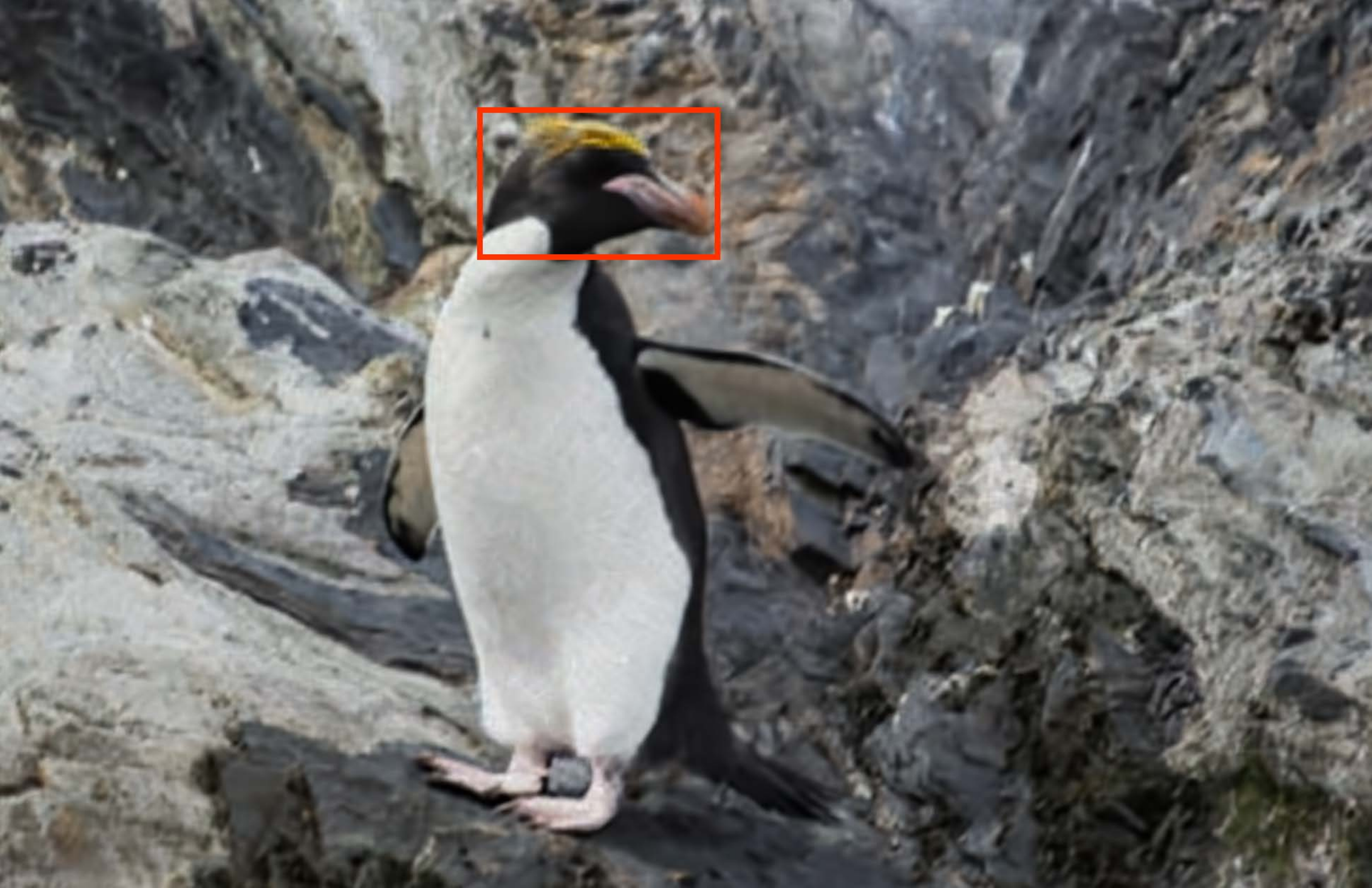} &
		\includegraphics[width=\swsix]{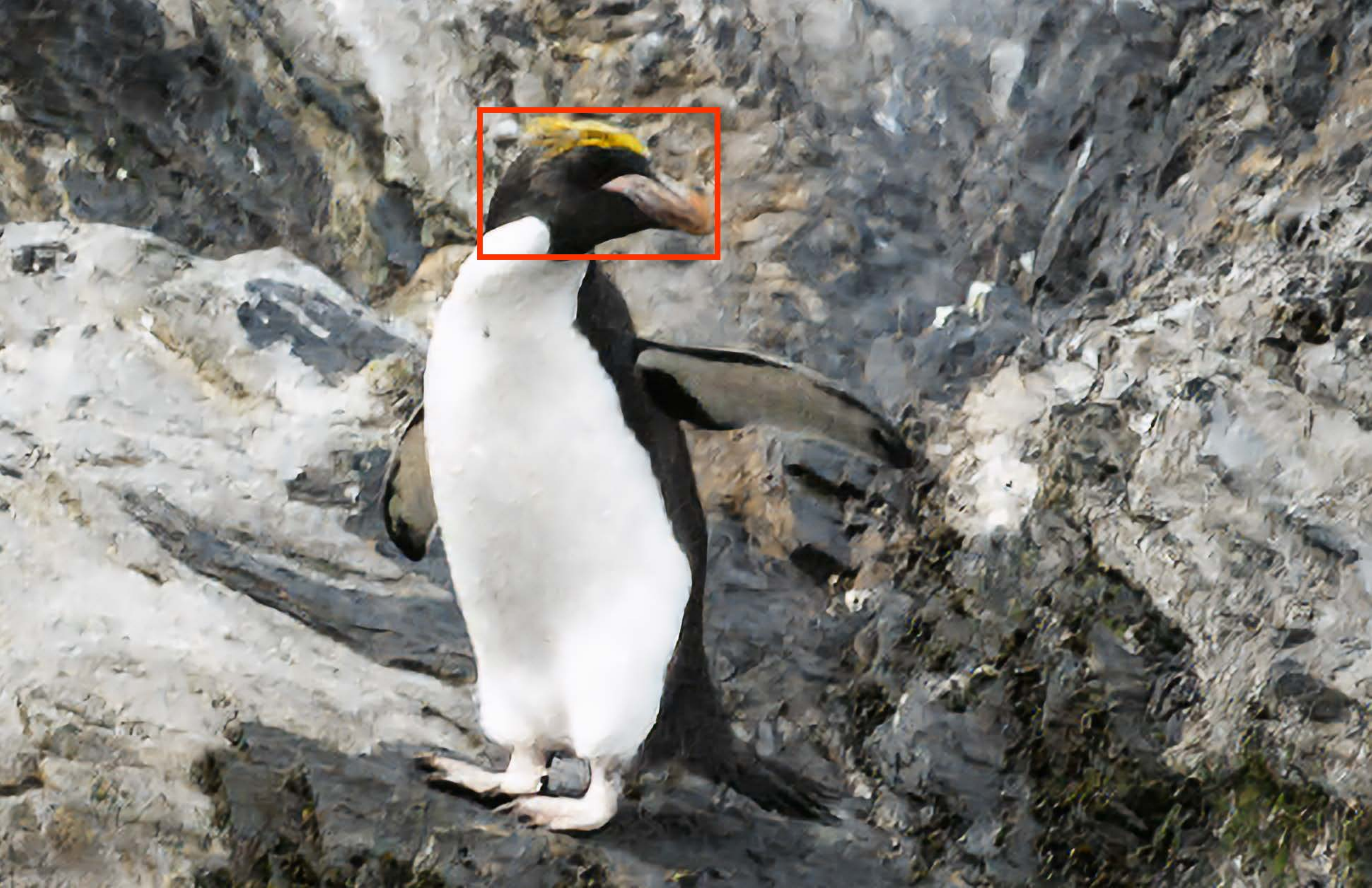} \\
			
		\includegraphics[width=\swsix]{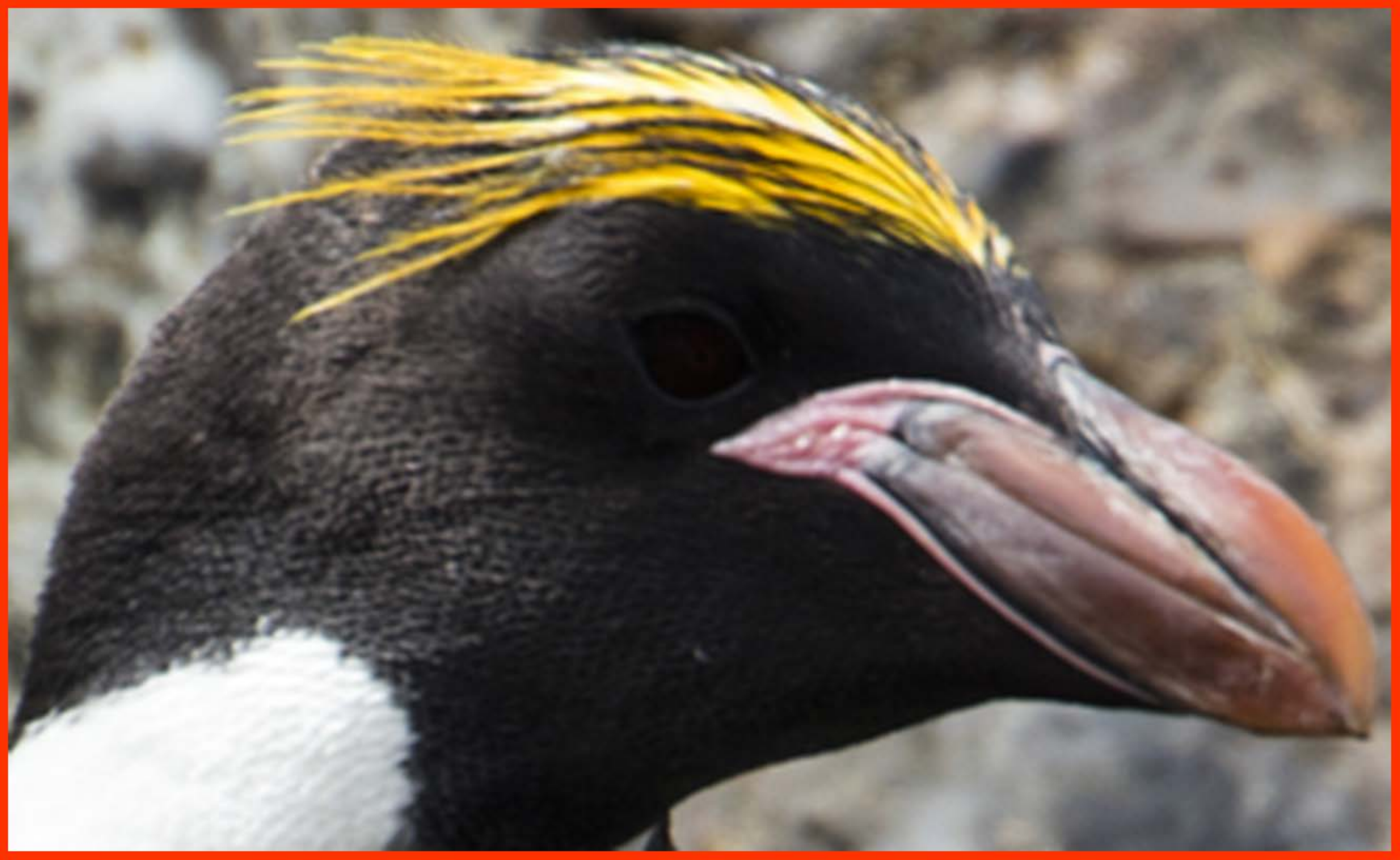} &
		\includegraphics[width=\swsix]{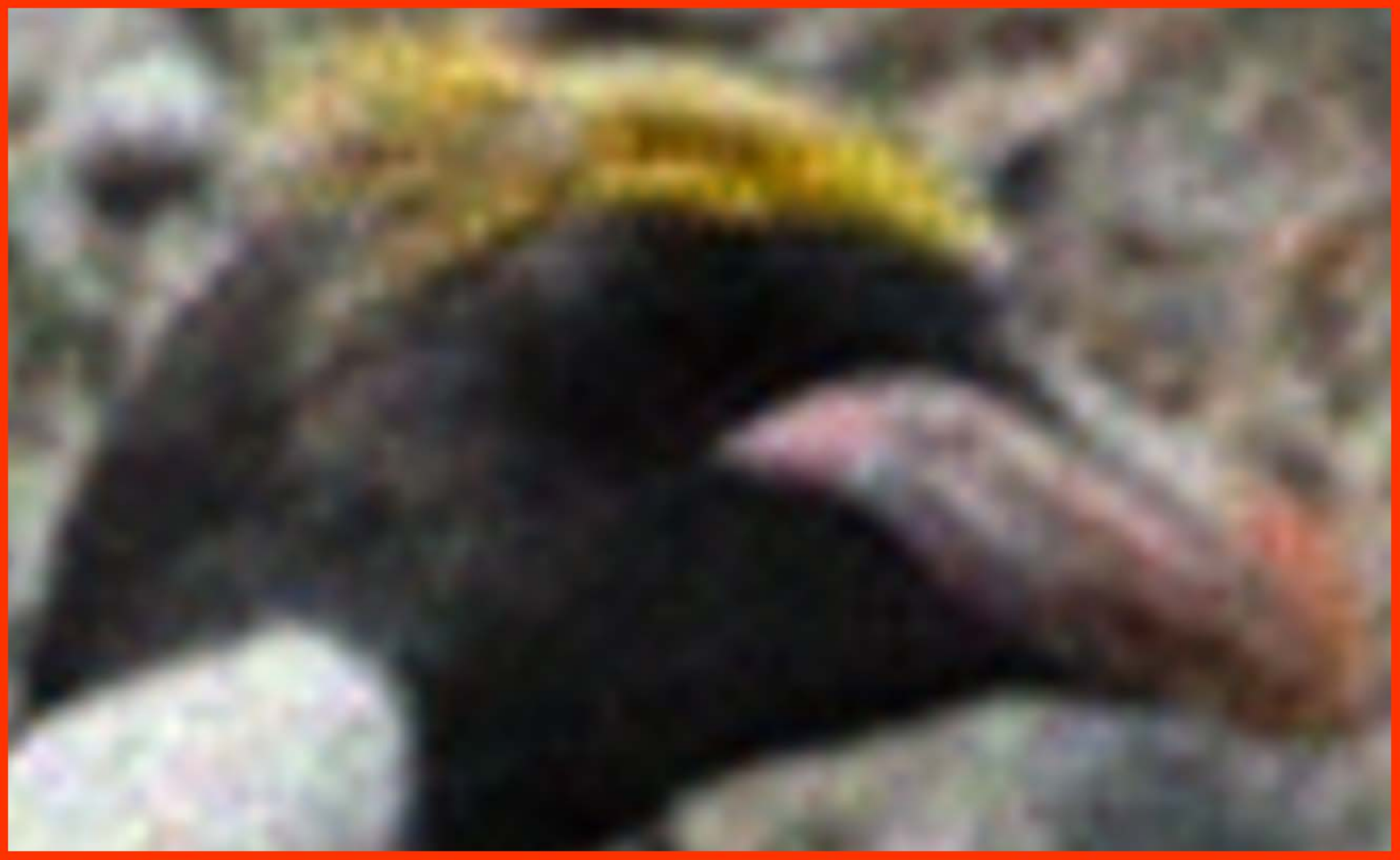} &
        \includegraphics[width=\swsix]{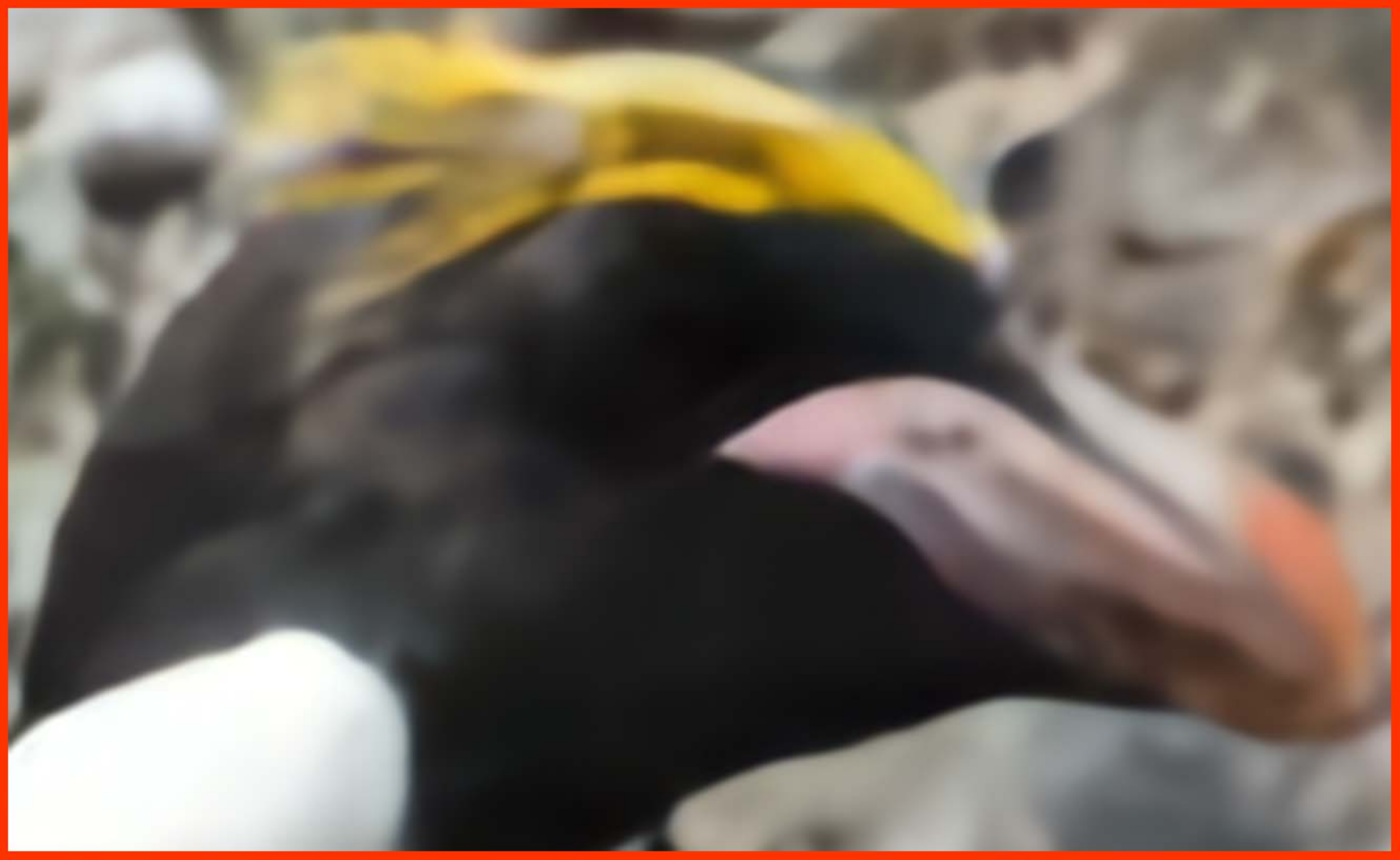} &
		\includegraphics[width=\swsix]{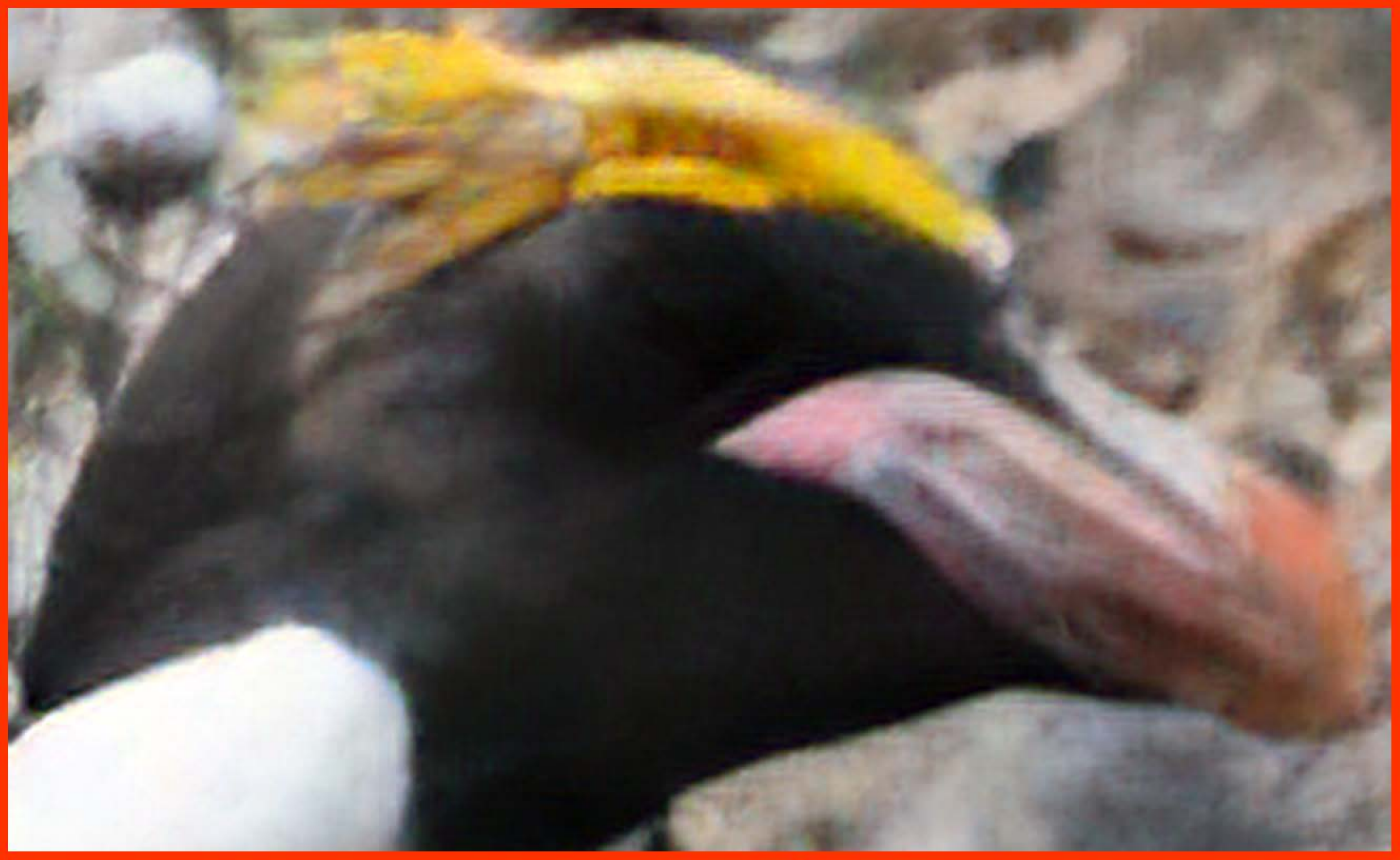} &
		\includegraphics[width=\swsix]{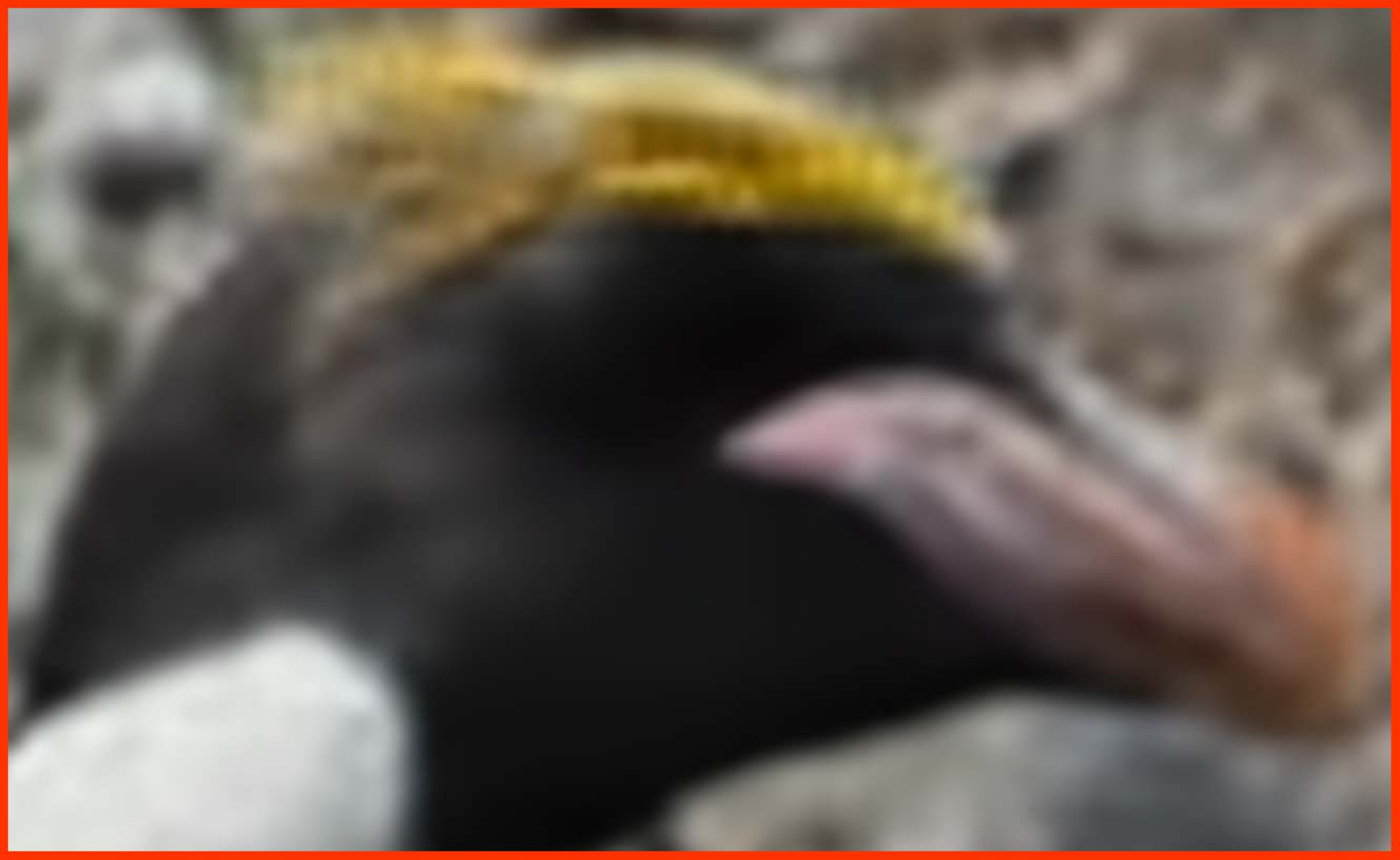} &
		\includegraphics[width=\swsix]{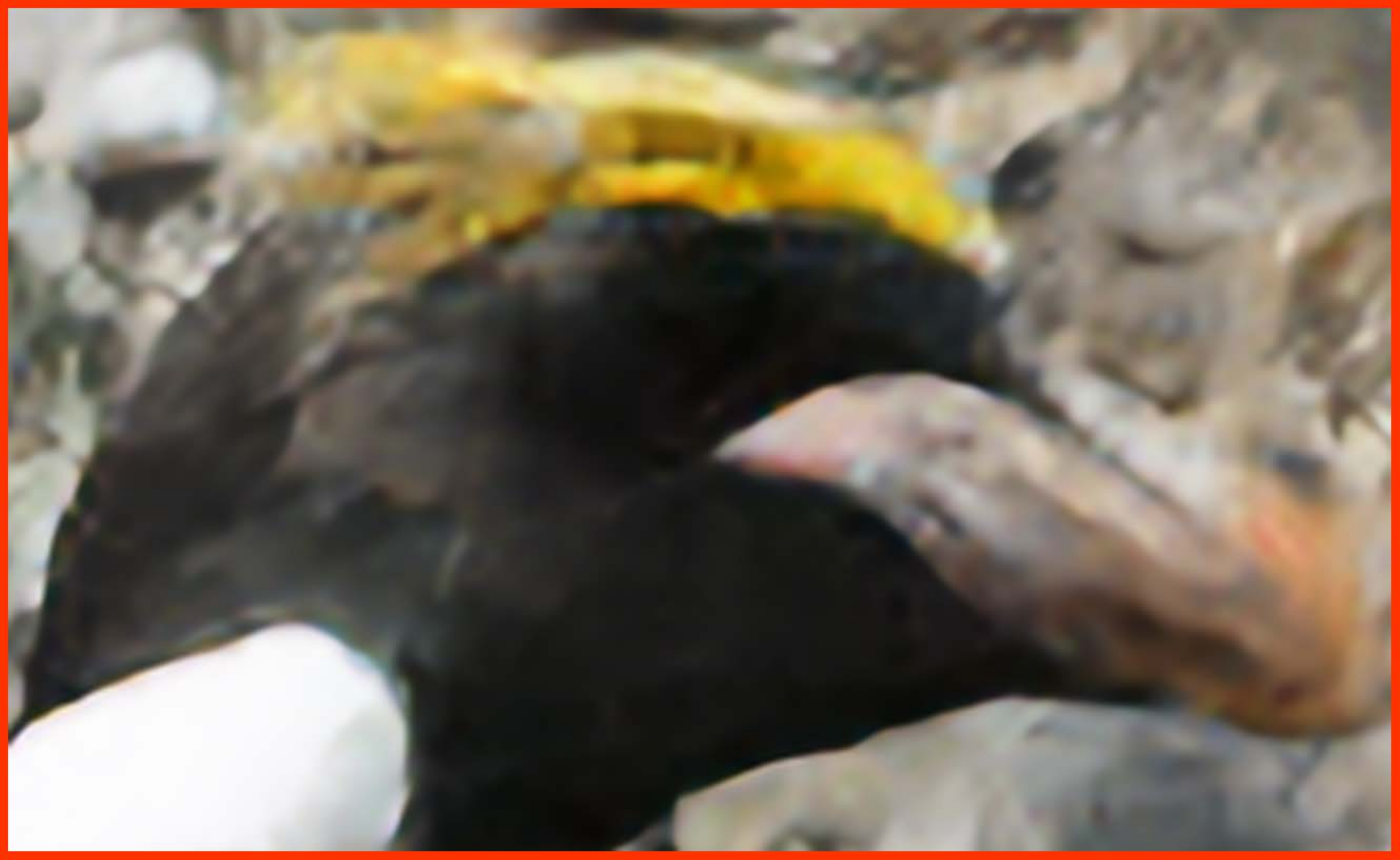} \\
		
		\footnotesize{(a) ground truth} & \footnotesize{(b) bicubic} &  \footnotesize{(c) EDSR$^+$~\cite{lim2017enhanced}} & \footnotesize{(d) SRGAN$^+$~\cite{ledig2016photo}} & \footnotesize{(e) BM3D+EDSR} & \footnotesize{(f) \textbf{CinCGAN (ours)}} \\
		\footnotesize{PSNR/SSIM} & \footnotesize{23.22/0.64} & \footnotesize{26.23/0.68} & \footnotesize{24.06/0.58} & \footnotesize{23.06/0.65} & \footnotesize{24.83/0.65} \\

		\includegraphics[width=\swsix]{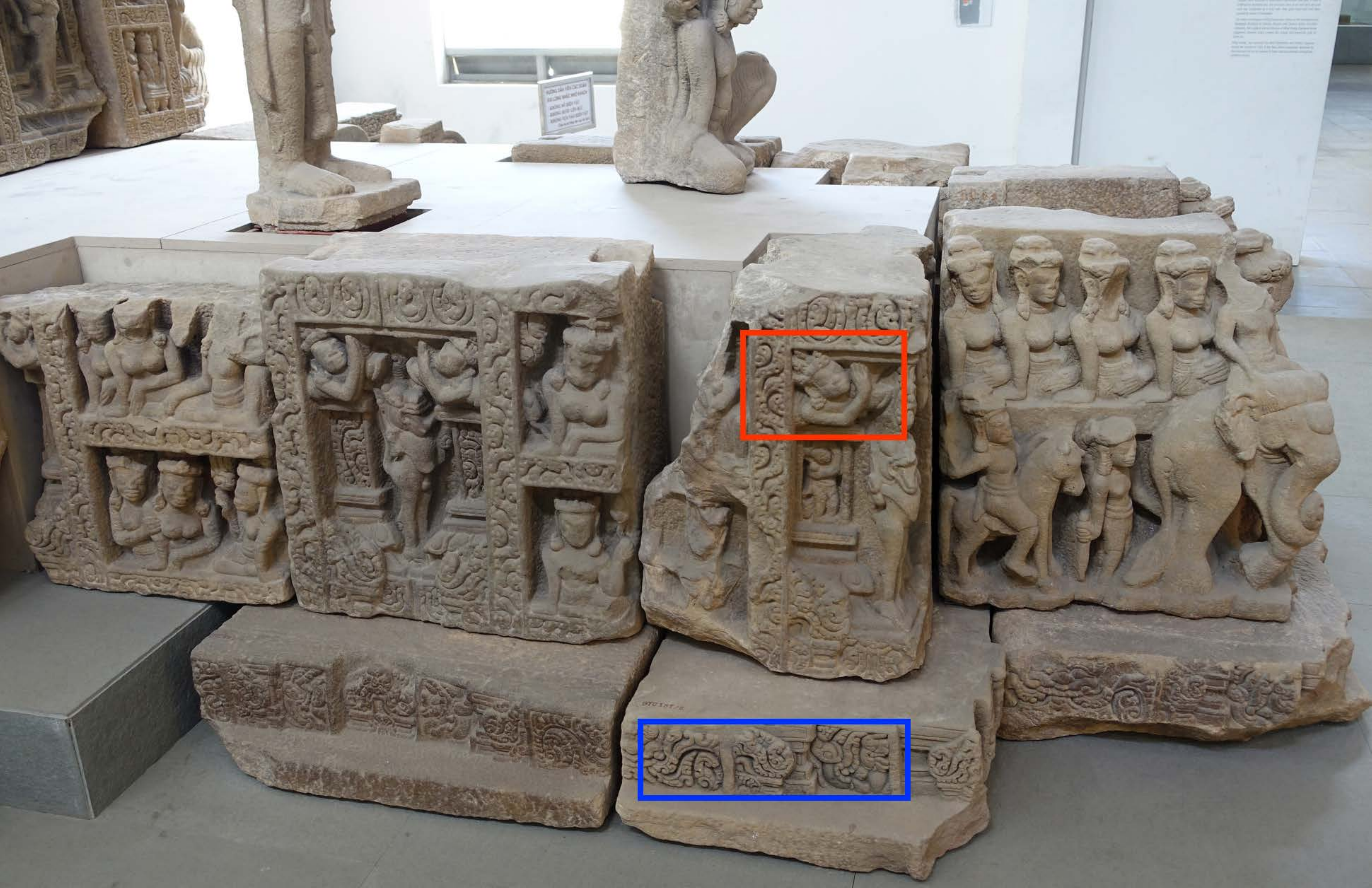} &
		\includegraphics[width=\swsix]{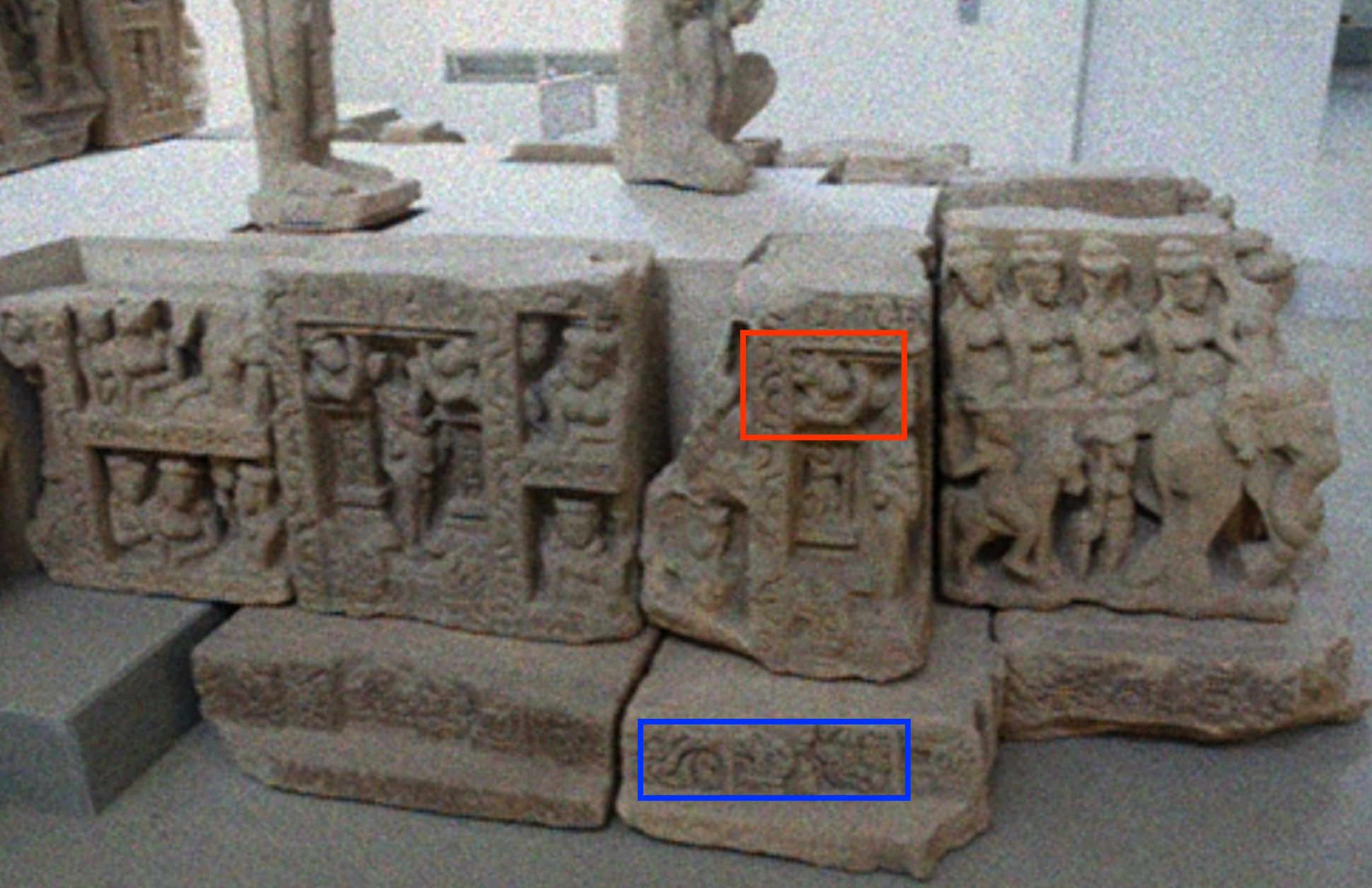} &
		\includegraphics[width=\swsix]{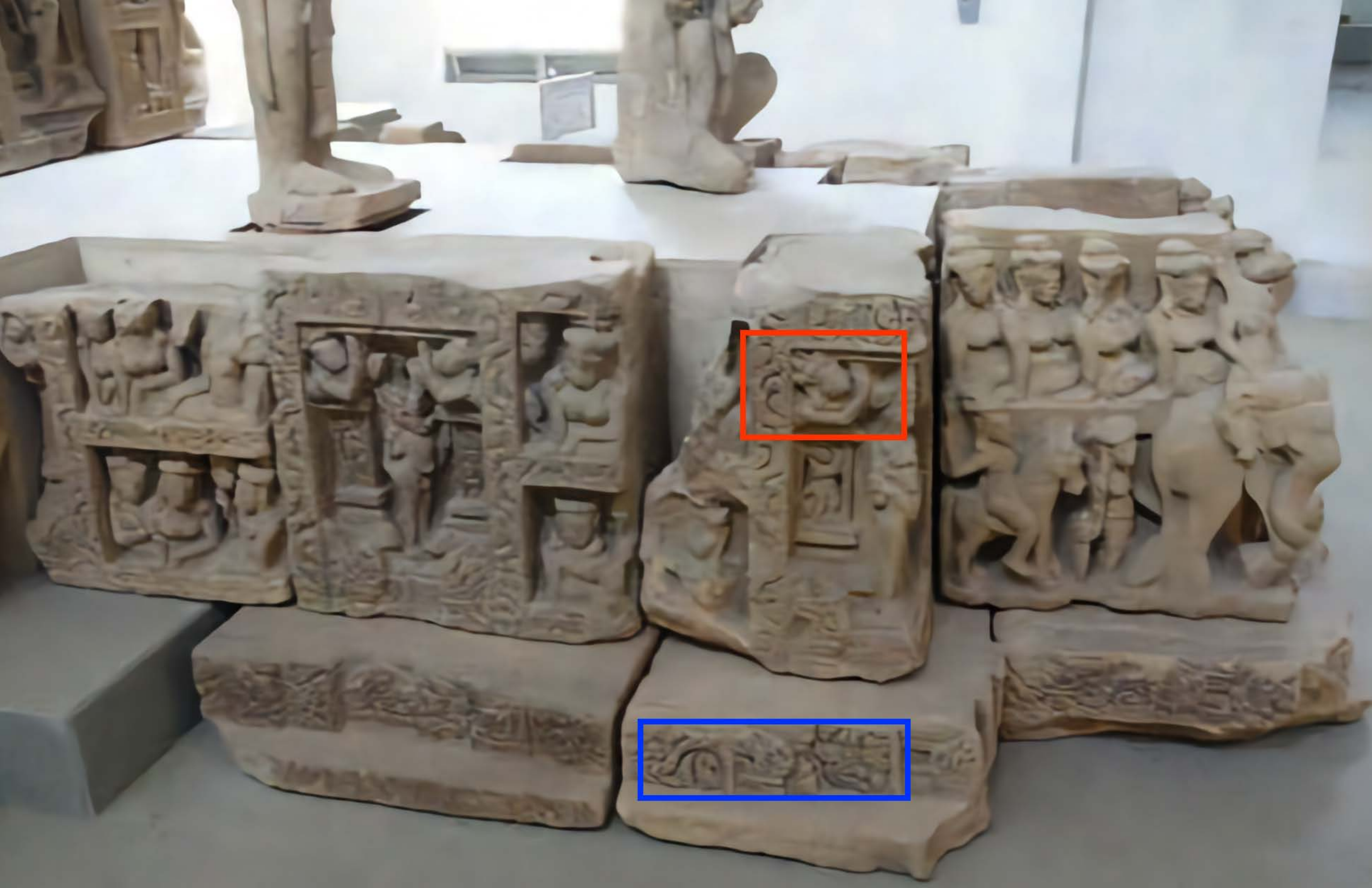} &
		\includegraphics[width=\swsix]{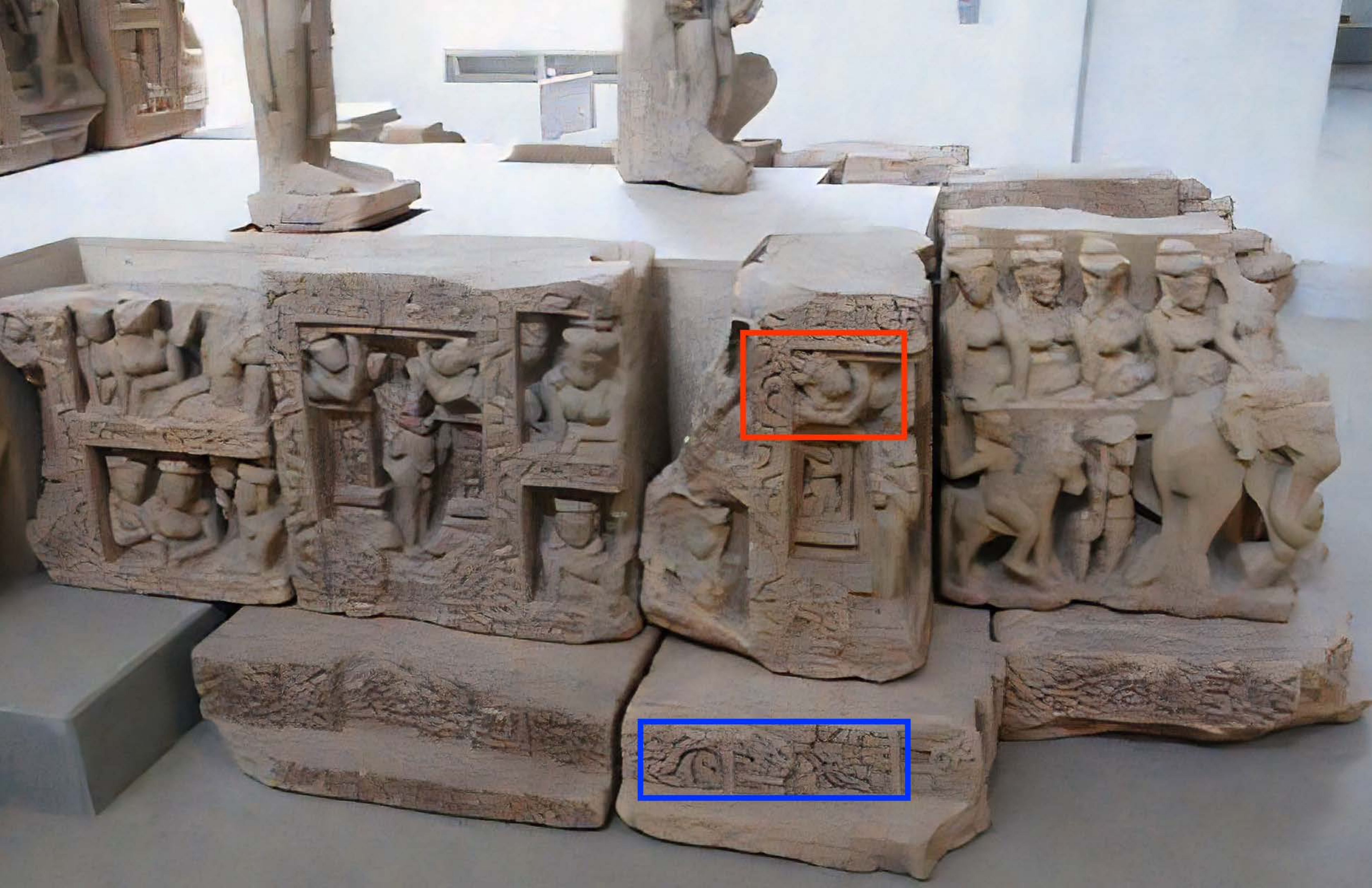} &
		\includegraphics[width=\swsix]{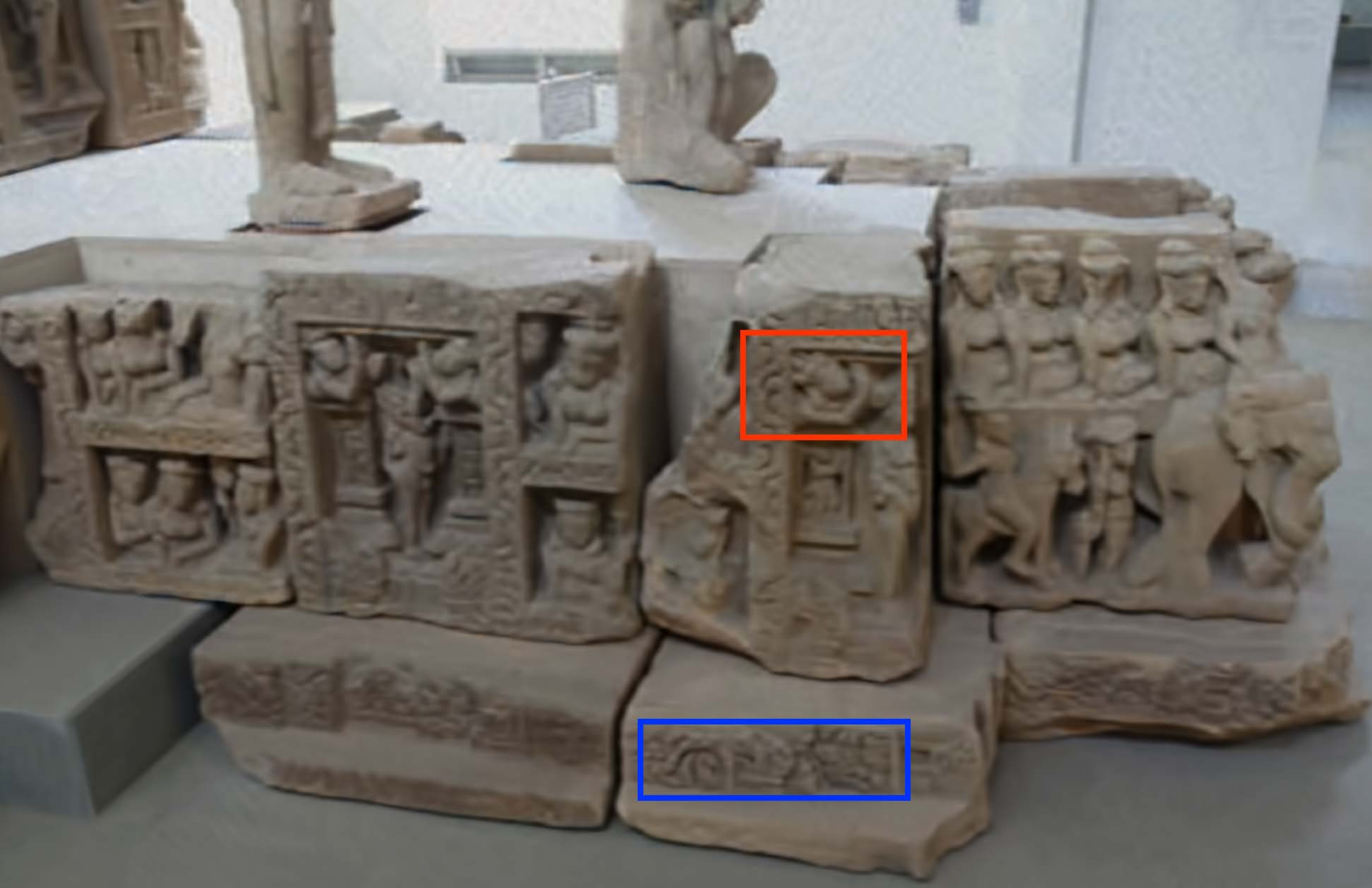} &
		\includegraphics[width=\swsix]{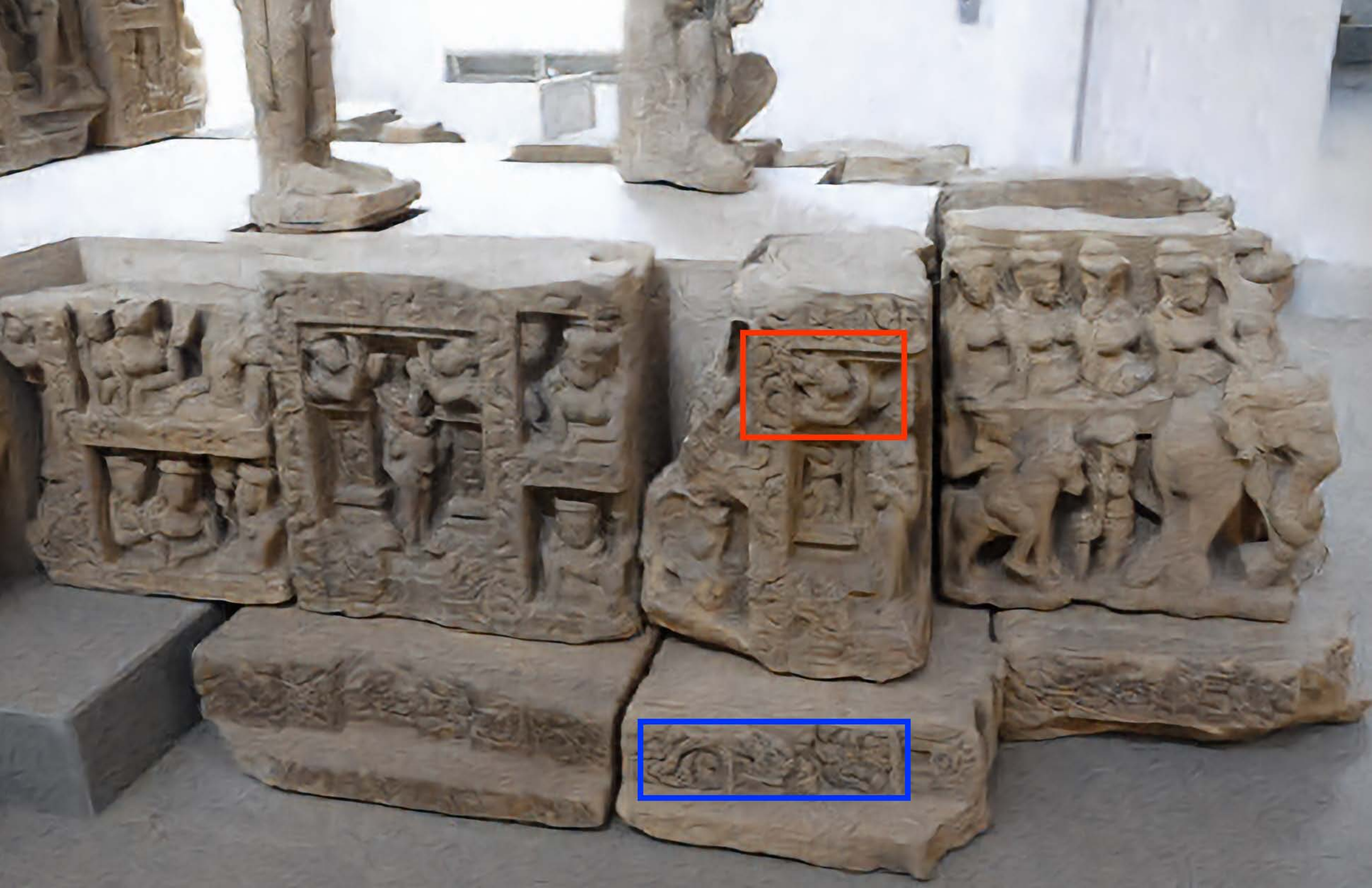} \\
		
		\includegraphics[width=\swsix]{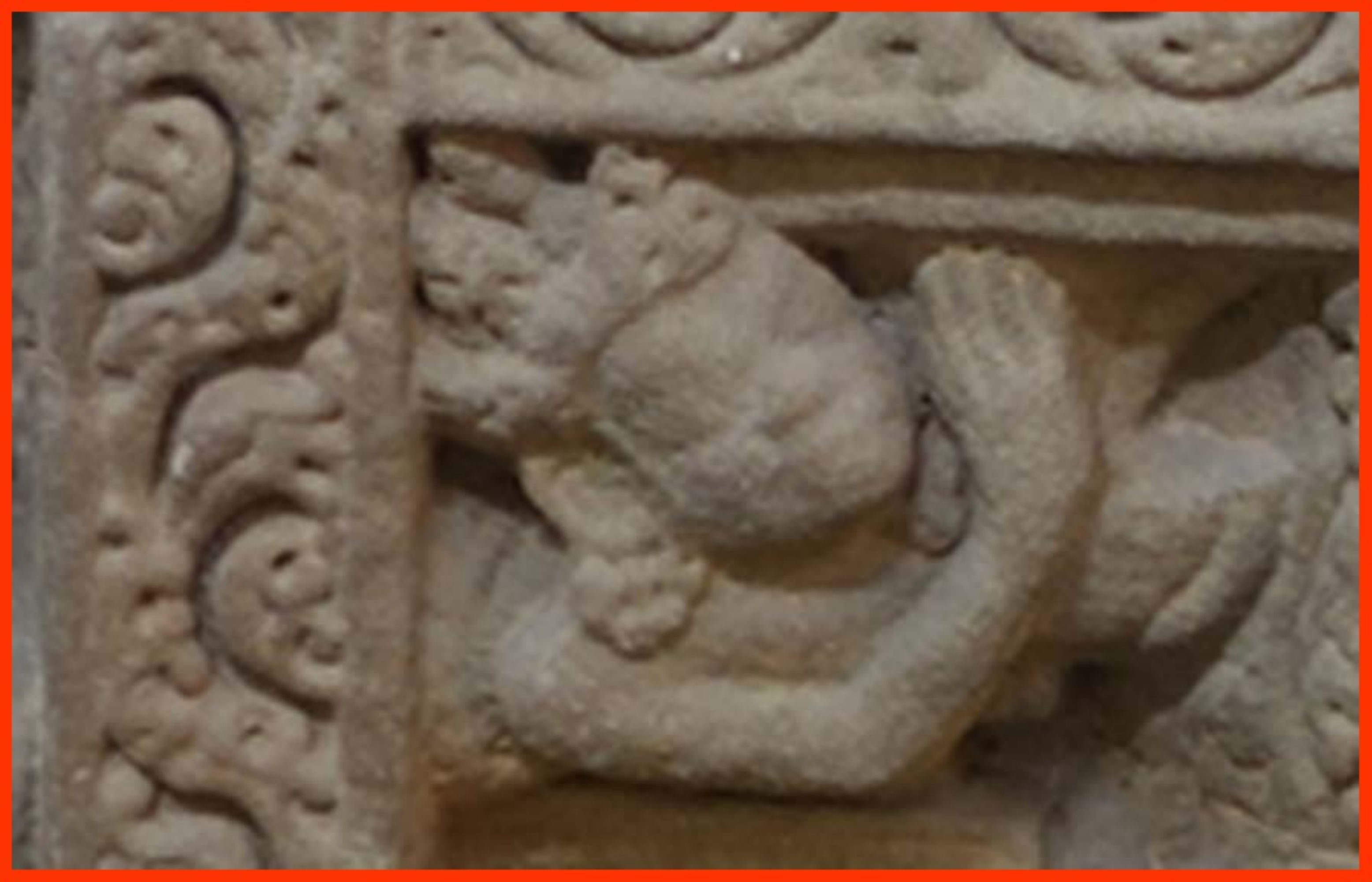} &
		\includegraphics[width=\swsix]{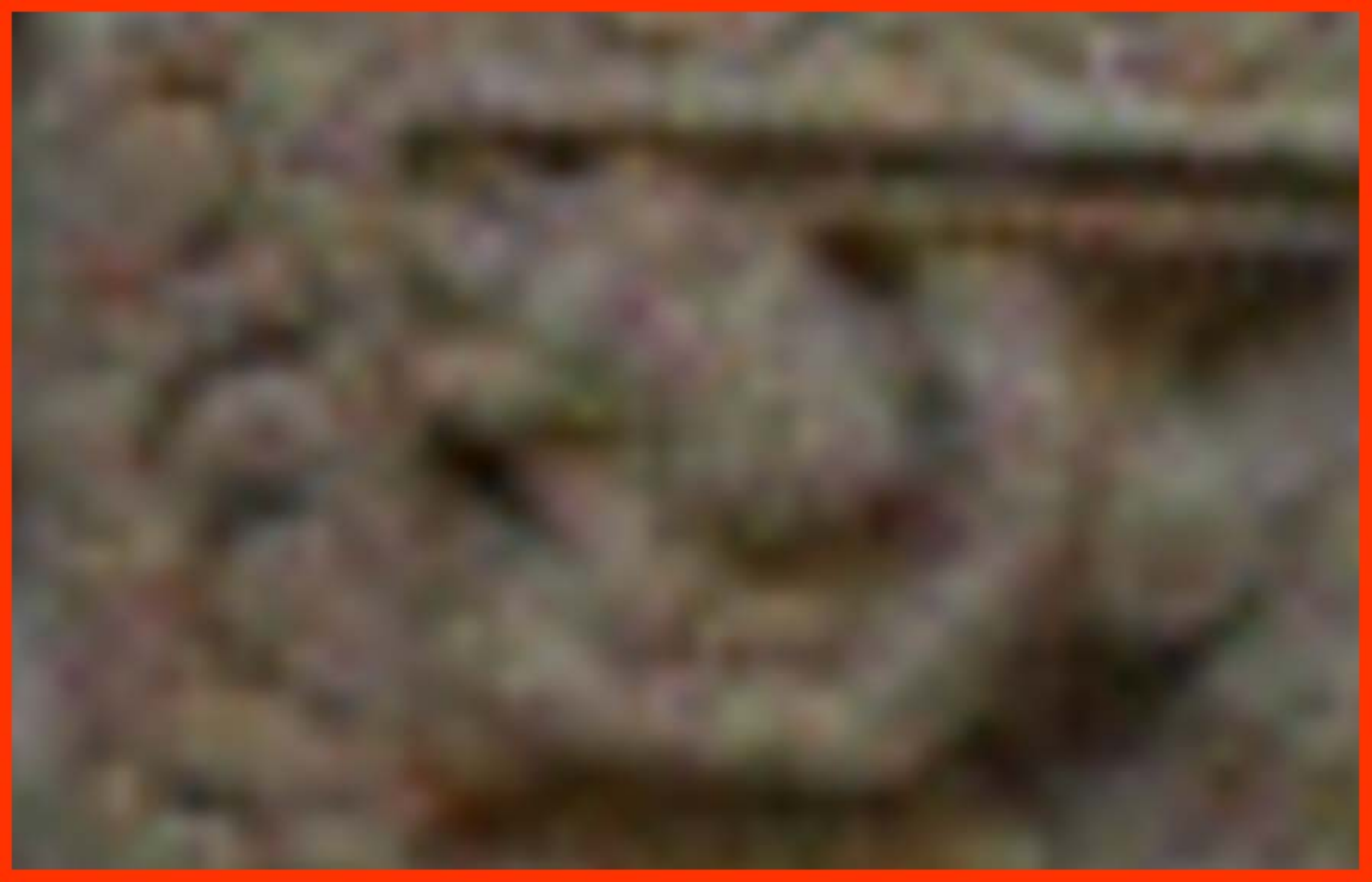} &
		\includegraphics[width=\swsix]{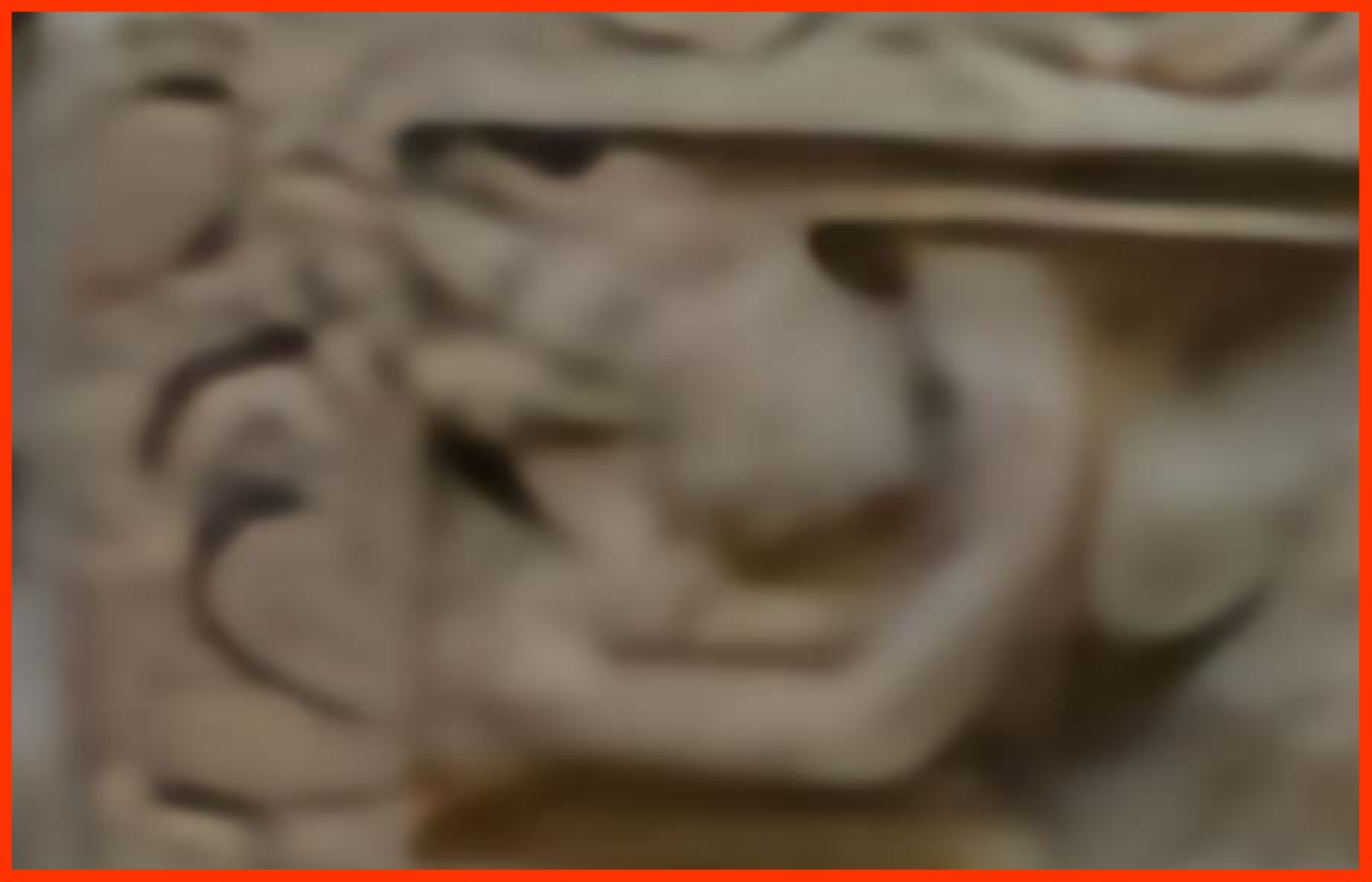} &
		\includegraphics[width=\swsix]{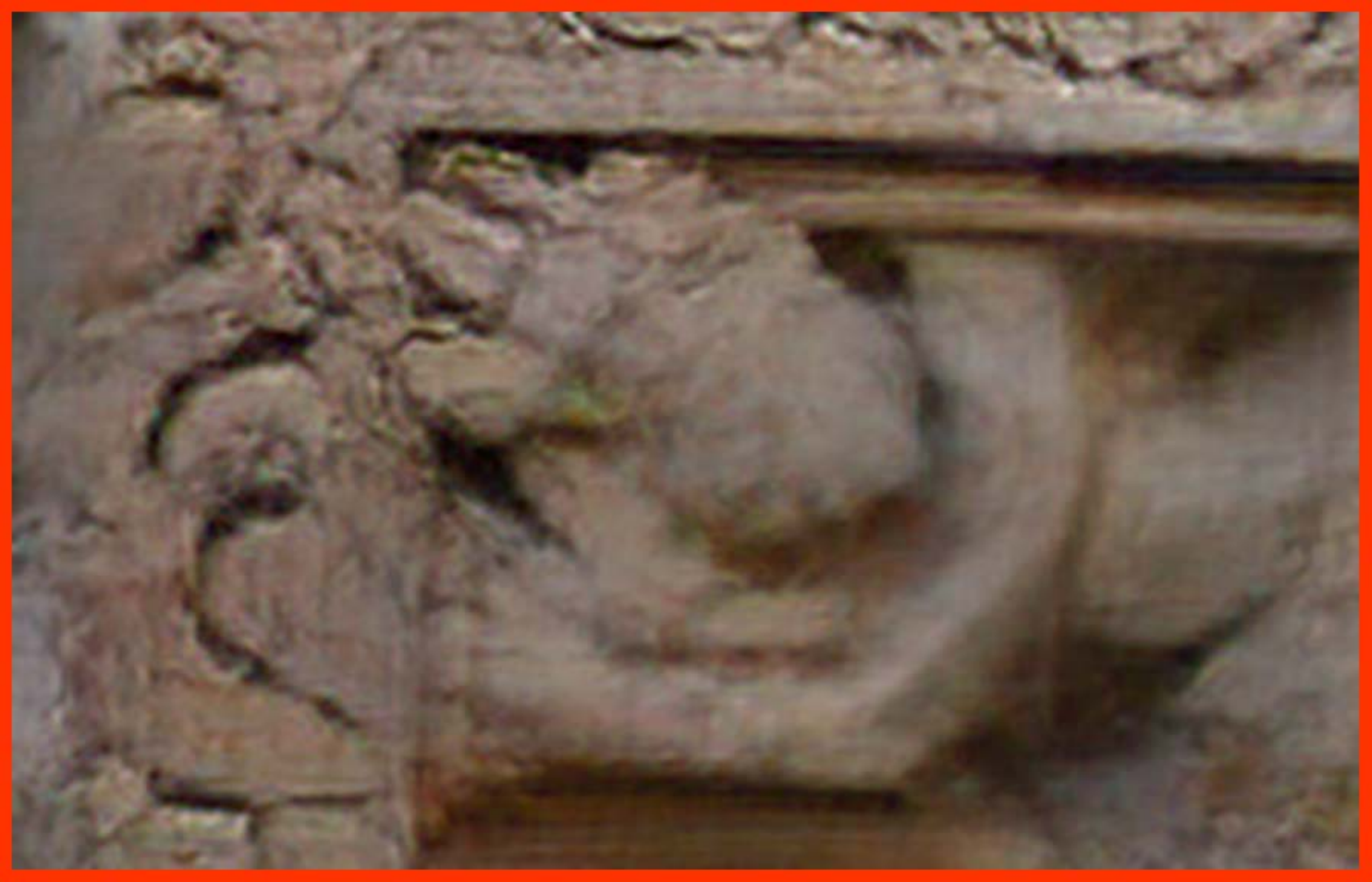} &
		\includegraphics[width=\swsix]{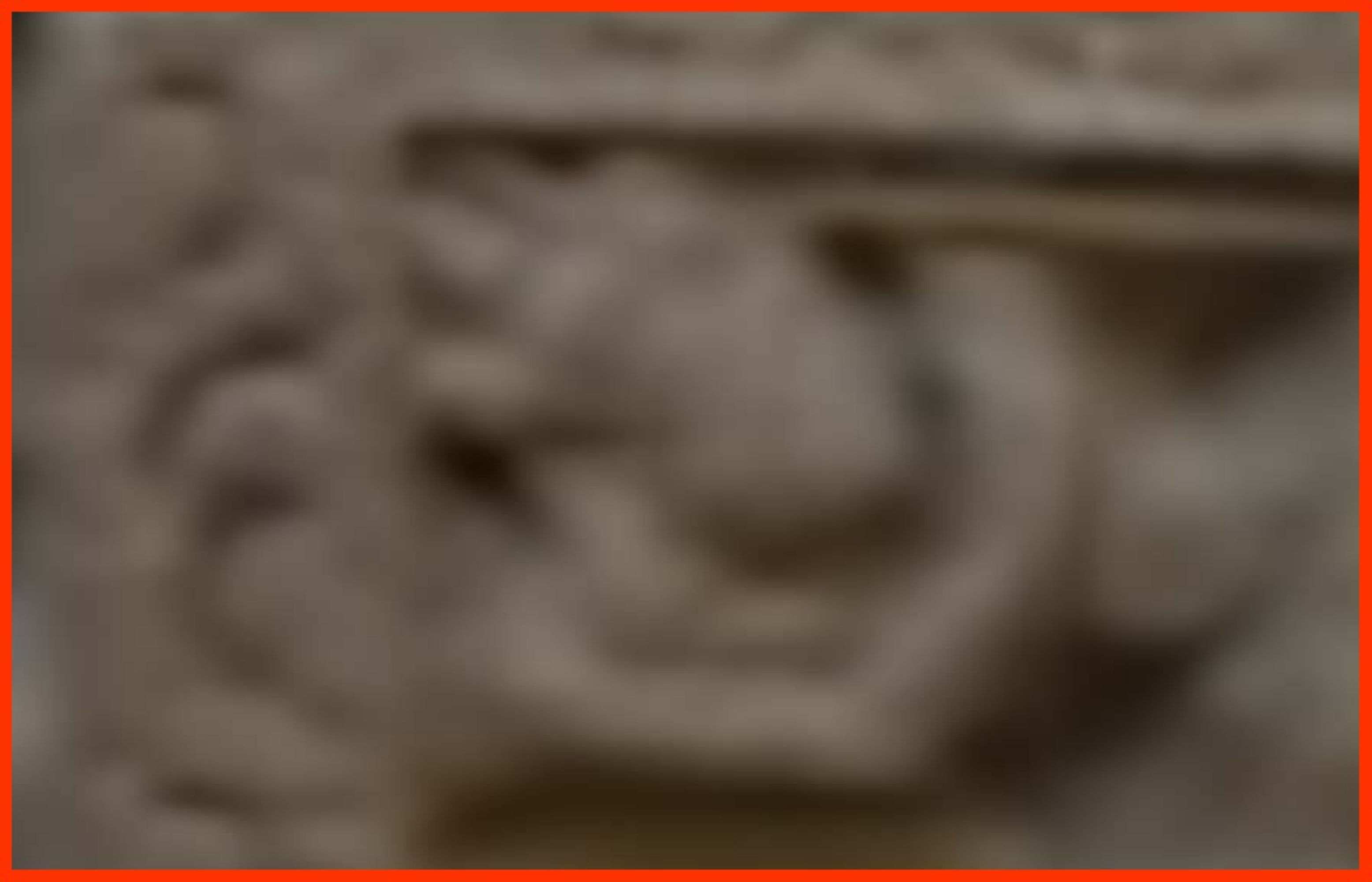} &
		\includegraphics[width=\swsix]{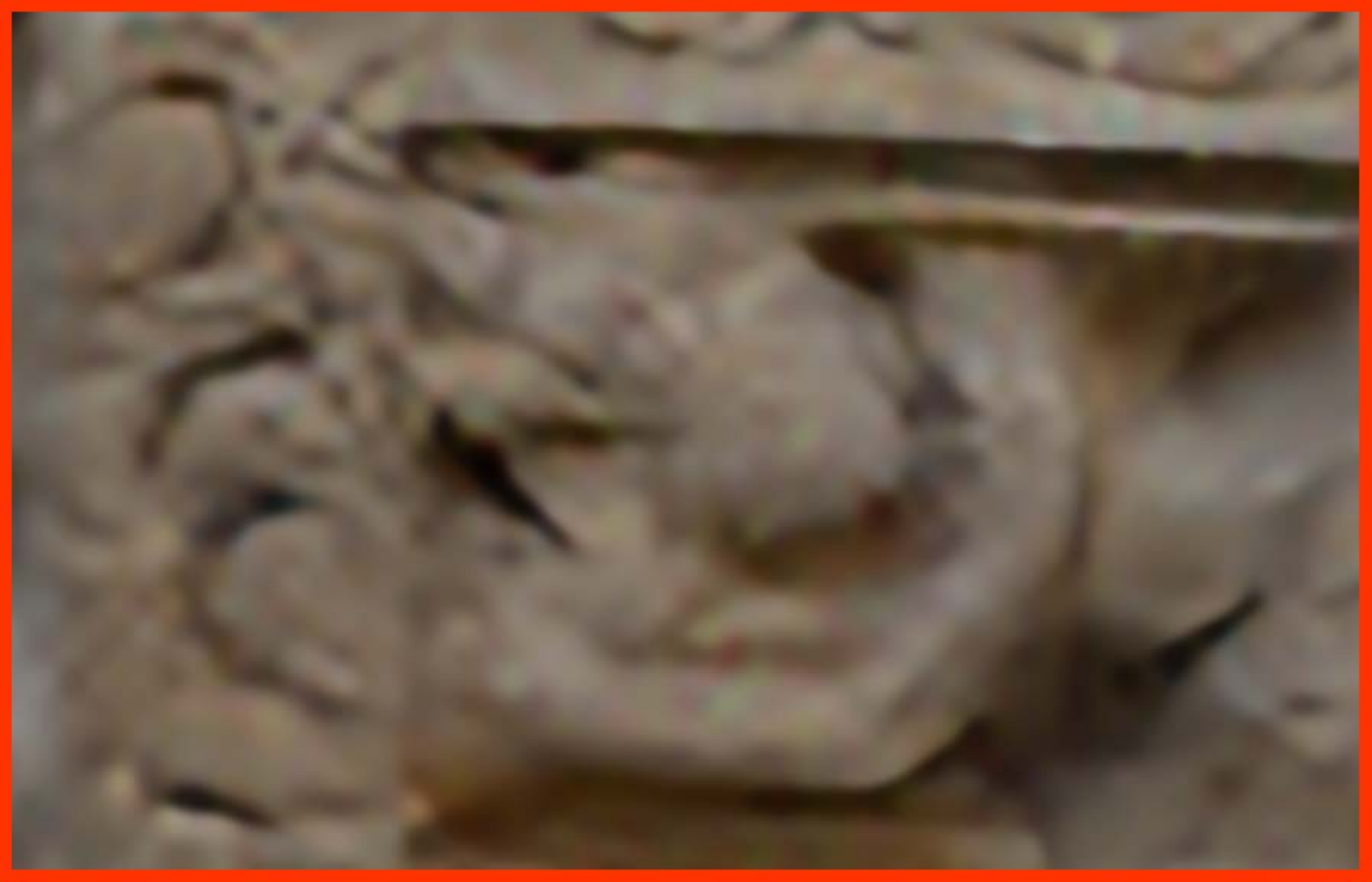} \\
		
		\includegraphics[width=\swsix]{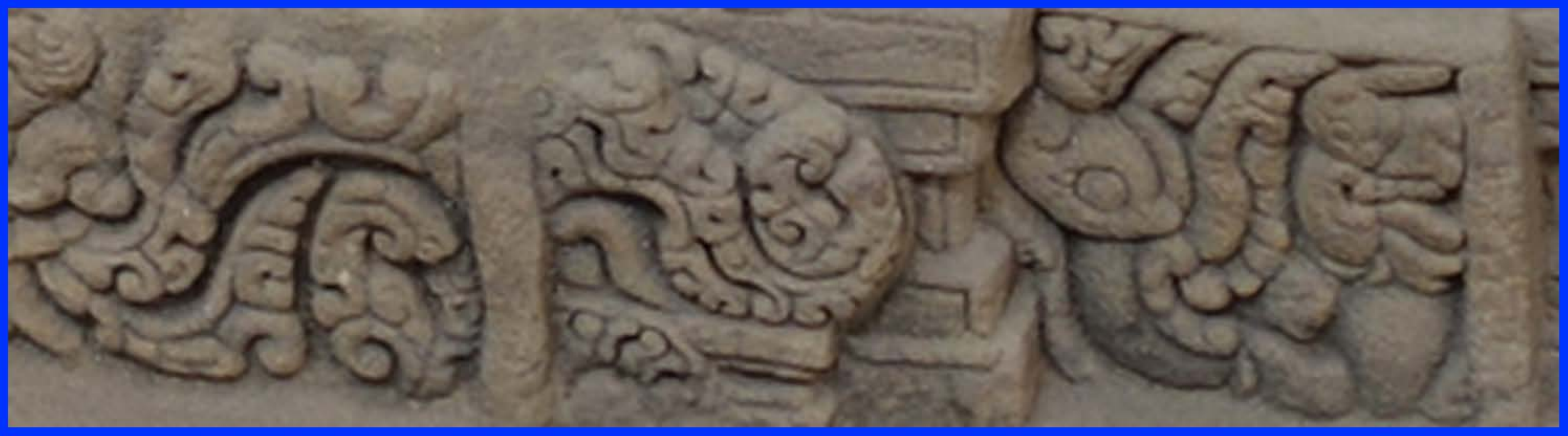} &
		\includegraphics[width=\swsix]{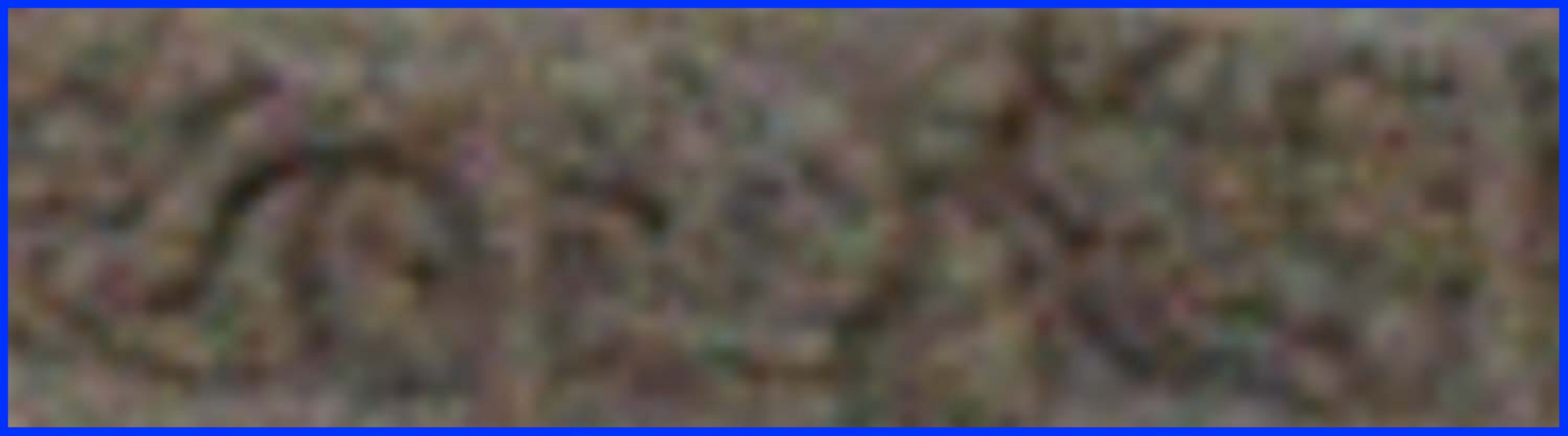} &
		\includegraphics[width=\swsix]{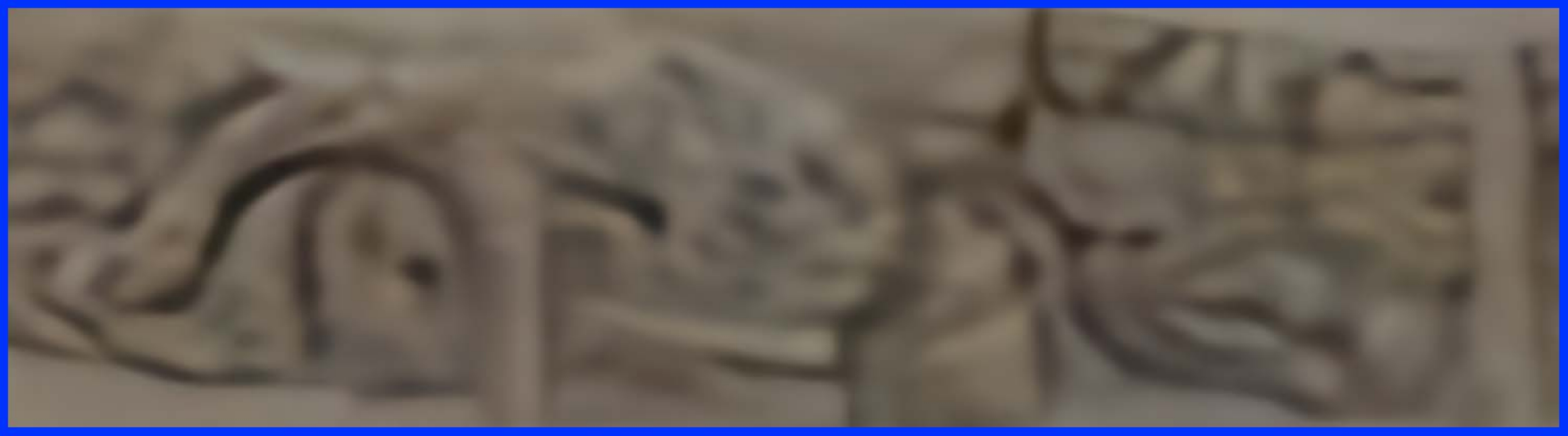} &
		\includegraphics[width=\swsix]{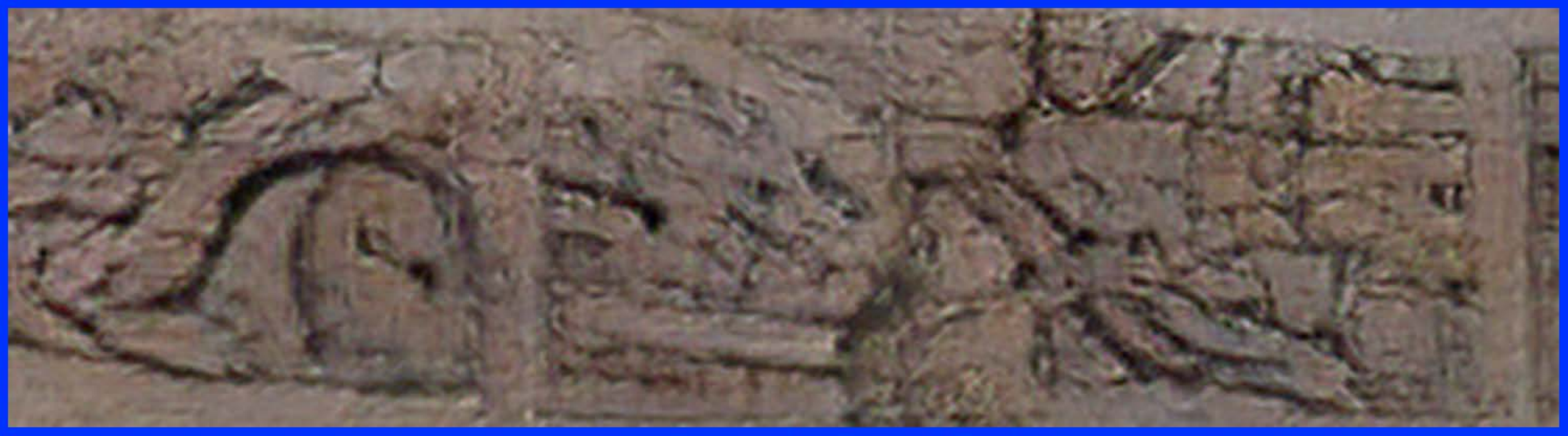} &
		\includegraphics[width=\swsix]{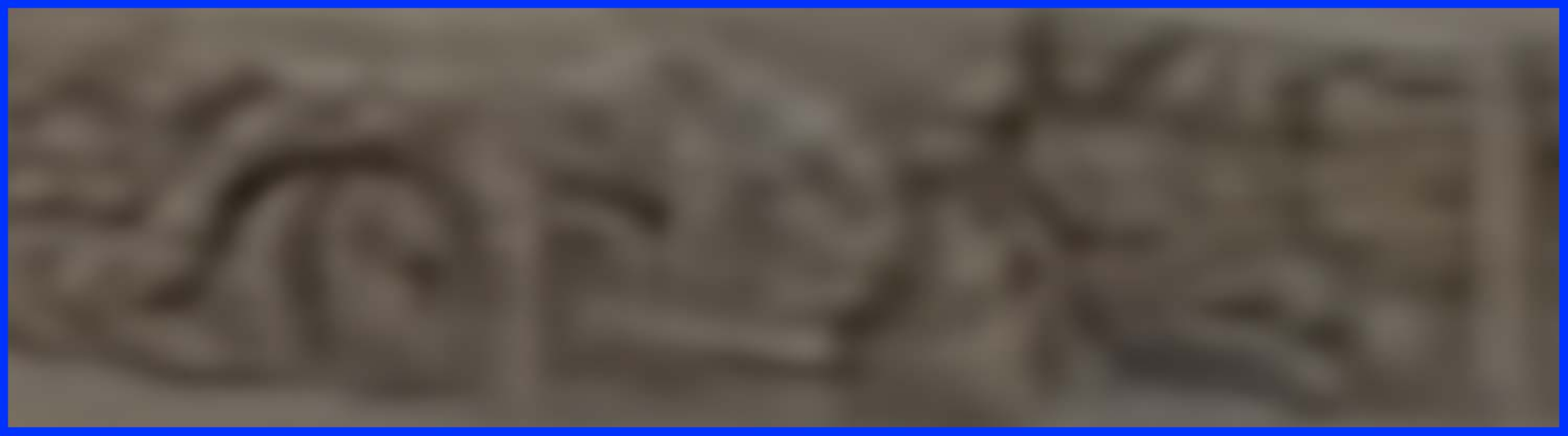} &
		\includegraphics[width=\swsix]{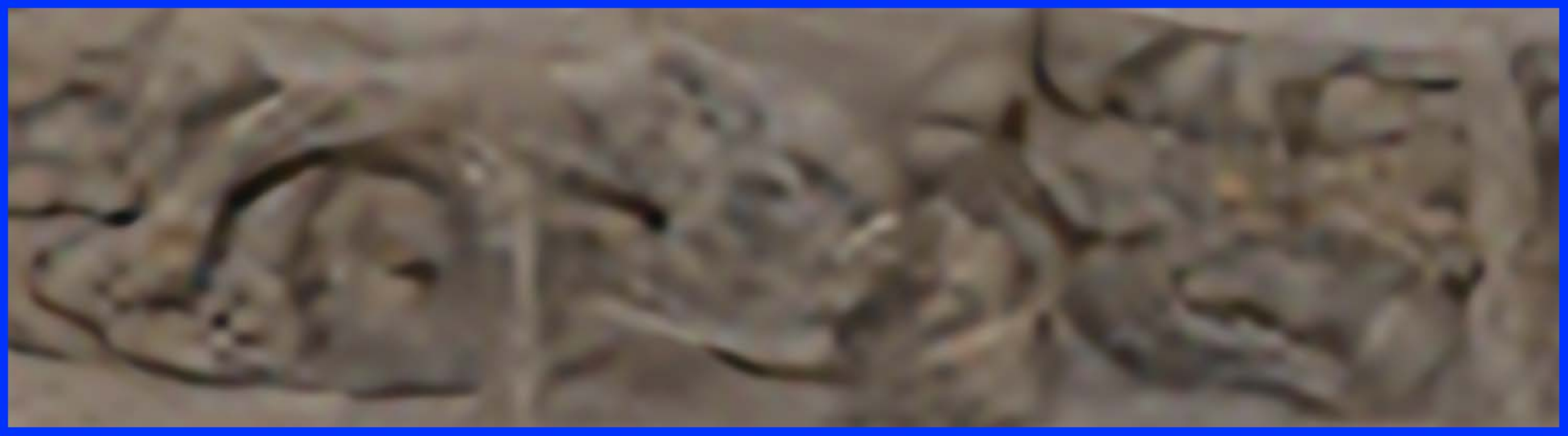} \\
		
		\footnotesize{(a) ground truth} & \footnotesize{(b) bicubic} & \footnotesize{(c) EDSR$^+$~\cite{lim2017enhanced}} & \footnotesize{(d) SRGAN$^+$~\cite{ledig2016photo}} & \footnotesize{(e) BM3D+EDSR} & \footnotesize{(f) \textbf{CinCGAN (ours)}}  \\
		\footnotesize{PSNR/SSIM} &  \footnotesize{22.25/0.68} &  \footnotesize{29.06/0.75} & \footnotesize{27.36/0.68} & \footnotesize{22.18/0.72} & \footnotesize{27.95/0.72} \\

		\includegraphics[width=\swsix]{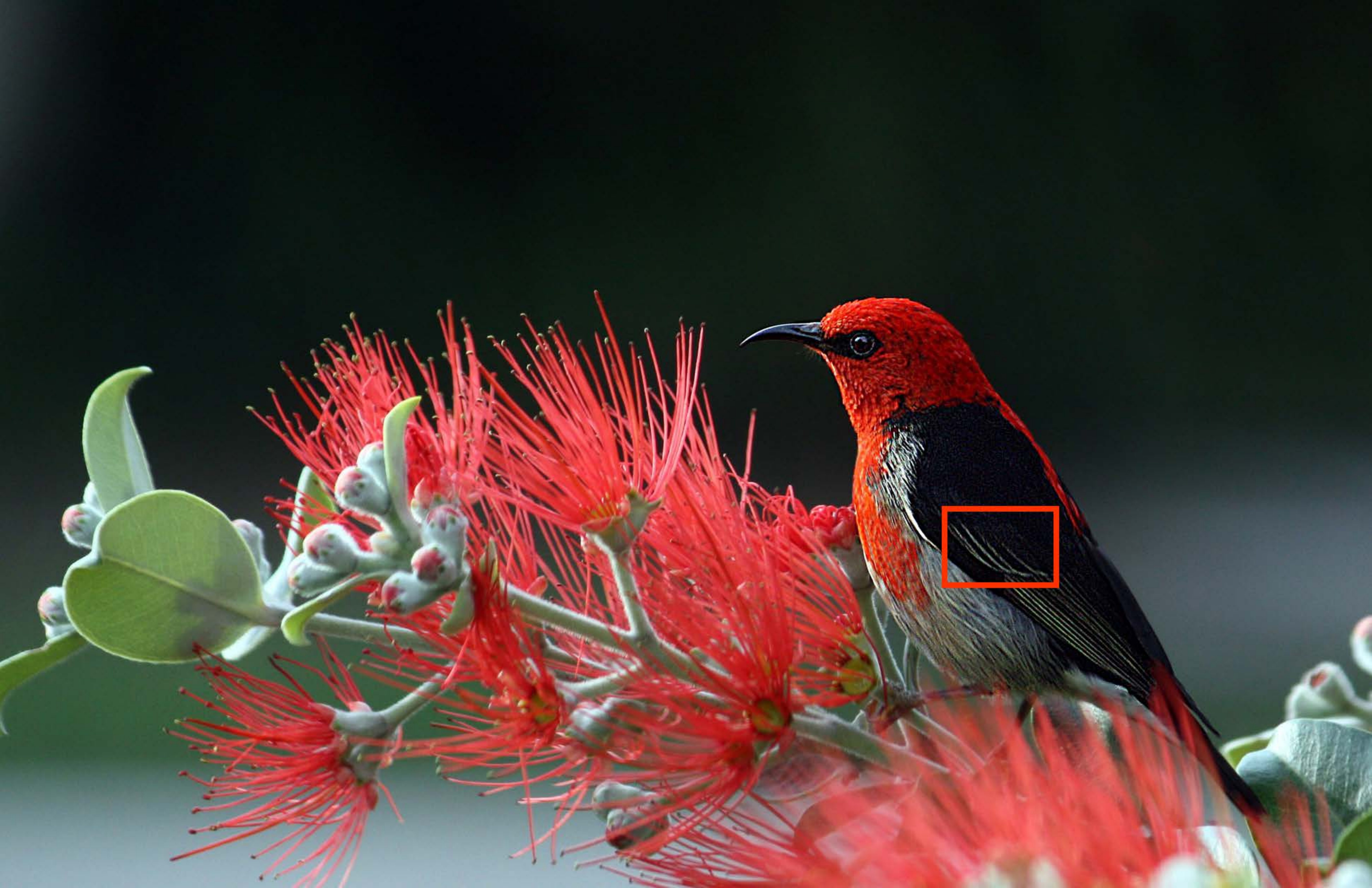} &
		\includegraphics[width=\swsix]{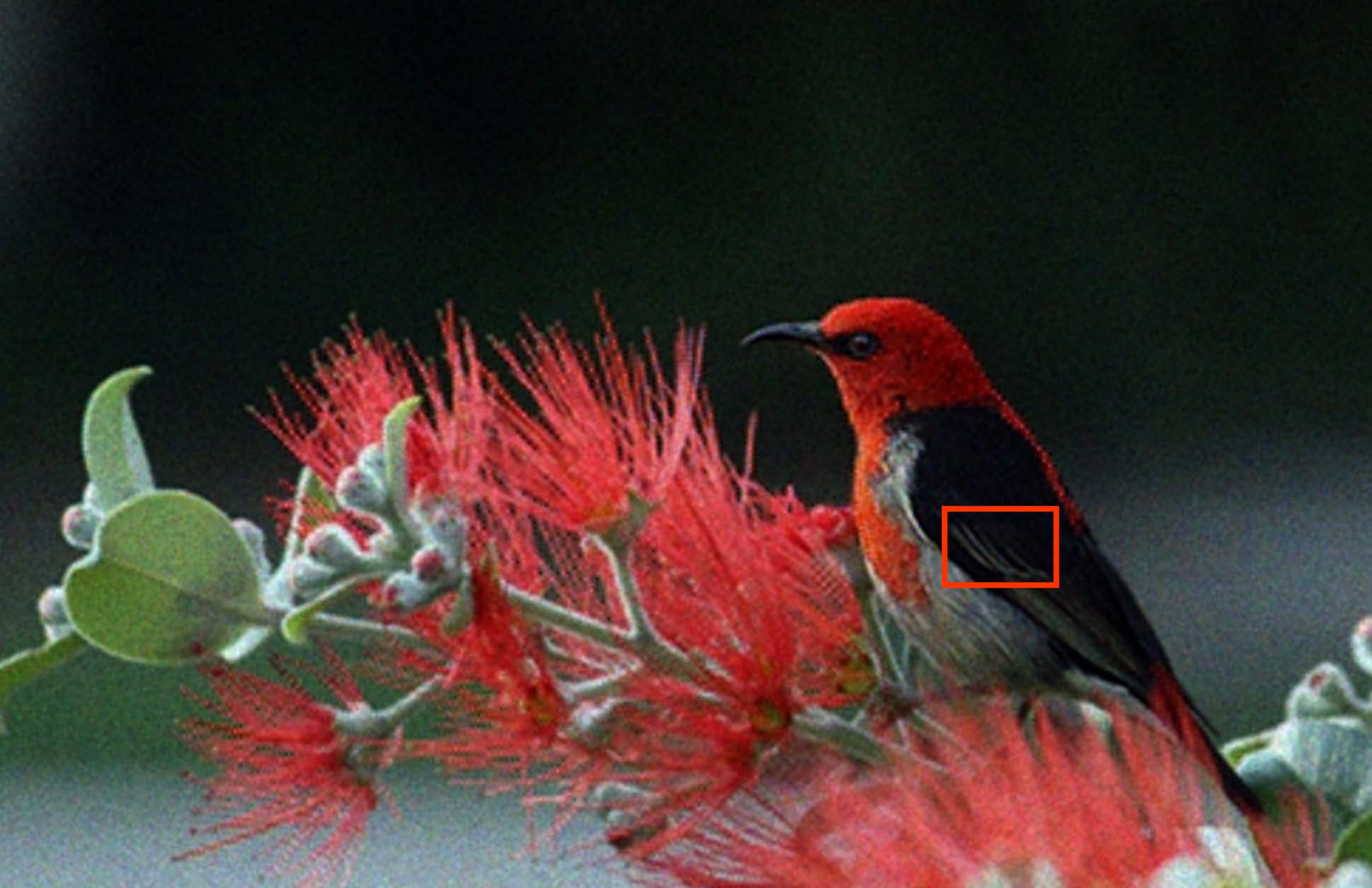} &
		\includegraphics[width=\swsix]{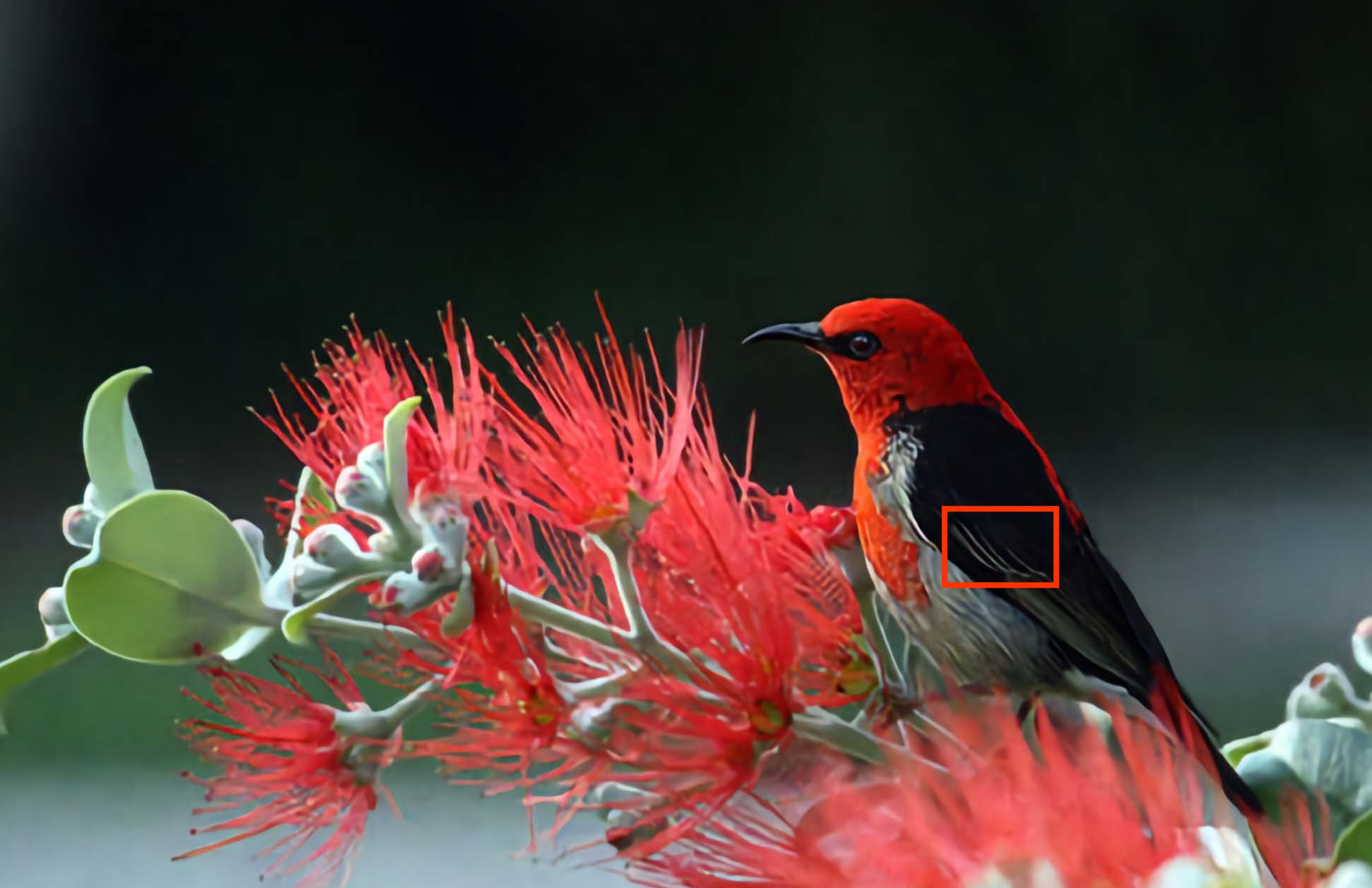} &
		\includegraphics[width=\swsix]{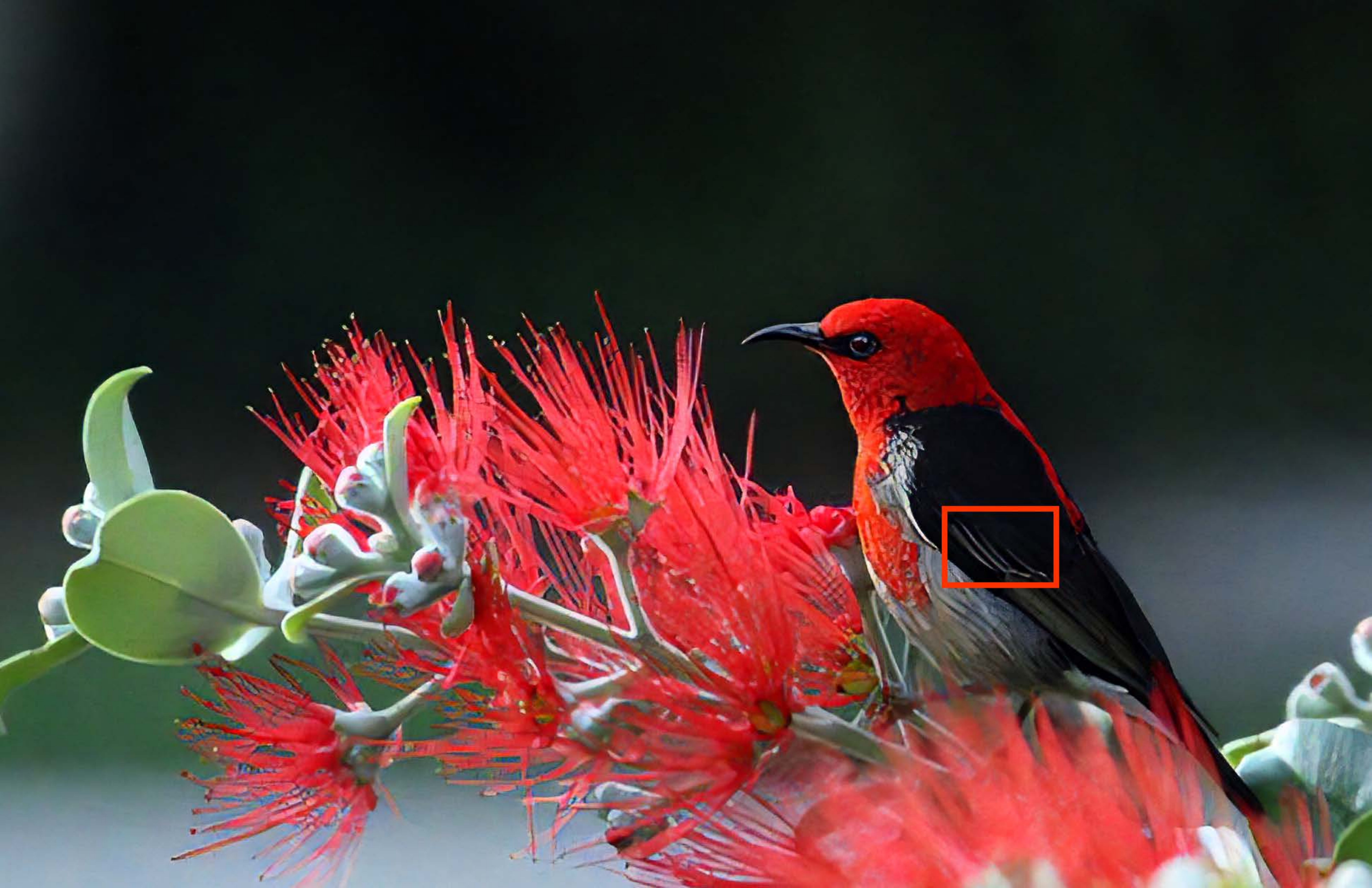} &
		\includegraphics[width=\swsix]{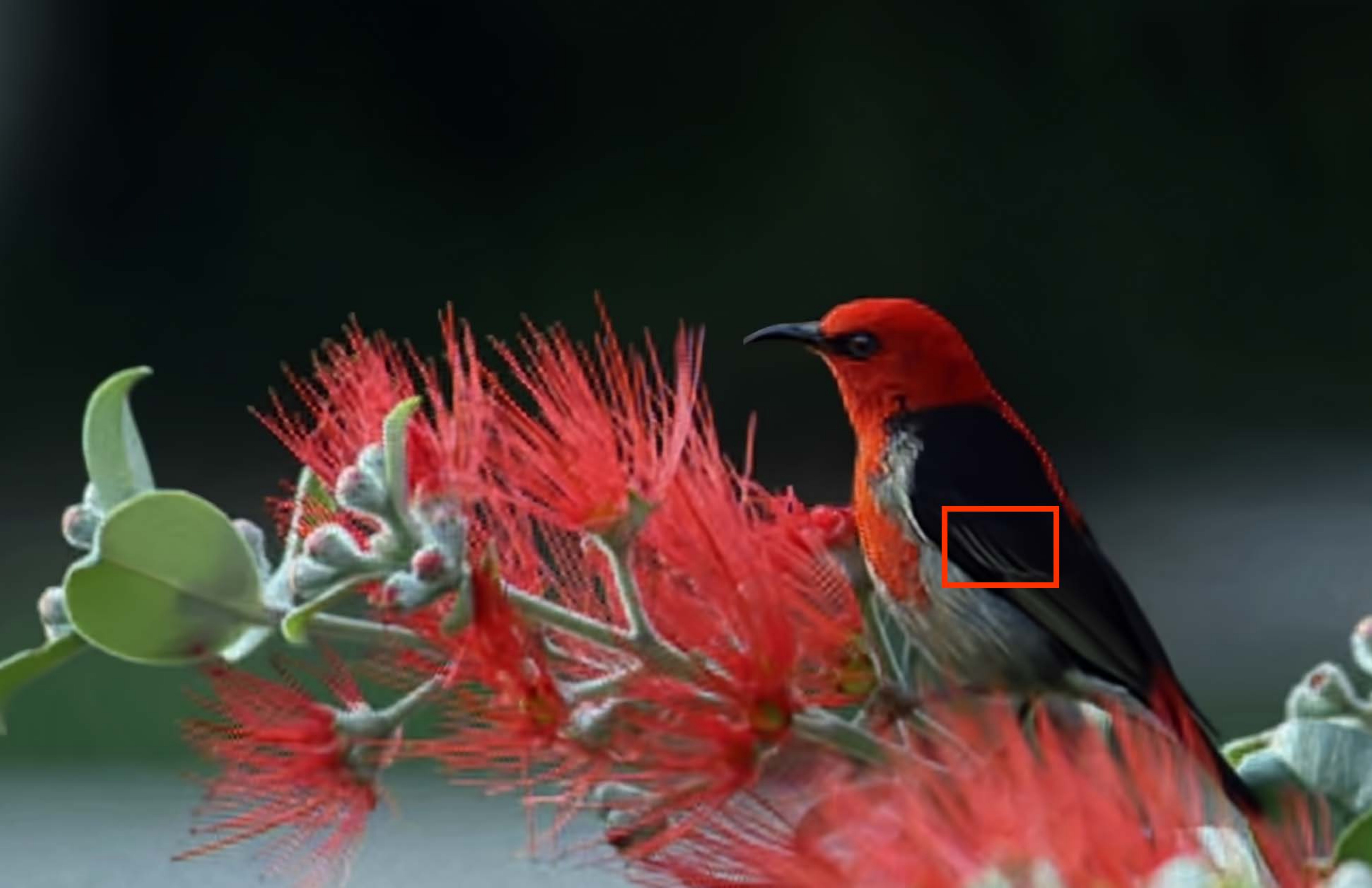} &
		\includegraphics[width=\swsix]{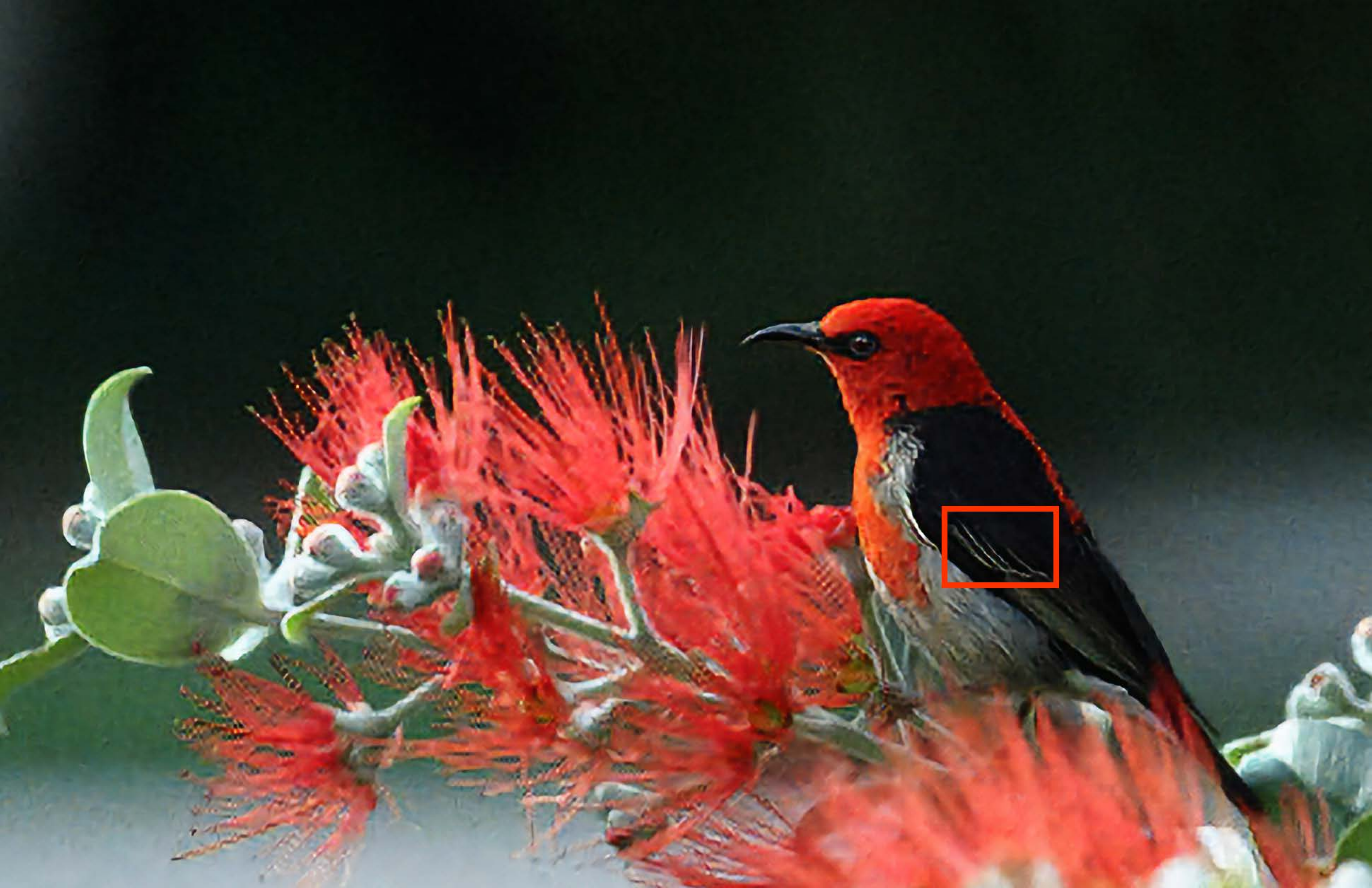} \\
		
		\includegraphics[width=\swsix]{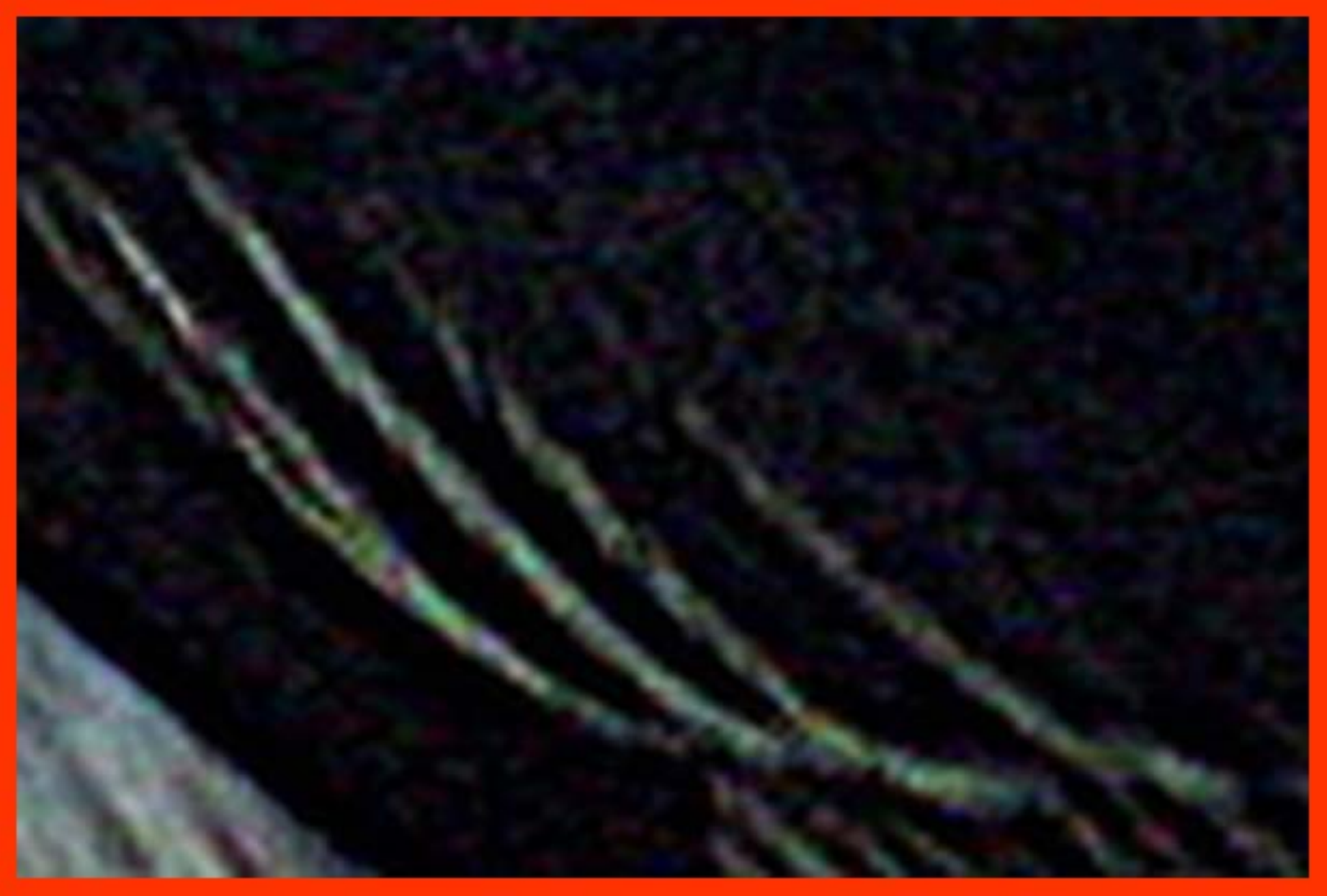} &
		\includegraphics[width=\swsix]{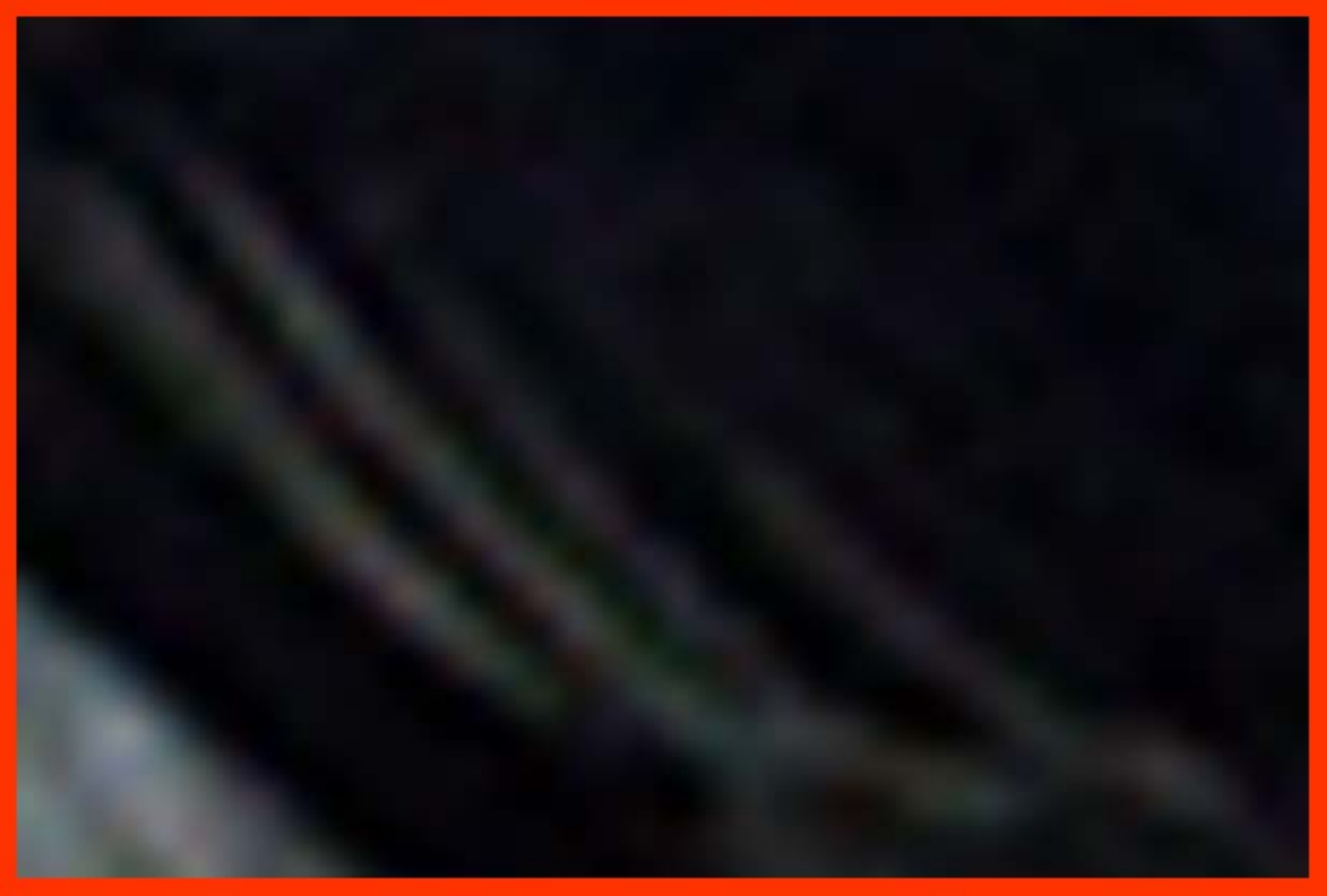} &
		\includegraphics[width=\swsix]{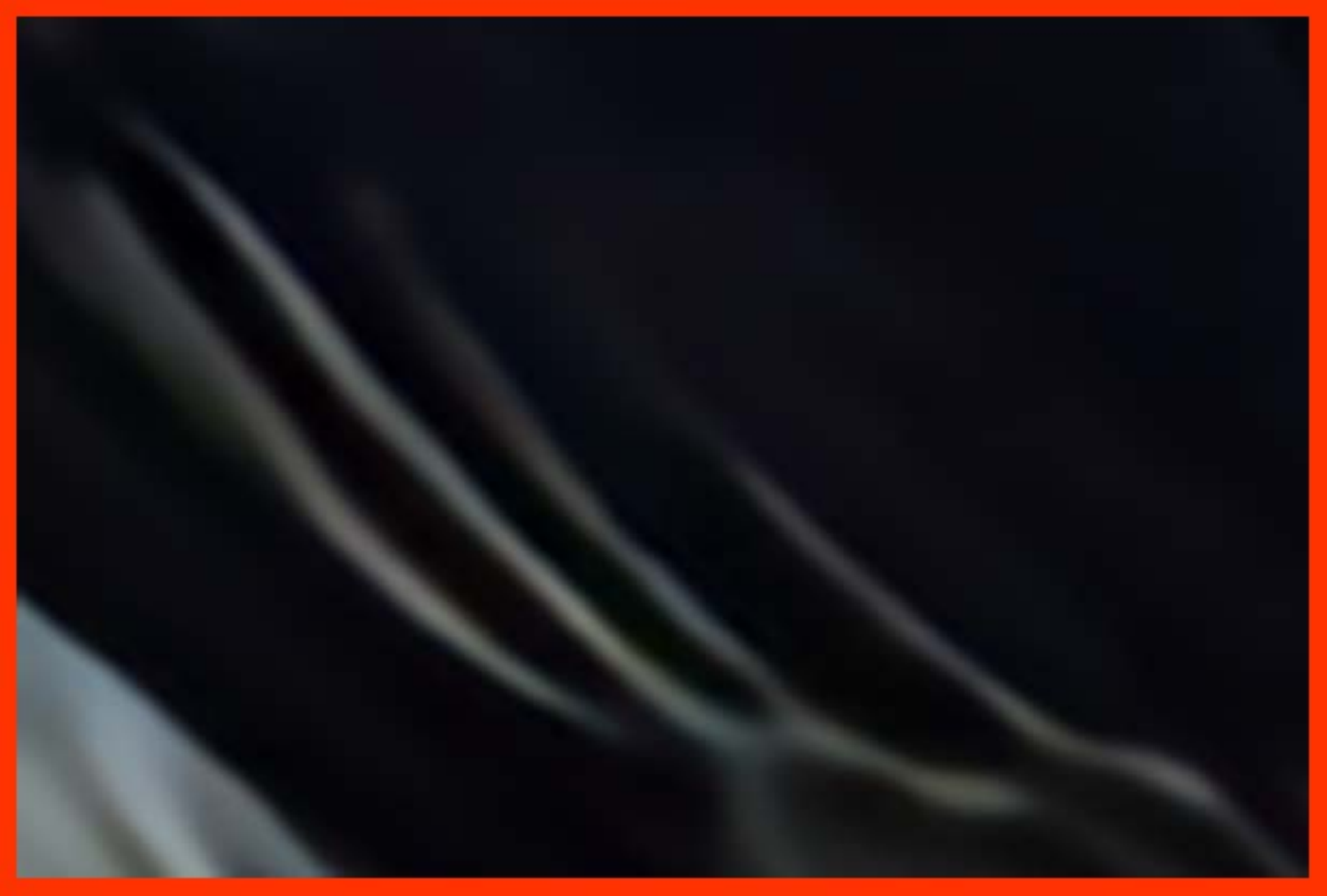} &
		\includegraphics[width=\swsix]{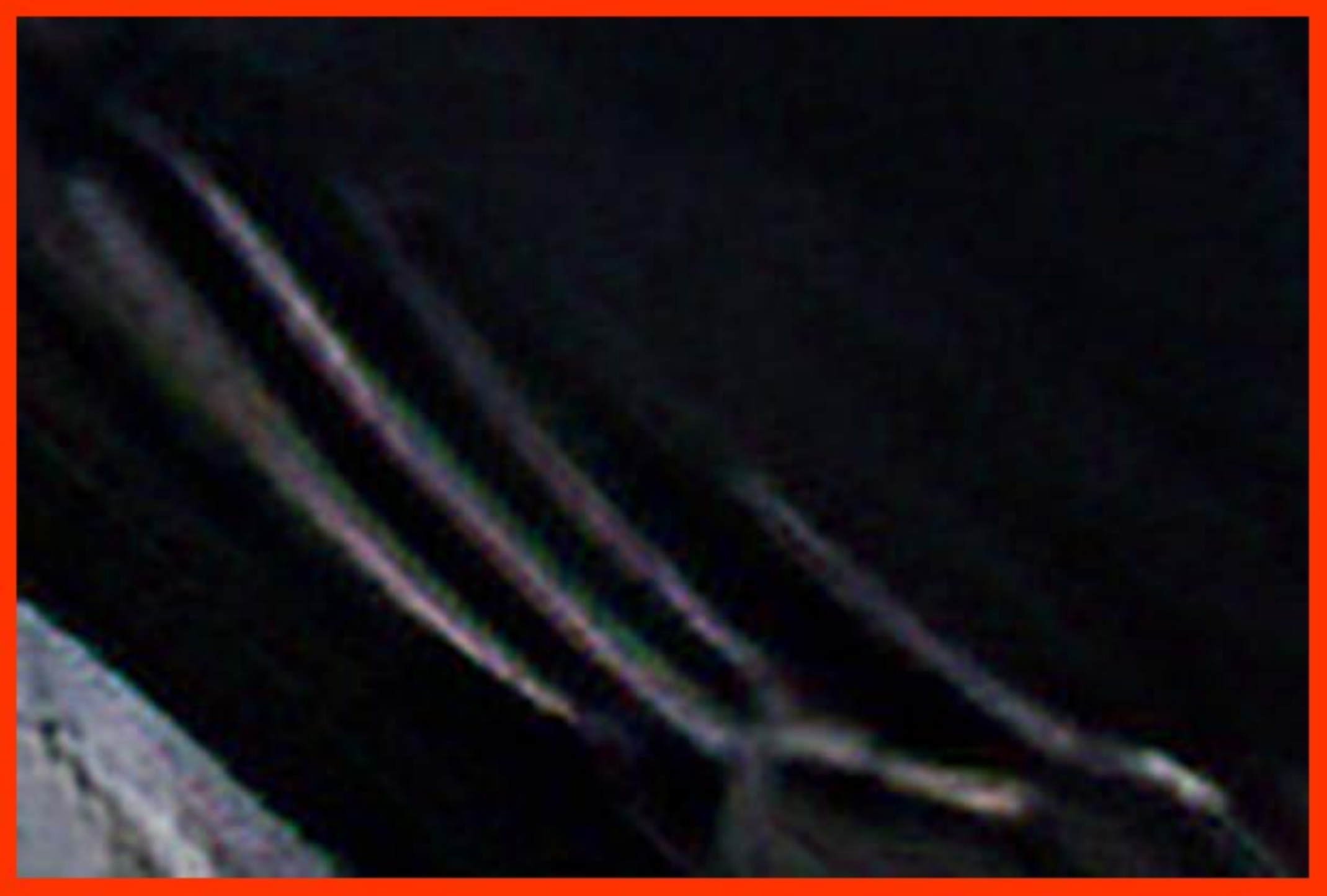} &
		\includegraphics[width=\swsix]{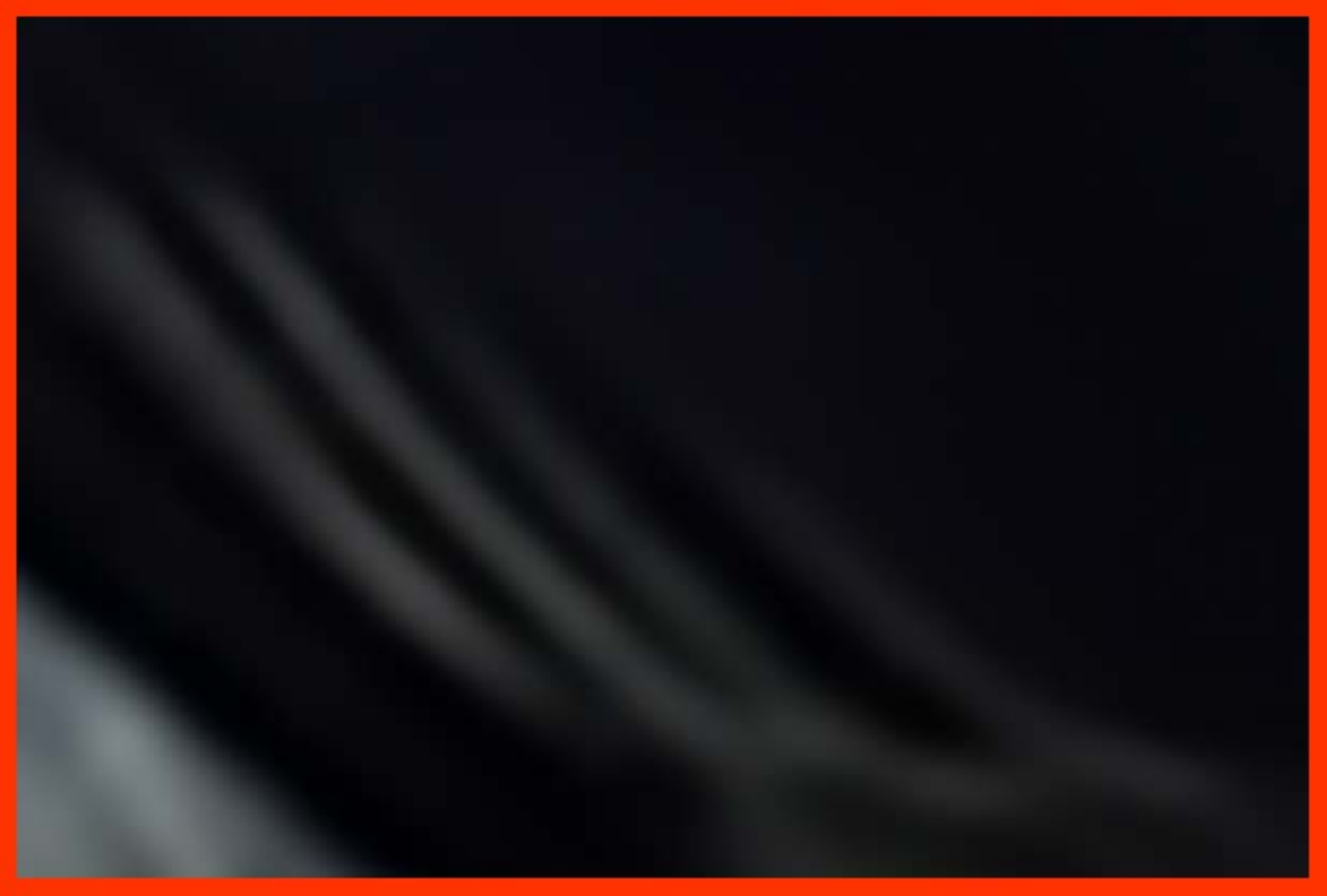} &
		\includegraphics[width=\swsix]{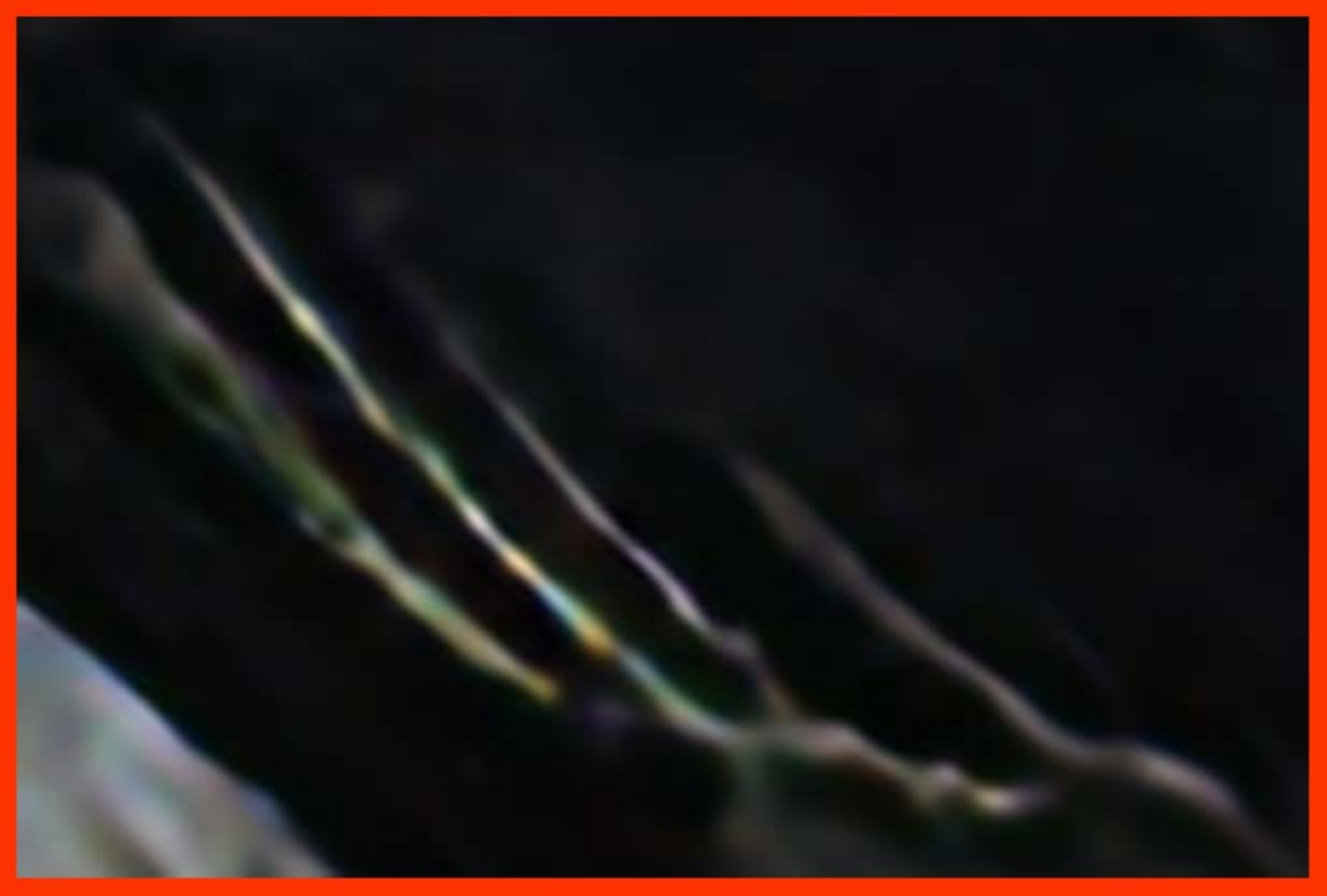} \\
		
		\footnotesize{(a) ground truth} & \footnotesize{(b) bicubic} &  \footnotesize{(c) EDSR$^+$} & \footnotesize{(d) SRGAN$^+$~\cite{ledig2016photo}} &  \footnotesize{(e) BM3D+EDSR} & \footnotesize{(f) \textbf{CinCGAN (ours)}} \\
		\footnotesize{PSNR/SSIM}  & \footnotesize{26.81/0.83} &  \footnotesize{30.28/0.88} & \footnotesize{29.05/0.85} & \footnotesize{26.84/0.86} & \footnotesize{28.26/0.84}
		\end{tabular}
	\end{center}
	\caption{Super-resolution results of ``0801'', ``0816'' and ``0853'' (DIV2K) with scale factor $\times 4$. EDSR$^+$ and SRGAN$^+$ are trained on paired NTIRE2018 track 2 dataset. BM3D+EDSR means using BM3D for denoising first and then using EDSR for super-resolution. The proposed CinCGAN model shows comparable results with SRGAN$^+$ and is better than BM3D+EDSR method.}
	\label{fig:res01_16_53}
\end{figure*}

\subsection{Training details}
We divide our training process into two steps. We first train the model $G_1$, $G_2$ and $D_1$ for mapping LR images to clean LR images (shown as \textit{LR$\to$clean LR} in Fig.~\ref{fig:struc}). The three parameters in (\ref{eq:objLR}) are set to be $w_1 = 10, w_2 = 5$ and $w_3 = 0.5$, respectively. We train our model with Adam optimizer~\cite{kingma2014adam} by setting $\beta_1 = 0.5$, $\beta_2 = 0.999$ and $\epsilon = 10^{-8}$, without weight decay. Learning rate is initialized as $2\times 10 ^{-4}$ and then decreased by a factor of 2 every 40000 iterations. 
The weights of filters in each layer are initialized using a normal distribution and the batch size is set as 16.
We train the model over 400000 iterations, until it converges.

We then jointly fine-tune the LR to HR model (shown as \textit{LR$\to$HR} in Fig.~\ref{fig:struc}). 
We initialize our \textit{SR} network by publicly available EDSR model\footnote{https://github.com/thstkdgus35/EDSR-PyTorch}.
We set parameters in (\ref{eq:objHR}) as $\lambda_1 = 10, \lambda_2 = 5$ and $\lambda_3 = 2$. 
The optimizer is set almost the same as training the \textit{LR$\to$clean LR} model, except for we initialize learning rate with $10^{-4}$. As to the weight of identity loss $\mathcal{L}^{LR}_{idt}$ in (\ref{eq:objLR}), we set $w_2 = 1$.
At each iteration, we update (\ref{eq:objLR}) and (\ref{eq:objHR}) in turn. We first train $G_1$ and $G_2$ to update the \textit{LR$\to$clean LR} network. We then train $G_1$, $SR$ and $G_3$ simultaneously to update the \textit{LR$\to$HR} network. 

We implement the proposed networks with PyTorch and train them on a Nvidia Tesla K80 GPU. It takes about 1 day to pre-train the \textit{LR$\to$clean LR} model and about 2 days to jointly fine-tune the \textit{LR$\to$HR} model.

\subsection{Results}

\renewcommand{\tabcolsep}{10pt}
\begin{table*}[!t]\footnotesize
	\renewcommand{\arraystretch}{1.5}
	\centering
	\caption{Quantitative evaluation on NTIRE 2018 track 2 dataset of the proposed CinCGAN model, in terms of PSNR and SSIM.}
	\label{table:QuantExp}
	\begin{tabular}{|c|c|c|c|c|c|c|c|}
		\hline
		method & bicubic     & FSRCNN~\cite{dong2016accelerating}         & EDSR~\cite{lim2017enhanced}	& EDSR$^+$	& SRGAN$^+$~\cite{ledig2016photo}	  & BM3D+EDSR      & \textbf{CinCGAN (ours)}   \\
		\hline
		\hline
		PSNR	& 22.85 & 22.79 & 22.67 & 25.77 & 24.33 & 22.88 & 24.33  \\
		\hline
		SSIM	& 0.65  & 0.61  & 0.62  & 0.71 & 0.67 & 0.68  & 0.69 \\
		\hline
	\end{tabular}
\end{table*}

We compare the performance of the proposed CinCGAN model with several state-of-the-art SISR methods: FSRCNN~\cite{dong2016accelerating}, EDSR~\cite{lim2017enhanced} and SRGAN~\cite{ledig2016photo}. 
We use the publicly available FSRCNN and EDSR models which are trained with paired LR and HR images, where the inputs are clean LR images down-sampled from HR images.
To make the results more comparable, we also fine-tune EDSR and SRGAN (labelled as EDSR$^+$ and SRGAN$^+$ respectively) with the paired track 2 dataset.
To emphasize the effectiveness of CinCGAN structure, we also try to first denoise the input LR images and then super-resolve the denoised images for comparison. BM3D~\cite{dabov2009bm3d} is one of the state-of-the-art image denoising approach, which is an efficient and powerful denoiser. Hence, we pre-process the test LR images with BM3D first, and then super-resolve it using EDSR (labelled as BM3D+EDSR). 


Table~\ref{table:QuantExp} shows the average PSNR and SSIM values of the restored test images.
It shows that FSRCNN and EDSR cannot work well if the blur and noises are unknown in the training process.
After fine-tuning by paired track 2 dataset, EDSR$^+$ and SRGAN$^+$ improve their results and our method can work comparably against SRGAN$^+$ in terms of PSNR and SSIM without paired training data. 
Although BM3D can remove noise, it also over-smooth the input images. The PSNR and SSIM values of BM3D+EDSR are lower than the proposed method.
Several subjective results are illustrated in Fig.~\ref{fig:res01_16_53}.


\renewcommand{\tabcolsep}{2.4pt}
\begin{figure*}
	\begin{center}
		\begin{tabular}{ccc}
		\hspace{-0.15in}
		\includegraphics[width=0.23\linewidth]{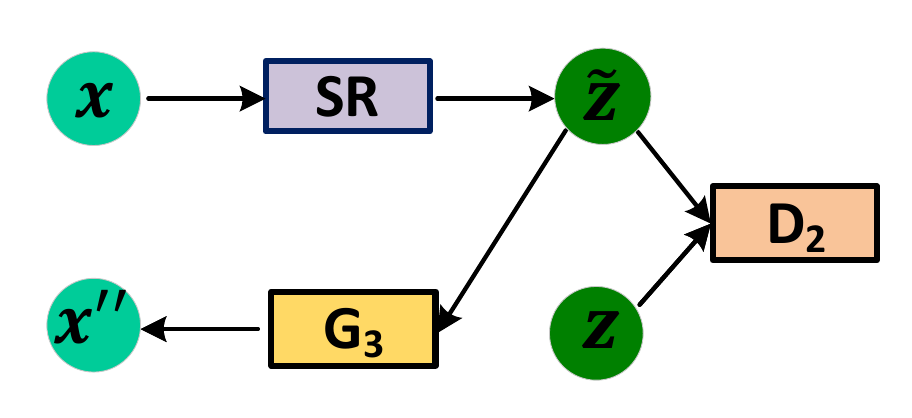} &
		\hspace{0.08in}
		\includegraphics[width=0.35\linewidth]{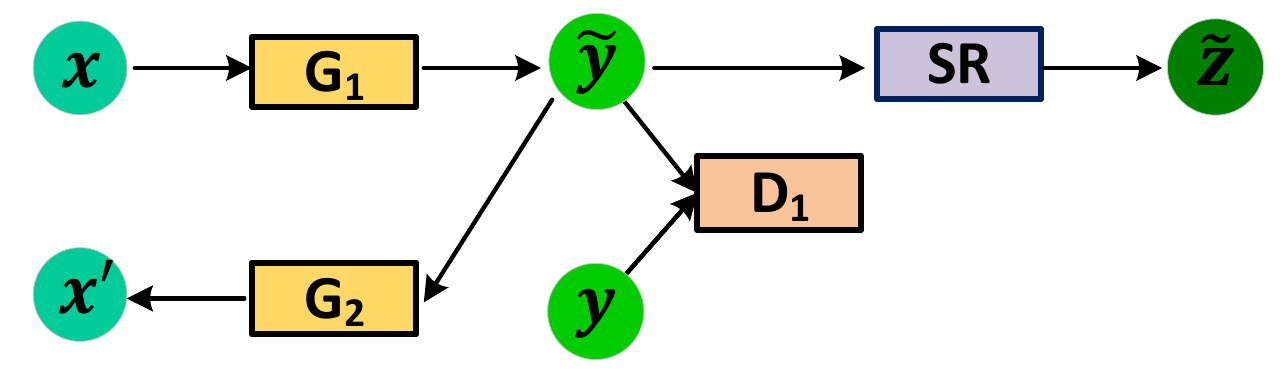} &
		\hspace{0.08in}
		\includegraphics[width=0.40\linewidth]{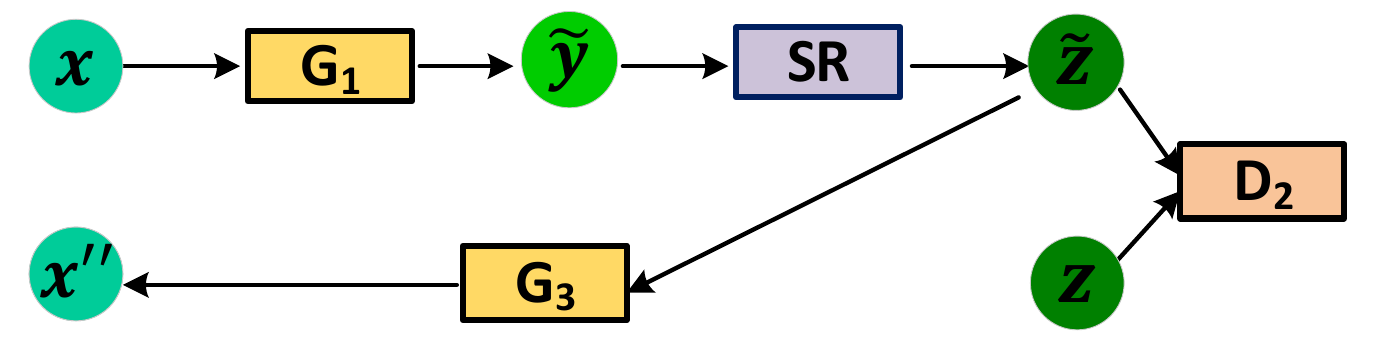} \\
		
		\footnotesize{(a) Structure 1} & \footnotesize{(b) Structure 2} & \footnotesize{(c) Structure 3}
		\end{tabular}
	\end{center}
	\caption{Experiments for validating the advantages of the proposed structure. (a) Structure 1: transform the LR images $x$ to HR images $z$ directly with one CycleGAN model; (b) Structure 2: remove $D_2$ and $G_3$ from the proposed CinCGAN model; (c) Structure 3: remove $D_1$ and $G_2$ from the proposed CinCGAN model.}
	\label{fig:steps}
\end{figure*}

\subsection{Ablation Study}
\label{sec:ablation}

\renewcommand{\tabcolsep}{0.2pt}
\renewcommand{\tabcolsep}{0.1pt}
\begin{figure*}
	\begin{center}
		\begin{tabular}{ccccc}
		\includegraphics[width=\swfive]{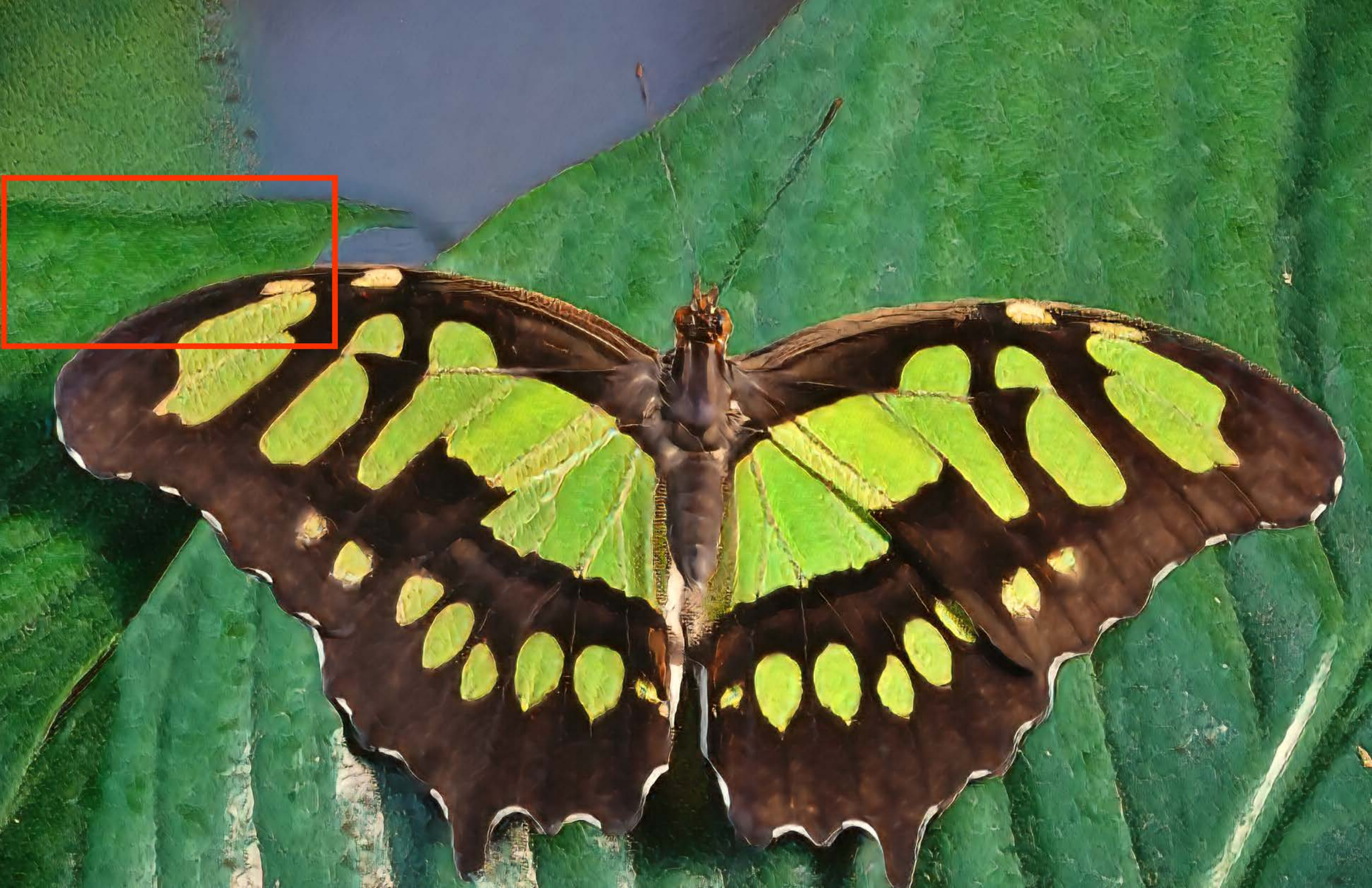} &
		\includegraphics[width=\swfive]{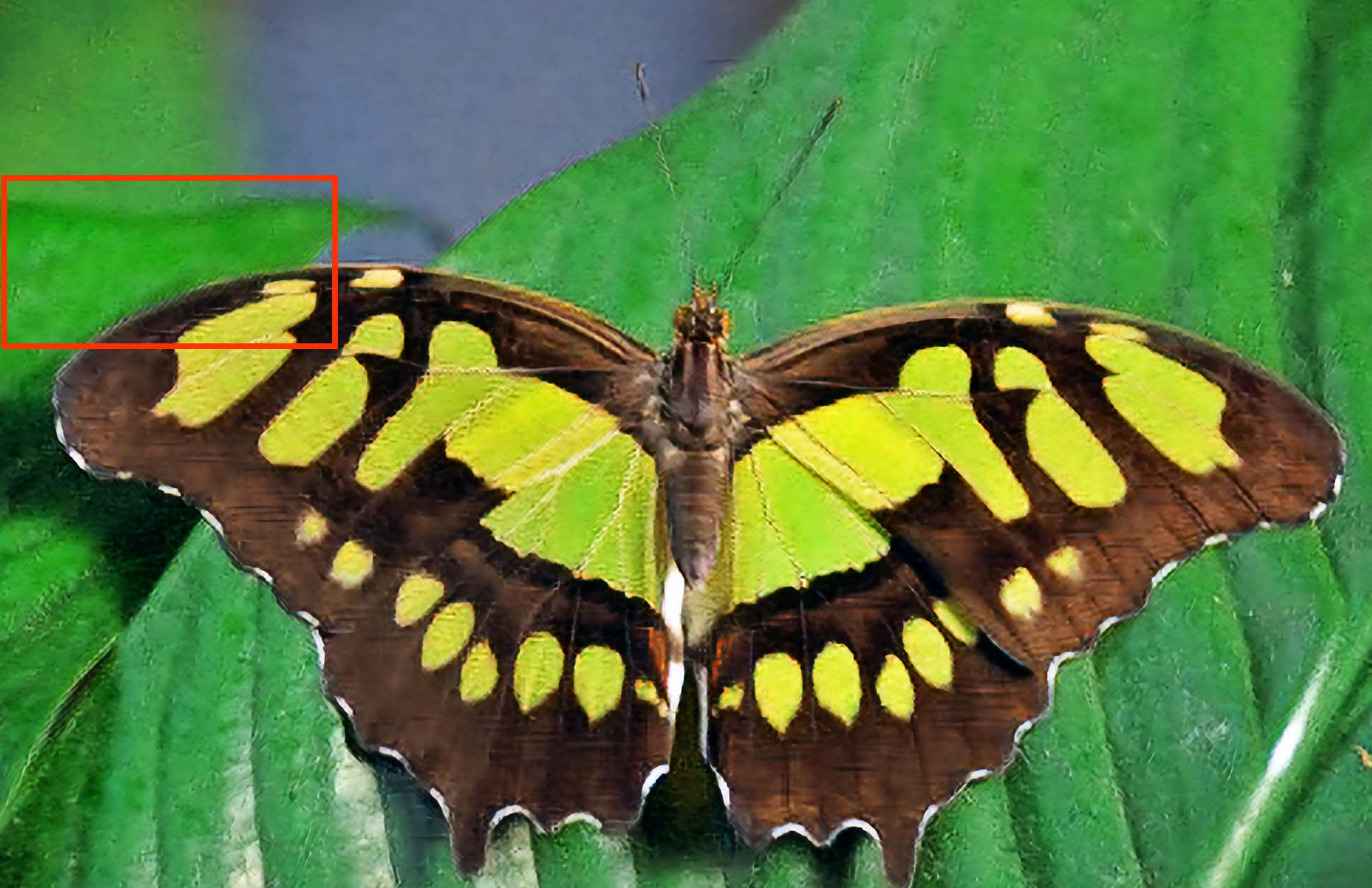} 	&	
		\includegraphics[width=\swfive]{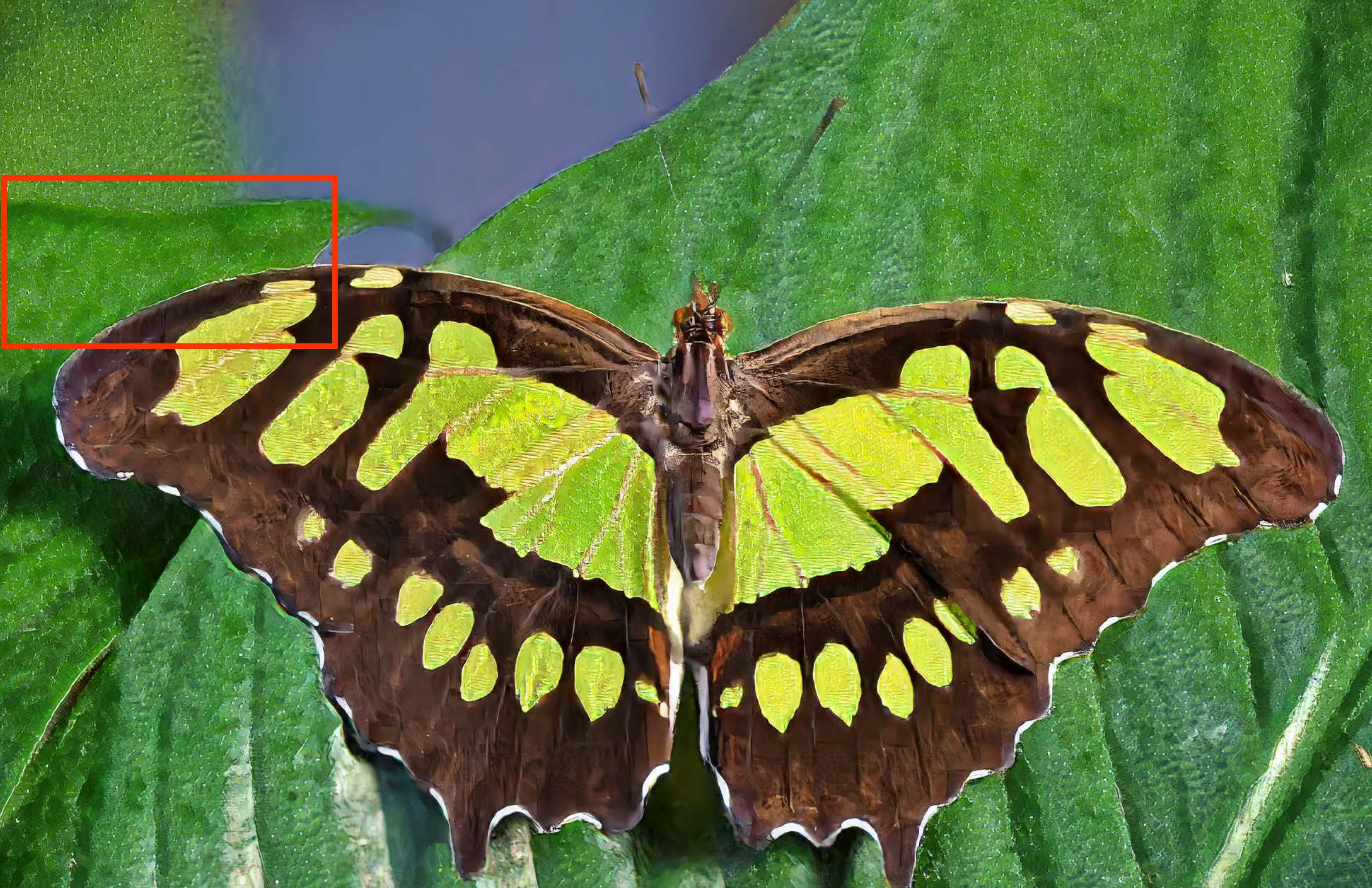} &
		\includegraphics[width=\swfive]{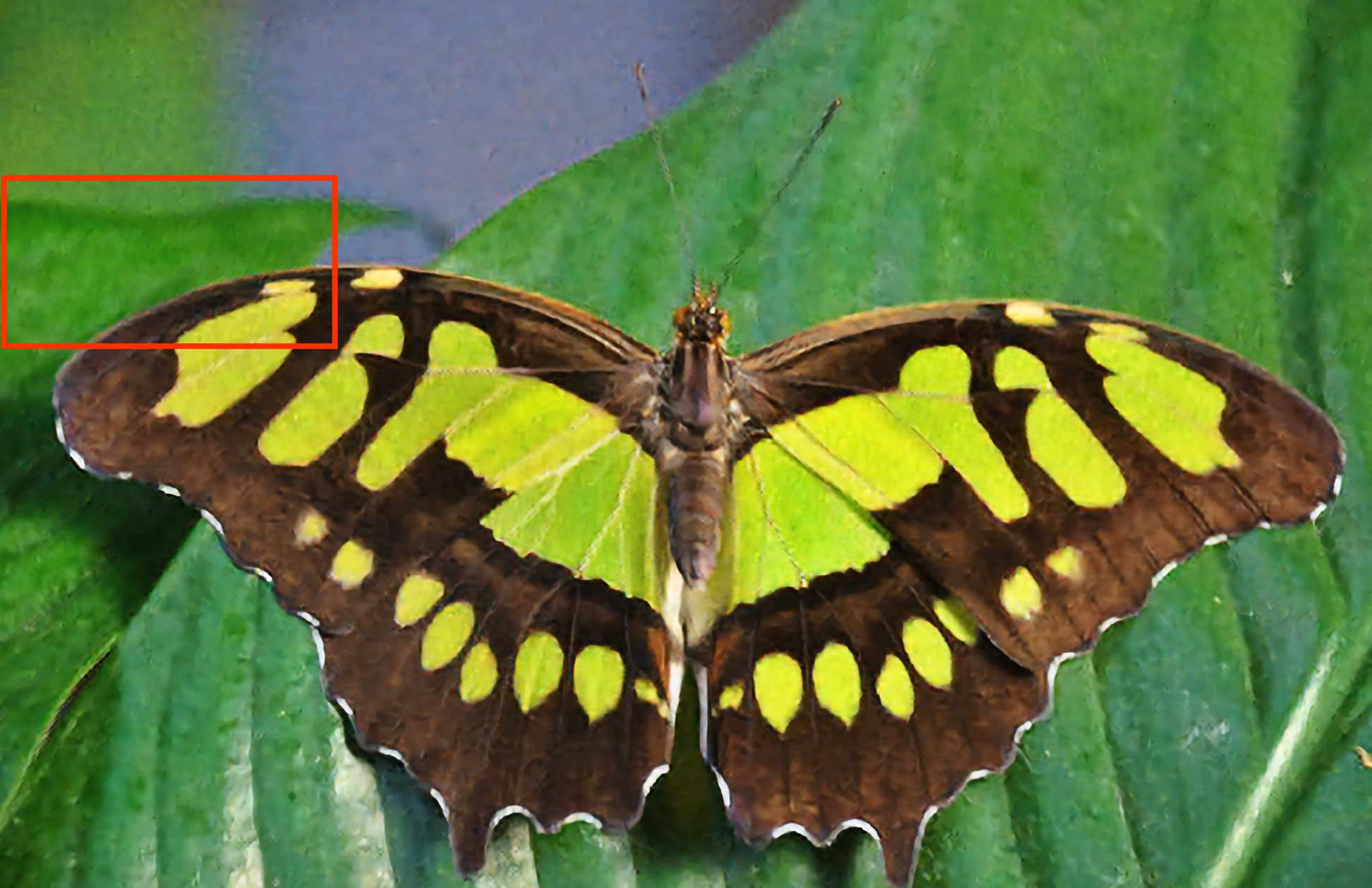} &
		\includegraphics[width=\swfive]{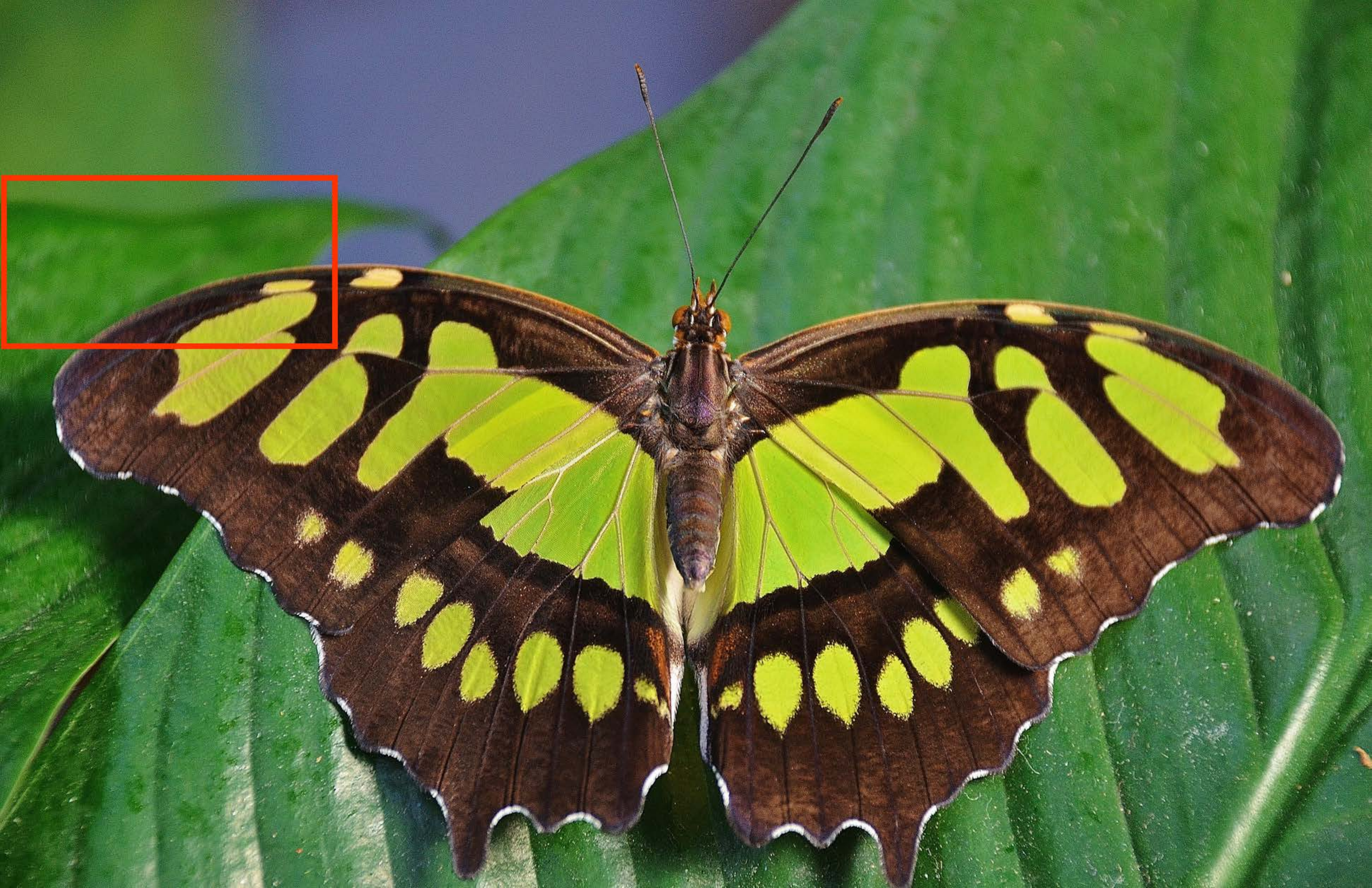} \\
		
		\includegraphics[width=\swfive]{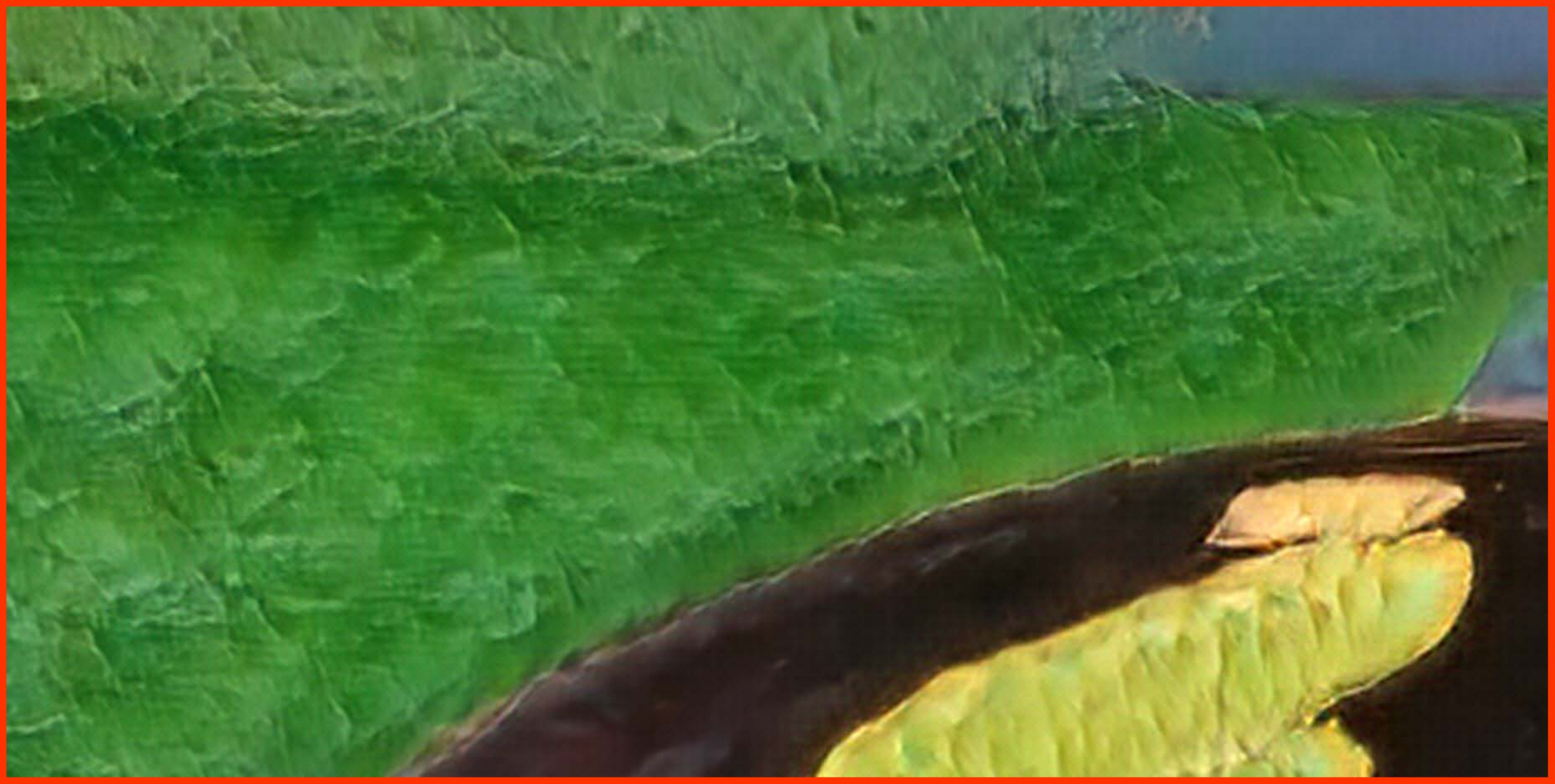} &
		\includegraphics[width=\swfive]{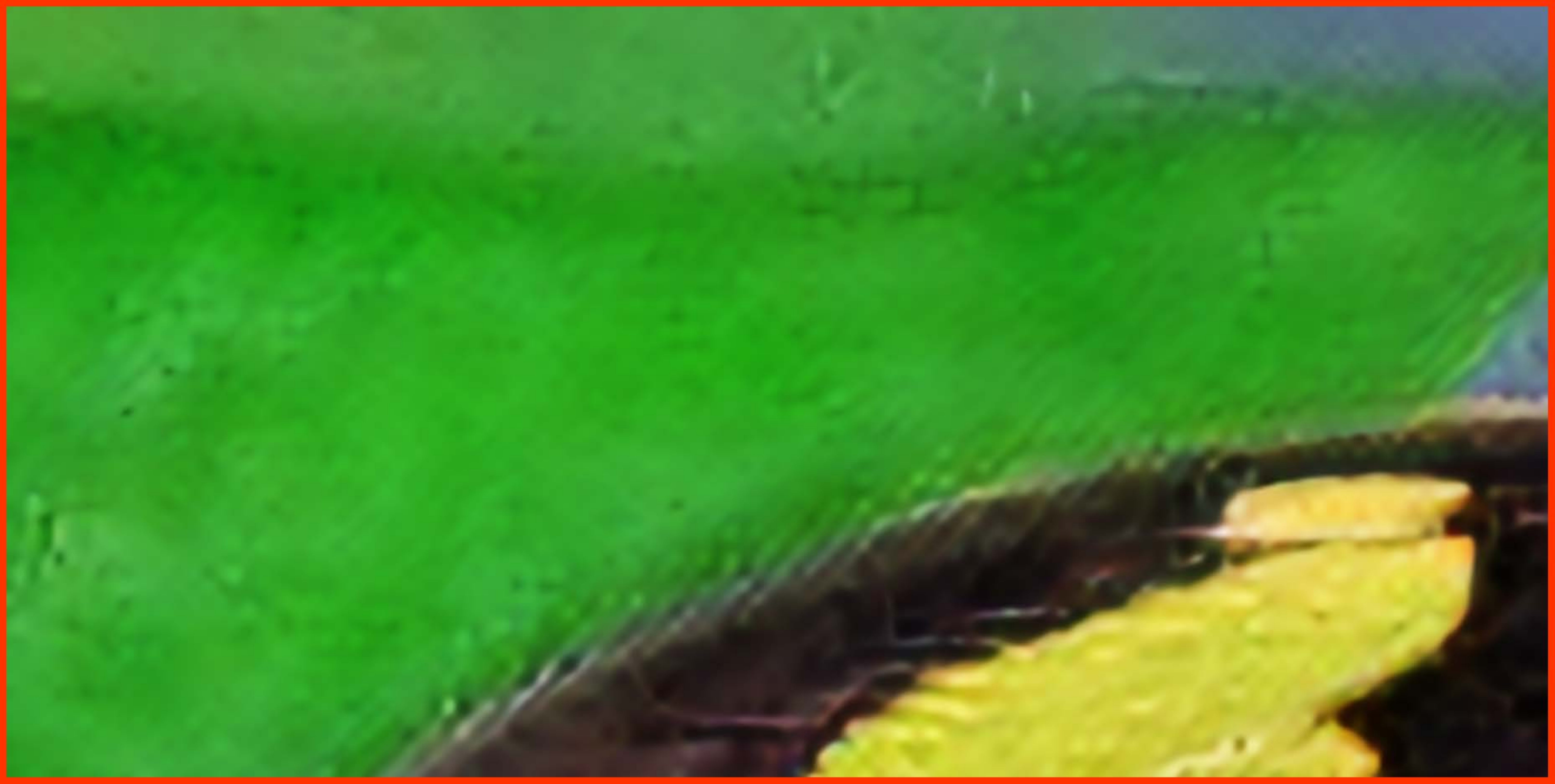} 	&	
		\includegraphics[width=\swfive]{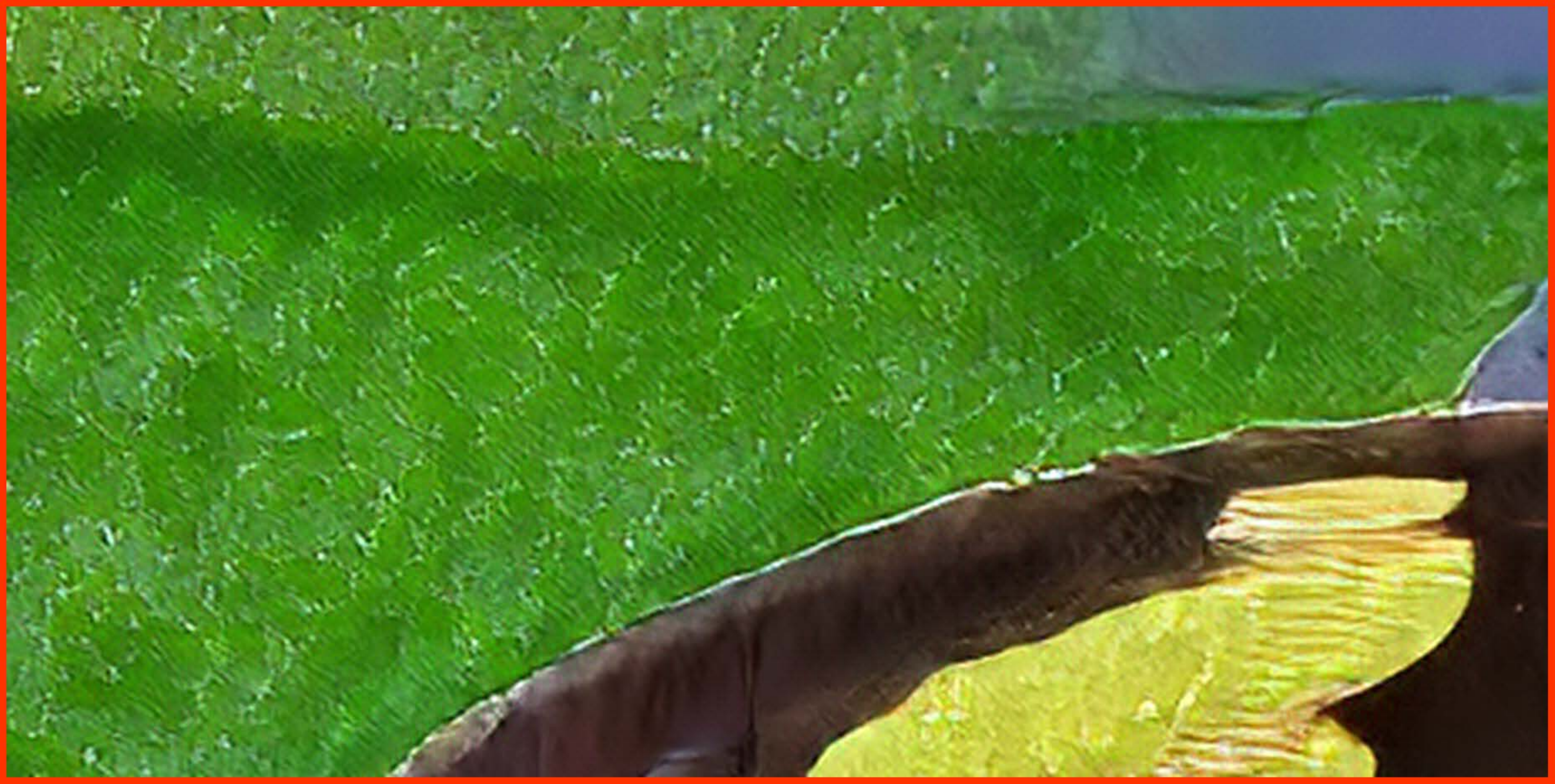} &
		\includegraphics[width=\swfive]{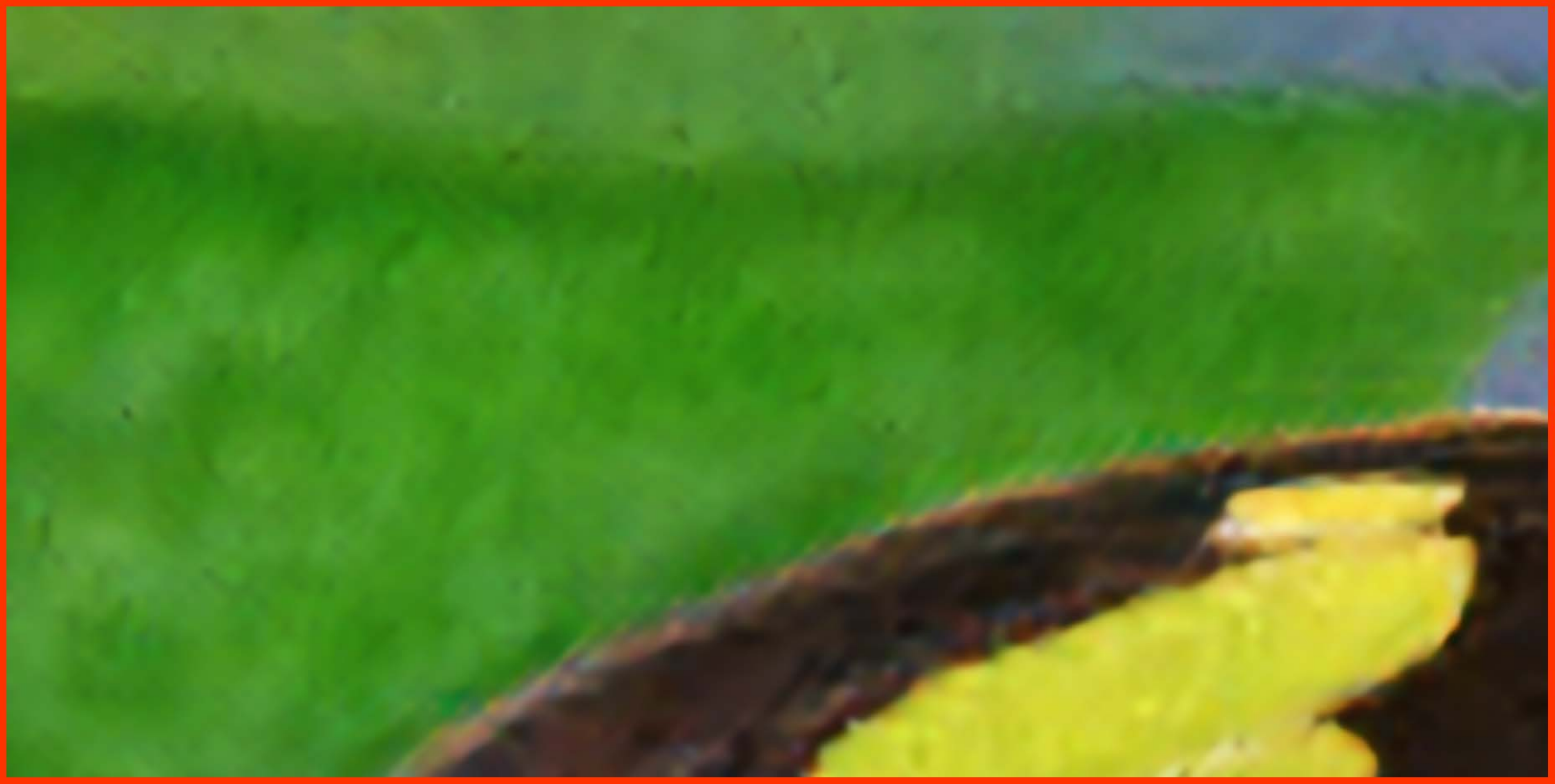} &
		\includegraphics[width=\swfive]{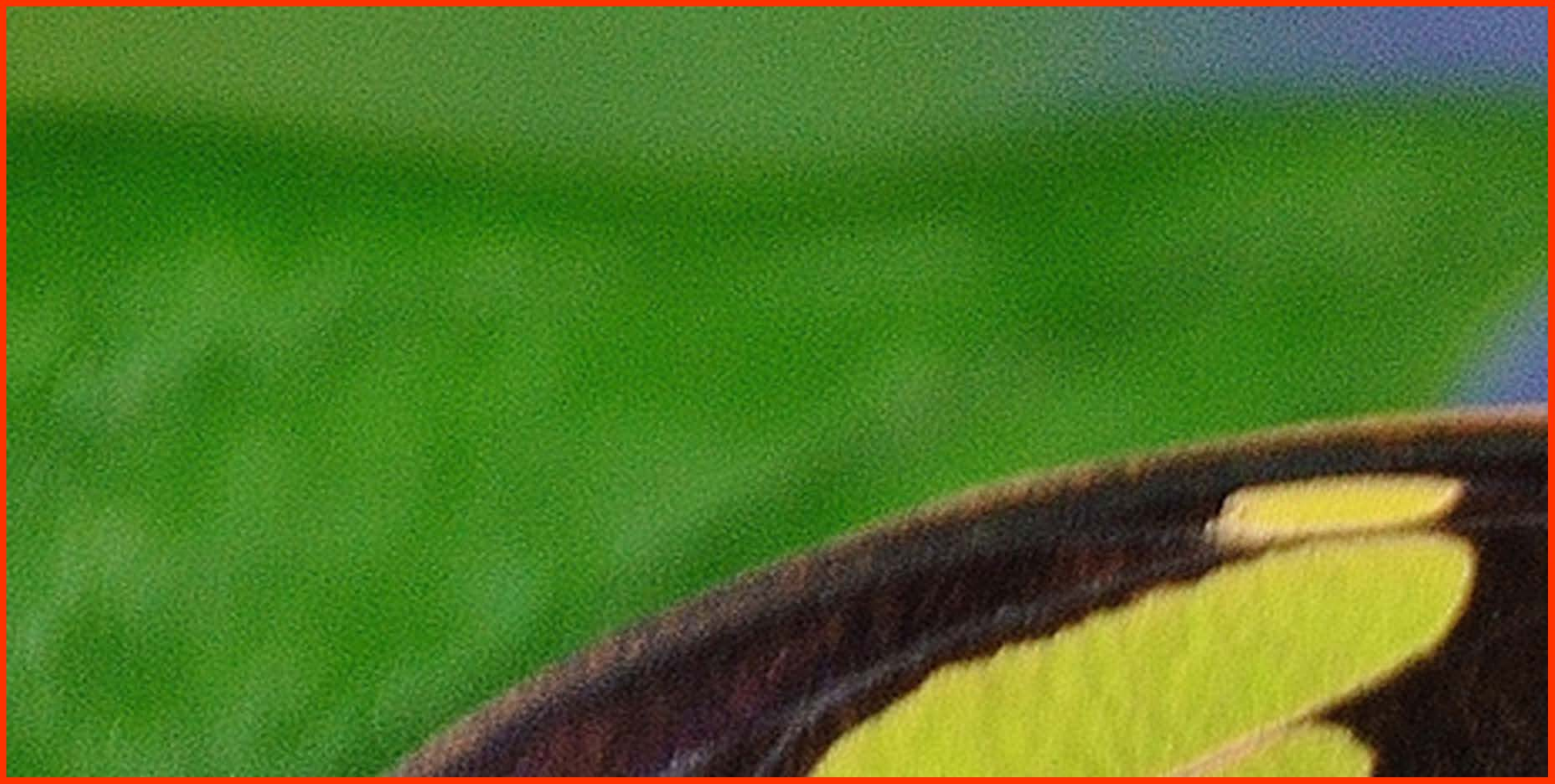} \\
		
       \footnotesize{(a) Structure 1} & \footnotesize{(b) Structure 2} & \footnotesize{(c) Structure 3} & \footnotesize{(d) \textbf{CinCGAN (ours)}}	 & \footnotesize{(e) ground truth}
		\end{tabular}
	\end{center}
	\caption{Super-resolution results of ``0829'' (DIV2K) with scale factor $\times 4$, for each frame structure as described in Fig.~\ref{fig:steps}.}
	\label{fig:res29}
\end{figure*}



To validate the advantages of the proposed CinCGAN model for the unsupervised SISR problem, we design some other network structures for comparison. 

{\flushleft \bf{Structure 1}}
The first frame structure is to restore LR images $X$ to HR images $Z$ using only one CycleGAN, \textit{i.e.} denoise, deblur and super-resolve the LR images at the same time. The structure of the model is shown in Fig.~\ref{fig:steps}(a), where we set an LR image $x$ as input to the \textit{SR} network directly. Correspondingly, we only minimize the total loss $\mathcal{L}^{HR}_{total}$ (with replacing $SR(G_1(\cdot))$ as $SR(\cdot)$ in Eq. (\ref{eq:ganHR})(\ref{eq:cycHR})(\ref{eq:tvHR})). However, during the training procedure, we found that the result $\tilde{z}$ are always unstable and there are a lot of undesired artifacts, as shown in Fig.~\ref{fig:res29}(a). It is hard for a single network to simultaneously denoise, deblur and up-sample the degraded images, especially when the degradation kernels are different from image to image and with unsupervised learning. 

{\flushleft \bf{Structure 2}}
We remove $D_2$ and $G_3$ from the proposed CinCGAN model for our second experiment.
We map the input LR images to a set of clean LR images using the same \textit{LR$\to$clean LR} networks shown in Fig.~\ref{fig:struc}; we then super-resolve the converted LR images directly using the $SR$ network. The whole structure is shown in Fig.~\ref{fig:steps}(b). The corresponding result is illustrated in Fig.~\ref{fig:res29}(b). As we can see, some negligible noise in the resulted clean LR images is magnified and now is visible in the super-resolved images, which affects the visual quality. 

%

{\flushleft \bf{Structure 3}}
Our third experiment is performed by removing $D_1$ and $G_2$ from the proposed CinCGAN model, as shown in Fig.~\ref{fig:steps}(c).
We use one CycleGAN for the LR to HR model, where we take $G_1 + SR$ as the forward network and $G_3$ as the inverse network. $D_2$ is used for distinguishing $\tilde{z}$ from $z$.
We load the pre-trained $G_1$ (in the \textit{LR$\to$clean LR} networks) and the downloaded EDSR models for initialization. 
Experimental results on Fig.~\ref{fig:res29}(c) show that the resulting $\tilde{z}$ are still noisy. 
Since without the $\mathcal{L}^{LR}_{cyc}$ and $\mathcal{L}^{LR}_{GAN}$ constraints on $G_1$ network ($\mathcal{L}^{LR}_{idt}$ and $\mathcal{L}^{LR}_{tv}$ are still used for this model), $G_1$ is unable to deonise and deblur. The whole model becomes similar to Structure 1.

{\flushleft \bf{Proposed Method}}
We then propose our final solution as shown in Fig.~\ref{fig:struc}: jointly fine-tune LR to HR networks with CinCGAN. We sequentially update the \textit{LR$\to$ clean LR}  and the \textit{LR$\to$HR} models. With the two constraint $\mathcal{L}^{LR}_{total}$ and $\mathcal{L}^{HR}_{total}$, the $G_1$ network can denoise and deblur the degraded input image $x$, while the \textit{SR} network can up-sample as well as further restore the resulted intermediate image $\tilde{y}$. The final resulted SR image is shown in Fig.~\ref{fig:res29}(d), which shows the best visual result comparing with other three structures.

\label{sec:expr}

\section{Conclusions}

We investigate the single image super-resolution problem with a more general assumption: the low-/high-resolution image pairs and the down-sampling process are unavailable. Inspired by the recent successful image-to-image translation applications, we resort to the unsupervised learning methods to solve this problem. Using generative adversarial networks (GAN), the proposed method contains two CycleGANs, where the second GAN covers the first one. The solution pipeline consists of three steps. First, we map the input LR images to the clean and bicubic-downsampled LR space with the first CycleGAN. We then stack another well-trained deep model with bicubic-downsampling assumption to up-sample the intermediate result to the desired size. Finally, we fine-tune the two modules in an end-to-end manner to get the high-resolution out. Experimental results demonstrate that the proposed unsupervised method achieves comparable results as the state-of-the-art supervised models.

\vspace{0.5in}
\textbf{Acknowledgement.}  This work is supported by SenseTime Group Limited and in part by the Projects of National Science Foundations of China (61571254), Guangdong Special Support plan (2015TQ01X16), and Shenzhen Fundamental Research fund (JCYJ20160513103916577).

\label{sec:conclude}

\clearpage

{\small
\bibliographystyle{ieee}
\bibliography{egbib}
}
\end{document}